\documentclass[10pt,twocolumn,letterpaper]{article}

\usepackage{wacv}
\usepackage{times}
\usepackage{epsfig}
\usepackage{graphicx}
\usepackage{amsmath}
\usepackage{amssymb}
\usepackage{comment}

\usepackage{color}
\usepackage{comment}
\usepackage{multirow}
\usepackage{multicol}

\usepackage[font=small]{caption}
\usepackage[font=small]{subcaption}
\usepackage{framed,enumitem}  
\usepackage{algorithm2e}
\usepackage{booktabs}
\usepackage{makecell}
\usepackage{array,ragged2e}
\usepackage{boldline}
\usepackage{adjustbox}

\def\wacvPaperID{16} 

\wacvfinalcopy 

\usepackage[pagebackref=true,breaklinks=true,colorlinks=true,bookmarks=false]{hyperref}

\begin{document}

\title{Domain Adaptive Knowledge Distillation for Driving Scene Semantic Segmentation}

\author{Divya Kothandaraman
\qquad
Athira Nambiar
\qquad
Anurag Mittal \\
{\tt\small ramandivya27@yahoo.in, \{anambiar,amittal\}@cse.iitm.ac.in}\\
Indian Institute of Technology Madras, India
}

\maketitle
\date{}

\begin{abstract}
Practical autonomous driving systems face two crucial challenges: memory constraints and domain gap issues. In this paper, we present a novel approach to learn domain adaptive knowledge in models with limited memory, thus bestowing the model with the ability to deal with these issues in a comprehensive manner. We term this as "Domain Adaptive Knowledge Distillation" and address the same in the context of unsupervised domain-adaptive semantic segmentation by proposing a multi-level distillation strategy to effectively distil knowledge at different levels. Further, we introduce a novel cross entropy loss that leverages pseudo labels from the teacher. These pseudo teacher labels play a multifaceted role towards: (i) knowledge distillation from the teacher network to the student network \& (ii) serving as a proxy for the ground truth for target domain images, where the problem is completely unsupervised. We introduce four paradigms for distilling domain adaptive knowledge and carry out extensive experiments and ablation studies on real-to-real as well as synthetic-to-real scenarios. Our experiments demonstrate the profound success of our proposed method. 
\end{abstract}


\section{Introduction}

Semantic segmentation is one of the fundamental tasks in perception for autonomous vehicles. The goal of semantic segmentation is to assign a label (car, tree, sky, etc.) to every pixel in the image. This information is quite relevant in the decision making of autonomous vehicles, as it provides information regarding shape/location of objects and the background. Recently, deep neural networks have imparted immense progress in this field. Some benchmark architectures include Fully Connected Networks \cite{long2015fully}, Dilated Residual Networks \cite{yu2017dilated},  and DeepLab \cite{chen2017deeplab}, PSPNet \cite{zhao2017pyramid}, and Gated-scnn \cite{takikawa2019gated}. 

Despite these advancements, there are some practical challenges that cause a hindrance to even the best performing models. One such major drawback is the \textbf{\textit{``domain-gap" issue}}. It particularly exists in `in-the-wild' scenarios, where neural networks trained on one domain won't generalize well on other domains, due to large domain gaps between the source (on which the neural network was trained) and the target (test) domains. This is mainly due to significant changes in the illumination, weather conditions and seasonal changes across domains. Additionally, collecting ground-truth (GT) annotations for different domains is a laborious and expensive task. This calls for an approach, that is capable of leveraging existing ground truth information in some domains to improve performance in other domains in an unsupervised manner. To this end, Unsupervised Domain Adaptation (UDA) has emerged as a solution. The goal of UDA is to align features (reduce domain gap) across the source (labelled) and target (unlabelled) domains, so that the adapted model performs well on both domains. 

Another key issue at the production end is the \textbf{\textit{memory constraint}} of the system. Most of the best-performing models are very deep and have high memory requirements. However, practical applications require the model to fit and evaluate on large images (as captured from a car) with limited memory, as in mobile devices. Many recent approaches have been inviting attention towards using compact light weight models \cite{mehta2018espnet},\cite{zhao2018icnet},\cite{paszke2016enet}. However, these models result in lower performance efficiency in comparison with the state-of-the-art (SOTA) architectures that use deep backbones. With respect to this, model compression techniques like Knowledge Distillation \cite{hinton2015distilling},\cite{ros2016training} have contributed significantly in fuelling the performance of a compact \textit{Student} network by transferring knowledge from a cumbersome \textit{Teacher} network.

In this paper, we propose a novel holistic end-to-end approach for distilling domain adaptive segmentation knowledge to tackle the challenges of domain gap and memory constraint, simultaneously. The goal is to distil domain-adaptive knowledge from a computationally expensive teacher network to a compact student network, to boost its performance in the target domain. To the best of our knowledge, this paper marks one of the first approaches towards this multifaceted goal. To facilitate our proposal, we propose a \textit{multi-level distillation} (MLD) strategy with novel loss functions to improve performance for both knowledge distillation (KD) and domain adaptation (DA) in a simultaneous manner. 

Our proposed method distils knowledge at two levels - feature space and output space - using a combination of KL divergence and MSE loss terms. Furthermore, in order to optimize our model for both tasks simultaneously, we discern the need for an additional loss function that is tailored explicitly to serve the multitudinous goal. In this regard, we leverage the teacher network soft predictions to generate pseudo labels. These pseudo labels enhance knowledge transfer at the output space. These labels are particularly relevant in the target domain, as they serves as a proxy for ground truth, which is otherwise unavailable. Our newly presented cross entropy loss function with pseudo-labels serves as a multifarious objective function, by serving as an emissary for the unavailable ground truth in the target domain, while also amplifying the distillation from the teacher to the student.  

We evaluate our proposed method on large-scale autonomous driving datasets. In addition to the benchmark synthetic-to-real adaptation case scenario (GTA5 to Cityscapes), we assess our pipeline on a real-to-real scenario too (BDD to Cityscapes). In summary, the key contributions of this paper are as follows: \
 
(1) A novel framework that distils domain adaptive knowledge to address two core practical problems (i.e. low memory networks \& domain gap) concurrently.; (2) A \textit{multi-level distillation} strategy carried out at the feature and output spaces.; (3) Proposal of a tailor-made loss function: \textit{cross entropy with pseudo teacher labels} for this multifaceted problem.; (4) Extensive experimental analysis and ablation studies on both \textit{synthetic-to-real}, and on \textit{real-to-real} cases.

\section{Related work}
\subsection{Domain adaptation} 
Domain adaptation is a subfield of machine learning that deals with employing a model trained on a source distribution in the context of a different (but related) target distribution. The goal is to reduce the domain gap and to find a common feature space that works well across domains. This problem has seen immense progress in the past. Domain gaps can be reduced in an adversarial way \cite{tzeng2014deep, tzeng2017adversarial}, by reconstruction \cite{ghifary2016deep, murez2018image} or by minimizing distance between distributions \cite{long2015learning, sun2016deep}. 

Domain-adaptive semantic segmentation is of keen interest to researchers as semantic labelling is an incredibly labor-intensive task. One of the first methods was reported in \cite{hoffman2016fcns}. The method uses fully convolutional networks for semantic segmentation and adversarial training for domain alignment. Kalluri et al. \cite{kalluri2018universal} performs semi supervised domain adaptation by using an entropy module that minimizes a similarity metric computing by projecting the representation at each pixel to an embedding space of different directions. Murez et al. \cite{murez2018image} uses cycleGAN to transform images from one domain to another, and then uses both the original and the transformed images to find a model agnostic feature space. Tsai et al. \cite{tsai2018learning} uses multi level adversarial learning to align features at different levels of the network. Vu et al. \cite{vu2019advent} uses losses based on the entropy of pixel wise predictions. Chen et al. \cite{chen2019crdoco} uses an image-to-image translation network, domain-specific task networks and cross domain consistency to find a common feature space. 

\begin{figure*}[!th]
\begin{center}
\centering
\includegraphics[width=15cm,height=7.7cm]{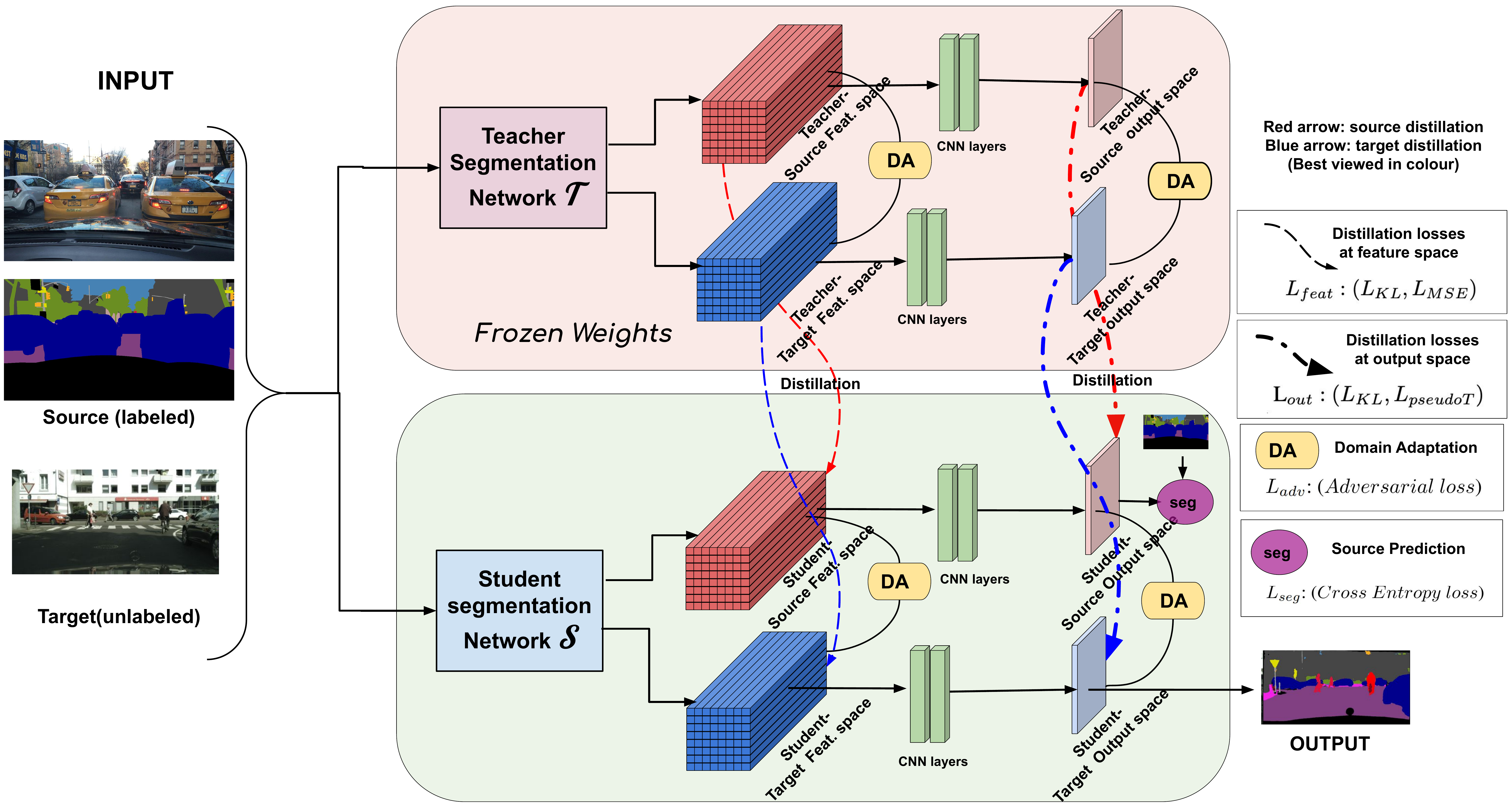}
\end{center}
\vspace{-0.3cm}
\caption{Our proposed framework for unsupervised domain adaptive knowledge distillation for driving scene semantic segmentation.
\label{fig:KDforSS}}
\end{figure*}

\subsection{Model compression via knowledge distillation:}
Recently, there has been a surge in papers addressing model compression techniques in various vision problems. A thorough review of various compression techniques is reported in the survey paper \cite{cheng2017survey}. One of the pioneer works in the field by Hinton et al. \cite{hinton2015distilling} showed that the extra information a.k.a. ``dark knowledge" encapsulated in the soft targets (class probabilities produced by teacher) could be used for training the compact student model. Motivated by this observation, several other works were followed leveraging more teacher information. For instance, Romero et al. \cite{romero2014fitnets} presented an idea for distilling from the intermediate feature representations, whereas Yim et al. \cite{yim2017gift} proposed to distil the knowledge in terms of flow between layers, which is calculated by computing the inner product between features from two layers. In addition to various KD models in image classification \cite{chen2018knowledge},\cite{cho2019efficacy}, it has been successfully applied in object detection \cite{chen2017learning}, video classification \cite{bhardwaj2019efficient} and speech recognition \cite{wong2016sequence} as well. The idea has been explored for semantic segmentation tasks as well. Ros et al. \cite{ros2016training} investigated the problem of road scene semantic segmentation leveraging different knowledge transfer frameworks based on the output probability of the teacher network. Further work includes distilling zero-order and first-order knowledge from teacher models \cite{xie2018improving} and distilling structured knowledge via pairwise and holistic distillation \cite{liu2019structured}. The paper on feature level distillation \cite{crasto2019mars} explores the topic in the setting of action recognition.\\

\subsection{Differences from past methods}
While we acknowledge that there has been good headway on the ideas of domain adaptation and knowledge distillation as two separate problems, it is notable that in this paper, we conceptualize the best of both in a single holistic model, in a novel way. To the best of our knowledge, this is for the first time that the usage of domain adaptation along with knowledge distillation has been explored in semantic segmentation. Albeit some early works addressed both of these concepts, those were on other problem settings such as Semi-supervised domain adaptation (SDA) \cite{ao2017fast},\cite{orbes2019knowledge} or upon clinical/acoustic platforms \cite{orbes2019knowledge},\cite{asami2017domain}. Further, they used knowledge distillation as a means for domain adaptation. On the contrary, we address a much more difficult problem setting  - Unsupervised Domain Adaptation (UDA), on autonomous driving scene understanding, an area of stupendous practical implication. Unlike leveraging distillation as a means for domain adaptation \cite{deng2019cluster},\cite{nath2019adapt},\cite{zhao2019multi}, the design of our architecture is very well fine-tuned to serve both the goals of compression \& domain adaptation in a specialized manner. Pseudo labels have been used for self-training in certain domain adaptation architectures \cite{zou2018unsupervised, choi2019pseudo, ge2020mutual, morerio2020generative}. Additionally, noisy labels have also been used for partial knowledge transfer \cite{li2017learning, wang2018kdgan}. In contrast, our paper proposes a method that leverages pseudo labels and improves both domain adaptation performance and knowledge distillation performance in a complementary manner. 

\section{Distillation model for UDA semantic segmentation}
We propose a novel framework for distilling domain adaptive knowledge from a cumbersome teacher network to a lightweight student network, in the context of unsupervised semantic segmentation. In this section, the overall network architecture, distillation strategies and the associated objective functions are explained.

\subsection{Overview of Network Architecture}

The proposed network architecture is shown in Fig. \ref{fig:KDforSS}. It consists of two neural networks \textit{viz.} a cumbersome teacher segmentation network \textit{$\mathcal{T}$} and a compact student segmentation network \textit{$\mathcal{S}$}. RGB images from both source and target domains and the segmentation maps for the corresponding source domain images are the inputs to the networks. The problem is completely unsupervised in the target domain \textit{i.e.,} no labels are available for the target domain.
First, the teacher \textit{$\mathcal{T}$} and the student \textit{$\mathcal{S}$} networks  are initialized with parameters trained for the task of domain adaptation (DA) on a source-target dataset pair. The  baseline DA model used in both student and teacher networks is detailed in later Section \ref{sec:DA}. The teacher network \textit{$\mathcal{T}$} (weights frozen after training) is used to distil knowledge to the compact student network \textit{$\mathcal{S}$} (distillations are shown in Fig. \ref{fig:KDforSS} via arrows). The task of the student network is to learn a domain-adaptive set of parameters, while taking advantage of the mastery of the teacher network. The transfer of knowledge can help the much smaller student network perform better than its innate capabilities.  

In order to facilitate efficient knowledge distillation, we propose a multi-level distillation scheme, wherein knowledge transfer is carried out at multiple levels of the network: feature space and output space, as explained in Section \ref{sec:MLD}. Further, two modes of distillations are designed at each level \textit{i.e.} source distillation (red arrows) and target distillation (blue arrows). In addition, at the output space, we introduce a loss function that leverages pseudo labels from the teacher network. These pseudo labels serve two purposes: knowledge distillation from the teacher to the student \& proxy for the target domain ground truth, which is otherwise unavailable. A detailed explanation on the pseudo teacher label based distillation is provided in Section \ref{sec:QTL}. 
\subsection{Baseline domain adaptation (DA) model}
\label{sec:DA}
It is important to use a good domain adaptation strategy for each of the teacher and student networks. To this end, we use the popular multi-level domain adaptation strategy \cite{tsai2018learning} to pretrain both the networks. To train the generator, segmentation loss $L_{seg}$  (cross entropy loss) is applied on images from the source domain, and adversarial loss $L_{adv}$ is applied for images from the target domain. The discriminators are optimised with a binary cross entropy loss applied on images from both the domains. The mathematical formulations \cite{tsai2018learning} are as follows:
\begin{equation}
    L_{seg} = -\sum_{h,w} \sum_{c \in C} Y_{s}^{(h, w,c)} \log(P_{s}^{(h, w,c)})   
\end{equation}
\begin{equation}
    L_{adv} = -\sum_{h,w}log(D(P_{t})^{h,w,1}) 
\end{equation}
where $Y_{s}$ represents ground truth, $P\textsubscript{s}$ is the probability map predictions for source images and 
$P_{t}$ is the probability map predictions of target images. \textit{C} represents the number of classes. The net equation to optimise the generator network is a sum of the segmentation and adversarial loss terms at feature and output level. This is optimized using the min-max criterion. 
\subsection{Multi-level distillation}
\label{sec:MLD}
We present a multi-level distillation method to transfer knowledge at both the feature and the output space to bestow the student network with improved performance capabilities while retaining its compactness. In this regard, we present the following loss functions:

\subsubsection{KL divergence loss $L_{KL}$}
The output of the networks are probability distributions. We use the KL divergence loss \cite{kullback1951information} to motivate the student to achieve distributions close to the teacher at the output level. This encourages the output of the student network to emulate the output of the teacher network. 
\begin{equation}
    L_{KL} = \lambda _{KL} * \sum_{i} KL(q_{i}^{s} || q_{i}^{t})
\end{equation}
Here, $\lambda _{KL}$ represents the weight hyperparameter for KL divergence distillation, $q_{i}^{s}$ denotes student domain probability maps and $q_{i}^{t}$ denotes the corresponding counterpart for teacher domain. To facilitate enhanced distillation, we enforce another KL divergence loss term at the high dimensional feature space (where feature-space adaptation is done), nevertheless with a lower weight hyperparameter. Note that, in consistency with multi-level DA, AdaptSegNet \cite{tsai2018learning}, we project the feature maps to segmentation probability maps. This enables us to apply KL divergence distillation at the feature space (Mathematically, KL divergence can be used to only constrain probability distributions). Thus, this loss is implemented at two stages viz., output and feature levels and we denote them as $L_{KL,out}$ and $L_{KL,feat}$ respectively. It is mathematically formulated as: 

\begin{equation}
L_{KL} = L_{KL,out} + L_{KL, feat} 
\end{equation}

\subsubsection{MSE loss $L_{MSE}$}

While the output space captures the crux of knowledge (scene layout, context, etc.) needed for distillation, the high dimensional feature space encodes complex representations and prove useful too. This leads us to impose a mean square (MSE) loss $L_{MSE}$ between the feature maps of the two networks, which pushes them to resemble each other. KL divergence at the feature space aims to align probability distributions (soft alignment), while MSE pushes the feature spaces to exactly resemble each other (hard alignment). While MSE can promote the final layers of the 2 networks to exactly resemble each other, probability distributions can be adversely affected. KL divergence (log based loss) acts as a regularizer, while simultaneously aiding distillation. $L_{MSE}$ is formally represented as:
\begin{equation}
    L_{MSE} = \lambda_{MSE}  * \sum_{i} ||(q_{i}^{s} - q_{i}^{t})||\textsuperscript{2}
\end{equation}
    
These feature maps correspond to the level at which feature-level domain adaptation is done. If the feature size between the student and teacher model is different, bilinear upsampling (proven to be effective in generating segmentation maps from feature maps) can be used to make the feature sizes same, followed by the application of the loss functions.

\begin{figure*}[!th]
\begin{center}
\includegraphics[width=12cm,height=3.3cm]{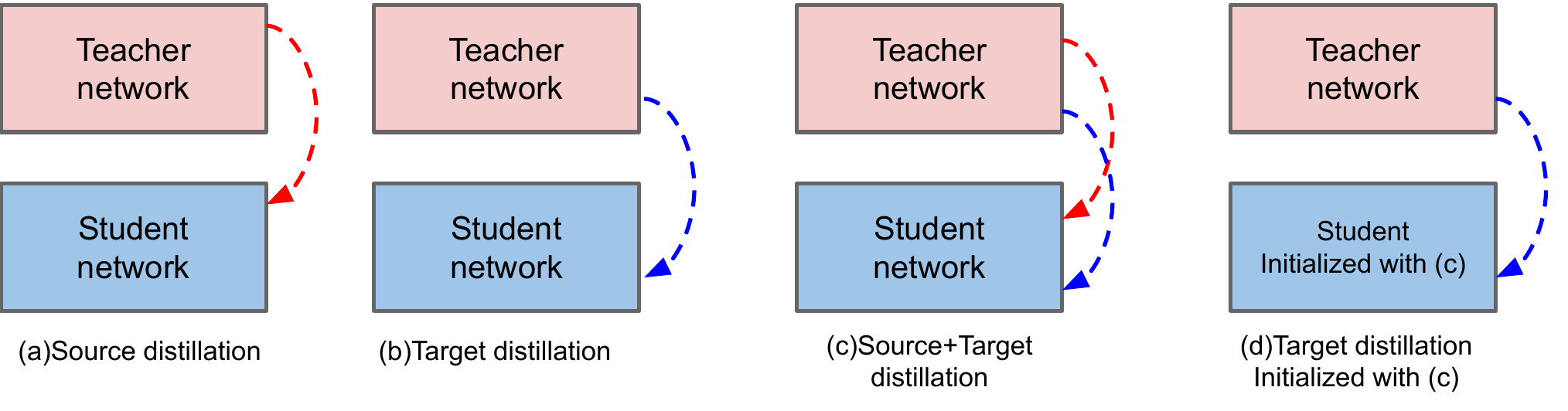}
\end{center}
\vspace{-0.5cm}
\caption{Various distillation paradigms. Source distillation and target distillation are shown in red and blue arrows respectively. (Best viewed in colour)}
\vspace{-0.2cm}
\label{fig:Distillations}
\end{figure*}

\subsubsection{Domain-adaptive distillation with pseudo teacher labels  $L_{pseudoT}$ } 
\label{sec:QTL}
This distillation happens at the output space, in addition to the distillation using KL-divergence loss. Based on the soft labels (probability maps) provided by the teacher, we determine the class that each pixel belongs to generate pseudo labels. Although these pseudo labels generated by the teacher are not totally accurate, the teacher network nevertheless performs better than the undistilled student network. These pseudo labels serve two purposes: \\ 

\noindent \textbf{(i) Knowledge distillation for images in the source domain} - The information from source domain pseudo-labels is distilled to the counterpart student network via multi-class cross entropy loss term. This is in addition to the segmentation multi-class cross entropy loss $L_{seg}$ that is applied using the ground truth for images from the source domain. \\\textbf{(ii) Proxy ground truth for images in the target domain}, where the problem is completely unsupervised. The enchantment of these pseudo-labels is mainly exhibited in the case of the student target domain, wherein teacher domain pseudo-labels serve as a proxy for the ground truth. On this notion, we apply  cross entropy loss between the student network target domain outputs and the corresponding pseudo-labels from teacher, that we term as $L_{pseudoT}$. While one may argue that for probability distributions, KL divergence and cross entropy play very similar roles and are thus equivalent to each other, we wish to emphasize that KL divergence is applied between the teacher and student probability distributions wherein the teacher distribution carries information about all classes. This can thus capture knowledge about semantically similar classes.
However, for the pseudo label cross entropy loss term, we deterministically predict the label for each pixel from the teacher network probability distribution. This carries no information about the labels for other classes, and can also introduce false positives, which is countered with the KL divergence loss term. We observe that these pseudo-labels significantly improve the results. Mathematically, 
\vspace{-0.4cm}
\begin{equation}
    L_{pseudoT} = \lambda_{pseudoT} * \sum_{h,w} \sum_{c} Y_{pseudoT}\log(P_{student})
\end{equation} 
$\lambda _{pseudoT}$ represents the hyperparameter for distillation with pseudo-labels, $Y_{pseudoT}$ denotes pseudo-labels from the teacher network, and $P_{student}$ denotes predictions (probability map) of the student network. The equation takes the same form for both source and target domain images. \\
\noindent \textbf{Overall objective function:}

The student network is trained with both domain adaptation losses (section \ref{sec:DA}), and distillation losses. The overall objective function for distillation is the weighted sum of the individual distillation loss functions. Taking advantage of the similitude between the source (s) and target (t) domains, we propose to introduce a scaling factor $\lambda_{target}$. For target domain distillation, we use parameters that are $\lambda_{target}$ times the corresponding parameters used for source domain distillation. 
\vspace{-0.4cm}
\begin{equation}
\begin{aligned}
L_{distill} = [L_{s-KL} + L_{s-MSE} + L_{s-pseudoT}]  + \\ \lambda_{target}*[L_{t-KL} + L_{t-MSE} + L_{t-pseudoT}].
\end{aligned}
\label{eq:lambda}
\end{equation}

\subsection{Distillation paradigms}
\label{sec:paradigms}

We institute four distillation paradigms (Fig. \ref{fig:Distillations}). Fig. \ref{fig:Distillations}(a) represents the case where only images from the source domain are distilled. Similarly, Fig. \ref{fig:Distillations}(b) represents the case where only images from the target domain are distilled. Fig. \ref{fig:Distillations}(c) represents source + target distillation, the complete model as proposed in Fig. \ref{fig:KDforSS}. Fig. \ref{fig:Distillations}(d) represents target distillation, where the student network is initialised with weights trained as in case \ref{fig:Distillations}(c). Each of these distillation paradigms contributes to improving performance of the overall architecture in different aspects, and we discuss these effects in detail in section \ref{sec:real2realExp}.

\section{Experimental Setup}
\subsection{Datasets \& Evaluation Protocols}
\vspace{-0.2cm}
We evaluate the proposed algorithm on three popular large-scale autonomous driving datasets. The \textbf{Cityscapes (CS)} \cite{cordts2016cityscapes} dataset and \textbf{Berkeley Deep Drive (BDD)} \cite{yu2018bdd100k} dataset are real scenes captured in Europe and the USA respectively. In particulat, BDD captures varying illumination conditions, seasonal changes. \textbf{GTA5} \cite{richter2016playing} is a popular synthetic driving dataset, and majorly emerged out of computer games. For Cityscapes, we use only the finely annotated images. The class definitions in all these datasets are compatible with each other. 
\begin{table*}[!htbp]
\centering
\begin{center}
\scalebox{0.64}{
\begin{tabular}{c c c c c c c c c c c c c c c c c c c c c c}
\hline
Experiment & mIoU & \rotatebox{90}{Road} & \rotatebox{90}{Sidewalk} & \rotatebox{90}{Building} & \rotatebox{90}{Wall} & \rotatebox{90}{Fence} & \rotatebox{90}{Pole} & \rotatebox{90}{Light} & \rotatebox{90}{Sign} & \rotatebox{90}{Veg} & \rotatebox{90}{Terrain} & \rotatebox{90}{Sky} & \rotatebox{90}{Person} & \rotatebox{90}{Rider} & \rotatebox{90}{Car} & \rotatebox{90}{Truck} & \rotatebox{90}{Bus} & \rotatebox{90}{Train} & \rotatebox{90}{MBike} & \rotatebox{90}{Bike} & mAcc \\ \\
\hline
\multicolumn{22}{c}{(I). Results on the real-to-real scenario: BDD to Cityscapes}\\
\hline
\textit{$\mathcal{T}$}: DRN-D-38 & \textbf{42.33} & 91.53 & 52.72 & 80.65 & 19.21 & 23.4 & 26.31 & 23.75 & 35.79 & 82.48 & 33.85 & 75.36 & 47.87 & 17.65 & 83.15 & 33.77 & 36.74 & 0.08 & 11.02 & 29.01 & \textbf{88.03} \\
\textit{$\mathcal{S}$}: DRN-D-22 & \textbf{38.33} & 89.25 & 49.03 & 78.2 & 16.22 & 20.27 & 21.17 & 17.28 & 33.51 & 80.8 & 28.89 & 71.97 & 45.44 & 12.7 & 76.1 & 21.35 & 32.18 & 0.91 & 6.35 & 26.66 & \textbf{86.12} \\
Distillation (a) & \textbf{42.33} & 92.41 & 53.91 & 80.57 & 19.3 & 22.89 & 26.88 & 23.03 & 36.05 & 82.59 & 34.67 & 77.42 & 48.39 & 15.39 & 82.8 & 27.57 & 40.04 & 0.03 & 9.22 & 31.12 & \textbf{88.42} \\
Distillation (b) & \textbf{43.73} & 92.65 & 56.1 & 81.55 & 17.04 & 26.65 & 26.1 & 25.66 & 39.36 & 83.61 & 37.4 & 76.29 & 50.11 & 14.41 & 83.6 & 33.25 & 41.03 & 0.34 & 10.64 & 35.07 & \textbf{88.89} \\
Distillation (c) & \textbf{43.97} & 92.73 & 56.51 & 81.57 & 16.34 & 26.21 & 27.18 & 25.28 & 38.47 & 83.96 & 37.74 & 79.17 & 50.78 & 15.82 & 83.67 & 33.59 & 40.95 & 0.08 & 10.03 & 35.36 & \textbf{89.09} \\
Distillation (d) & \textbf{44.15} & 92.64 & 55.91 & 81.74 & 16.3 & 26.53 & 26.51 & 25.88 & 38.72 & 83.97 & 37.27 & 78.61 & 50.77 & 16.13 & 83.94 & 36.14 & 39.87 & 0.28 & 11.35 & 36.28 & \textbf{89.08} \\
\hline
\multicolumn{22}{c}{(II). Results on the synthetic-to-real scenario: GTA to Cityscapes}\\
\hline
\textit{$\mathcal{T}$}: DRN-D-38 & \textbf{32.49}	&	87.02 &	40.43 &	73.12 &	17.29 &	10.85 &	21.25 &	19.32 &	10.69 &	71.88 &	26.1 &	58.64 &	41.34 &	9.26 &	75.84 &	15.87 &	21.26 &	10.12 &	5.86 &	1.13 &		\textbf{82.23} \\
\textit{$\mathcal{S}$}: DRN-D-22 & \textbf{30.39} &		86.88 &	38.49 &	69.92 &	13.98 &	12 &	19.73 &	18.05 &	10.09 &	69.09 &	25.39 &	51.06 &	39.64 &	4.93 &	73.5 &	12.71 &	17.36 &	5.6 &	6.59 &	2.44 &		\textbf{80.83} \\
Distillation (a) & \textbf{31.91}	&	87.59 &	39.42 &	72.19 &	15.99 &	11.16 &	20.26 &	20.46 &	11.89 &	71.76 &	26.93 &	60.5 &	41.49 &	4.33 &	76.21 &	14.19 &	18.89 &	5.48 &	6.3 &	1.68 &		\textbf{82.63} \\
Distillation (b) & \textbf{32.98} &		88.76 &	43.5 &	73.34 &	18.56 &	11.32 &	20.39 &	18.2 &	11.5 &	72.67 &	29.45 &	59.47 &	43.35 &	6.65 &	78.53 &	17.24 &	23.71 &	3.48 &	4.97 &	1.5 &		\textbf{83.51} \\
Distillation (c) & \textbf{33.36} &		88.88 &	43.99 &	73.69 &	19.88 &	11.37 &	20.91 &	19.18 &	10.94 &	73.16 &	29.57 &	59.37 &	43.43 &	7.5 &	78.83 &	17.9 &	23.07 &	4.94 &	5.83 &	1.39 &		\textbf{83.72} \\
Distillation (d) & \textbf{33.81} & 89.07 &	44.84 &	74.15 &	20.03 &	11.41 &	22.06 &	19.19 &	11.83 &	73.58 &	29.58 &	59.9 &	43.59 &	7.64 &	79.12 &	18.13 &	22.09 &	8.18 &	6.47 &	1.45 & \textbf{83.9} \\
\hline
\hline
\end{tabular}}
\end{center}
\vspace{-0.7cm}
\caption{Results of our proposed Domain Adaptive Knowledge Distillation method on synthetic-to-real and real-to-real cases scenarios 
}
\label{tab:gta2cs_mainresults}
\end{table*}

We evaluate our models on the validation images of the target domain dataset, using the standard segmentation evaluation metrics: Intersection over Union (\textbf{IoU}) and pixel wise accuracy. The IoU score is calculated as the ratio of intersection and union of the ground truth mask and the predicted mask for each class. We report classwise IoUs for and also compute the \textbf{mIoU} (mean IoU) as the mean of all classwise IoU scores. Additionally, we report the overall \textbf{pixel accuracy} as the ratio of the pixels classified correctly to the number of overall pixels. 
\vspace{-0.2cm}
\subsection{Implementation details}
\vspace{-0.2cm}
We follow the standard procedure for training a GAN and alternate between two steps: i) Fix the discriminator and train the segmentation network with segmentation loss (source domain), domain discrimination loss (target domain) and distillation losses (source and target domains) ii) Fix the generator network and train the discriminator. Details regarding the learning rate, optimizers and GPU configurations are provided in the supplementary material. 


\section{Experimental results}
The performance of our model is evaluated in various distillation paradigms on real-to-real and synthetic-to-real scenarios. Extensive ablation studies are also carried out to determine the performance of various components of our model under various settings.

\subsection{Real to real adaptation: Berkeley Deep Drive to Cityscapes}
\label{sec:real2realExp}
The results of domain-adaptive distillation from a real environment to another real environment viz., BDD to CS are shown in Table \ref{tab:gta2cs_mainresults}(I) (Visual results in Fig. \ref{fig:visualisations_gta2cs}). The first two rows represent the baseline teacher(DRN-D-38, 26.5 M parameters) and student(DRN-D-22, 16.4 M parameters) networks respectively, that are trained using multi-level DA \cite{tsai2018learning}. The results in the subsequent four rows correspond to distillation via the four distillation paradigms (section \ref{sec:paradigms}). The weights for knowledge distillation (in both source and target domains) are set at 0.1, 0.01 and 1.0 for $\lambda_{KL}$ (output level), $\lambda_{MSE}$ (feature level) and $\lambda_{pseudoT}$ (output level) respectively. For KL divergence distillation at the feature level, we set the hyperparameter as one-tenth of its counterpart at the output level.

\vspace{-0.1cm}

\textbf{Analysis of the effects of distilling via the four distillation paradigms ( as per Fig \ref{fig:Distillations}):} As we can observe in Table \ref{tab:gta2cs_mainresults}, source distillation (case (a)) and target distillation (case (b)) outperform the corresponding undistilled student network by 10.43 \% and 14.08 \% (4 and 5.4 mIoU points in terms of absolute numbers) respectively. Target distillation performs better than source distillation. This can be ascribable to the fact that distilling only in the target domain reduces the bias of the model towards the source domain, and improves performance in the target domain. In addition, our network in case (b) (target distillation) outperforms even the teacher network by 3.30 \% (1.4 mIoU points in terms of absolute numbers). Further, in concurrence with our intuition, case (c) (where both source and target distillation are done simultaneously) performs better than both case (a) and case (b), achieving an mIoU of 43.97 (14.71 \% and 3.87 \% relative improvement over the student and teacher networks respectively). Among all the distillation case studies, we observe that case (d) performs the best with a improvement of 15.18 \% and 4.29 \% (5.82 and 1.82 mIoU points respectively) over the student and teacher networks respectively. This is because the model in this case has the best of both worlds - it is initialised with the weights of case (c) thus providing it with the pre-requisite information necessary for source distillation. Finetuning it with target distillation brings target distillation at the focus of the model thus improves the results in the target domain via generalization. We also accredit the improved performance gain of the student network via knowledge distillation to the customized objective functions that we employed in this task, such as cross-entropy loss with pseudo teacher labels $L_{pseudoT}$. We conjecture that the dark knowledge \cite{hinton2015distilling} encapsulated within these pseudo-labels in both source and the target domains substantially augment the individual performance of the compact student network. The fact that our network surpasses even the teacher network by 4.29 \% (1.82 mIoU points in terms of absolute numbers), which is an appreciable number, is commendable given that the teacher network has almost 10 M parameters more than the student network. We notice similar trends in the mAcc scores too.

\subsection{Synthetic to real adaptation: GTA5 to Cityscapes}
\begin{figure*}[!htbp]
    \centering
    \captionsetup[subfigure]{labelformat=empty}
        \begin{subfigure}[b]{0.22\textwidth}
    \includegraphics[scale=0.15]{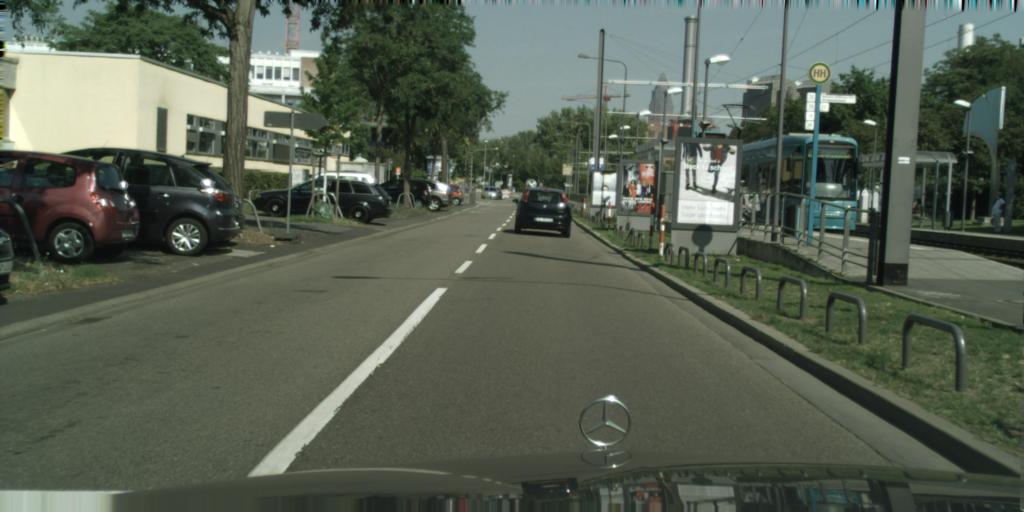}
    \caption{Image}
    \end{subfigure}
    \begin{subfigure}[b]{0.22\textwidth}
    \includegraphics[scale=0.15]{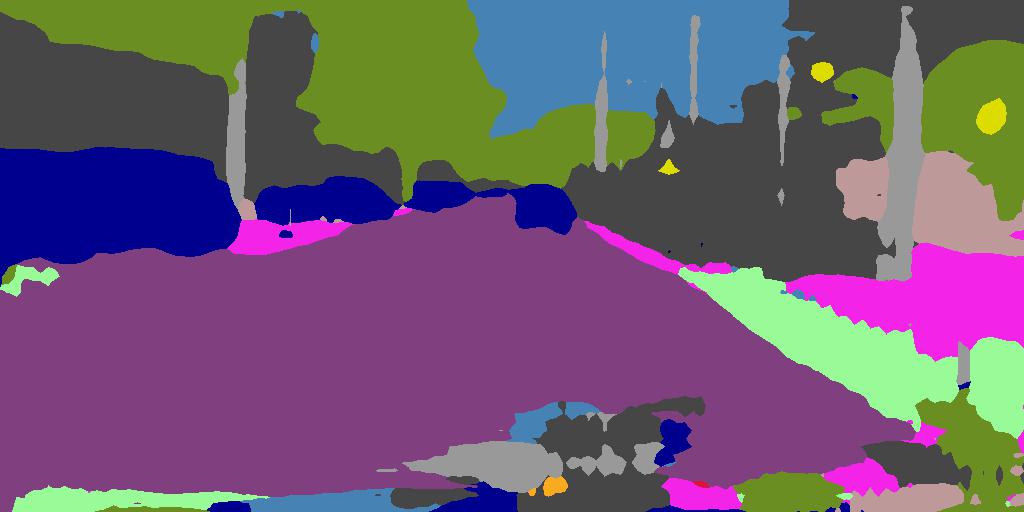}
    \caption{Undistilled Student}
    \end{subfigure}
    \begin{subfigure}[b]{0.22\textwidth}
    \includegraphics[scale=0.15]{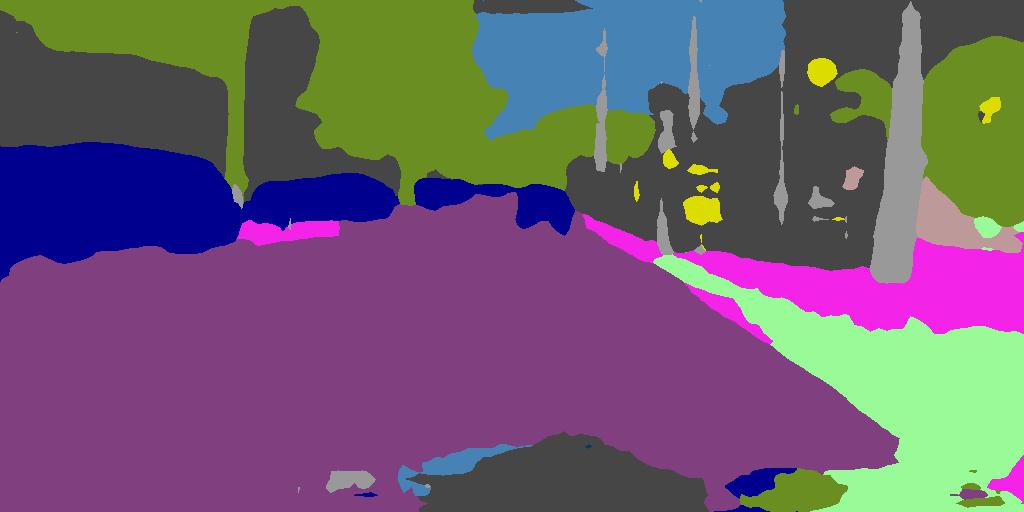}
    \caption{Ours (d)}
    \end{subfigure}
    \begin{subfigure}[b]{0.22\textwidth}
    \includegraphics[scale=0.15]{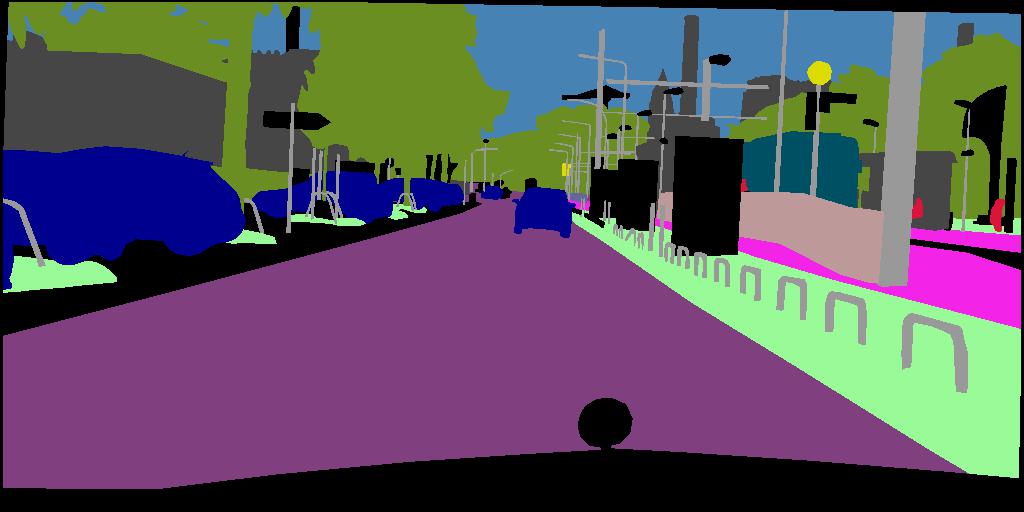}
    \caption{GT}
    \end{subfigure} \\
    
    \begin{subfigure}[b]{0.22\textwidth}
    \includegraphics[scale=0.15]{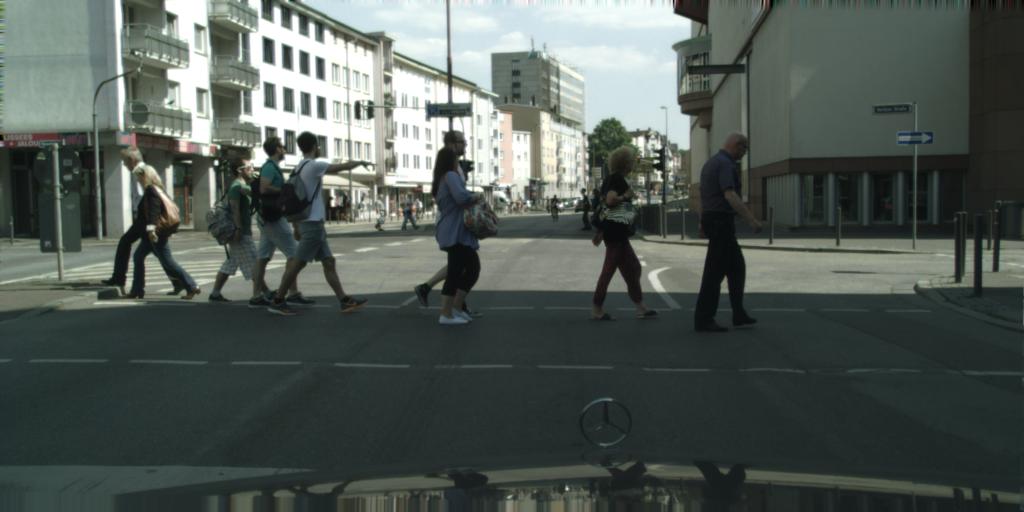}
    \caption{Image}
    \end{subfigure}
    \begin{subfigure}[b]{0.22\textwidth}
    \includegraphics[scale=0.15]{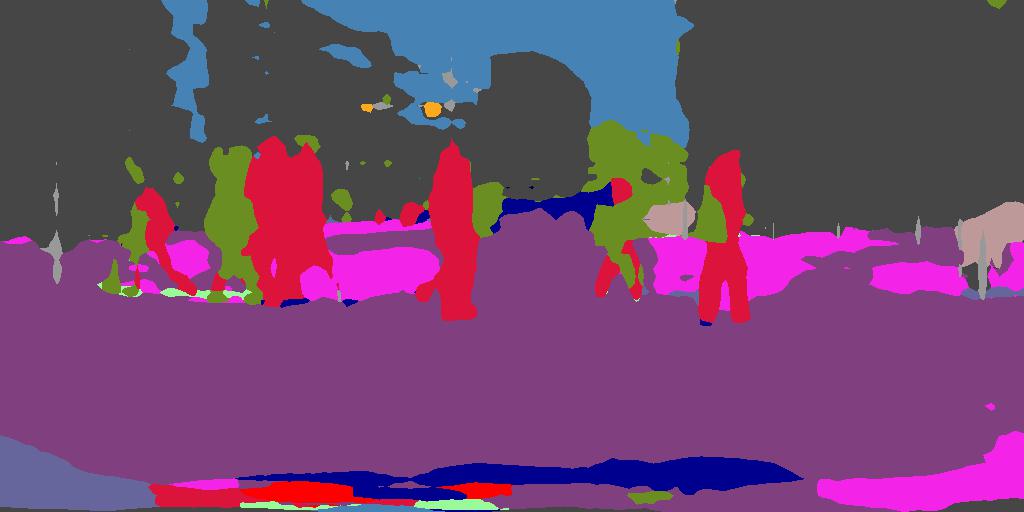}
    \caption{Undistilled Student}
    \end{subfigure}
    \begin{subfigure}[b]{0.22\textwidth}
    \includegraphics[scale=0.15]{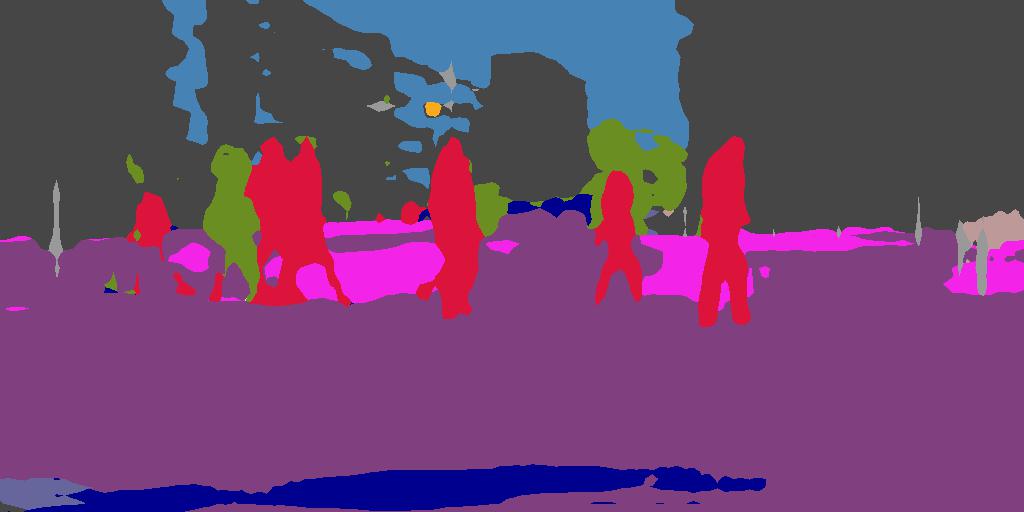}
    \caption{Ours (d)}
    \end{subfigure}
    \begin{subfigure}[b]{0.22\textwidth}
    \includegraphics[scale=0.15]{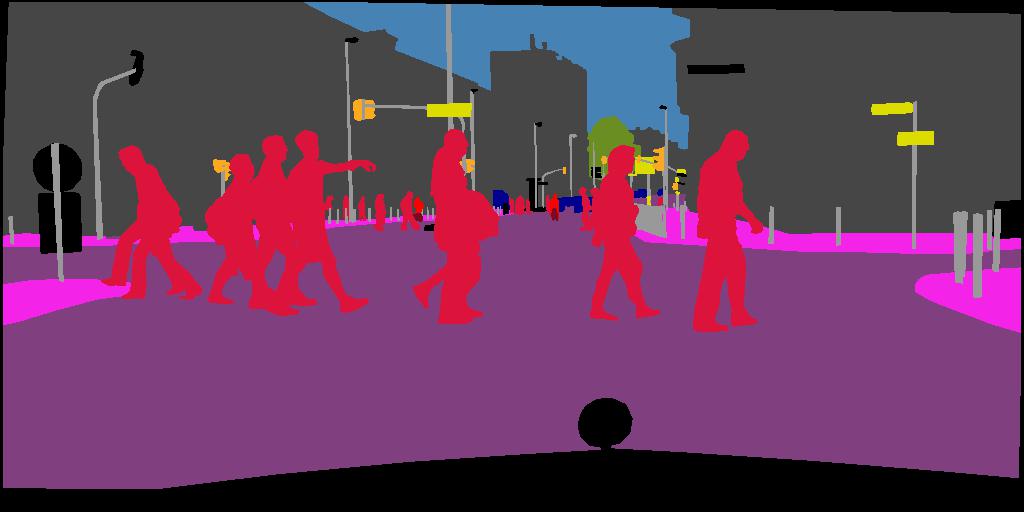}
    \caption{GT}
    \end{subfigure} \\

    \caption{Visual results - First row: BDD to Cityscapes and Second Row: GTA5 to Cityscapes (More results can be found in the supplementary material). Ours (d) corresponds to distillation paradigm (d) as per Fig. \ref{fig:Distillations}}
    \label{fig:visualisations_gta2cs}
        \vspace{-0.6cm}
    
\end{figure*}

Experiments on the synthetic to real dataset pair: GTA5 to Cityscapes with DRN-D-38 teacher (26.5 M parameters) and DRN-D-22 student (16.4 M parameters) are presented in Table \ref{tab:gta2cs_mainresults}(II) (Visual results in Fig. \ref{fig:visualisations_gta2cs}). The trends are analogous to the real to real adaptation case. Source distillation (case (a)) shows an improvement of 5 \% over the undistilled student network (1.52 mIoU points). The other distillation paradigms (case b,c,d) not just outperform the undistilled student network by 8.52 \% - 11.25 \% (1.52 - 3.42 mIoU points in terms of absolute numbers), but also the teacher model by 1.50 \% - 4.06 \% (0.49 - 1.32 mIoU points in terms of absolute numbers). This improvement can be explained on the basis of the knowledge transfer from the teacher to student.

\vspace{-0.1cm}
\noindent \textbf{Effect of change of teacher and student networks:} It is important to study the effects of change of teacher and student networks in knowledge distillation algorithms (Table \ref{tab:ablation}(a)). In addition to the baseline DRN-D-38 teacher \& DRN-D-22 student networks, we use DeepLab (ResNet-101, 44.5 M parameters) as the teacher network and experiment with DRN-D-38 (26.5 M parameters) and DRN-D-22 (16.4 M parameters) as the student networks, with \cite{tsai2018learning} baseline architectures. Our distilled student networks demonstrate an improvement of 13.14 \% - 19.11 \% (4.27 - 5.81 mIoU points in terms of absolute numbers) over the corresponding undistilled student networks, thus showing the effect of knowledge transfer from a powerful and heavy teacher network to different lightweight student networks.

\noindent \textbf{Effect of change of Domain Adaptation backbone:} 
Since our pipeline is immune to the internal domain adaptation strategy used in the teacher and student networks, it can be treated as a blackbox and can be replaced by any other powerful DA approach. Our domain adaptive distillation pipeline will help the student network to benefit accordingly. Table \ref{tab:ablation}(b) shows the effect of using the domain adaptation method, ADVENT \cite{vu2019advent} with DeepLab (ResNet-101, 44.5 M parameters) backbone, as the teacher.  We demonstrate an improvement of 20.63 \% and 14.9 \% (6.27 and 4.86 mIoU points respectively) over the corresponding undistilled DRN-D-22 and DRN-D-38 student networks respectively.\\
\textbf{SOTA comparisons}: Since this paper marks the first approach towards this goal of tackling both domain adaptation and model compression issues in a simultaneous manner, the results shown in this paper set the benchmark for the problem. In addition, in the interest of thorough analysis, we compare the performance of our student model on the synthetic-to-real case against the performance of the current SOTA UDA methods on models with similar sizes, for fair comparison. The results are shown in Table \ref{tab:sota}. With the same backbone architecture (DRN-D-22, first row) and baseline DA strategy, our best model outperforms the popular multi-source DA method AdaptSegNet \cite{tsai2018learning} by 4-6 mIoU points (14\%-20\%) with a model size that is $0.6x$ the model size of its counterpart. By extension, our model scales appropriately when evaluated against all other methods that have similar model sizes and are trained using similar DA strategies. This shows that our method enables our relatively lightweight model achieve SOTA results against comparatively heavier models, which is a testimony to the success of our proposed method. 

\begin{table}[!htb]
\footnotesize
\begin{subtable}{\linewidth}\centering
\begin{tabular}{ccc}
\hline
\hline
Network & mIoU & Params(M) \\
\hline
\textit{$\mathcal{T}$}: DRN-D-38  & 32.49 & 26.5 \\
\textit{$\mathcal{S}$}: DRN-D-22(No dist.) & 30.39 & 16.4 \\
\textit{$\mathcal{S}$}: DRN-D-22( DA dist. (Ours)) &\textbf{ 33.36} & 16.4 \\
\hline
\textit{$\mathcal{T}$}: DeepLab & 42.5 & 44.5 \\
\textit{$\mathcal{S}$}: DRN-D-38 (No dist.) & 32.49 & 26.5 \\
\textit{$\mathcal{S}$}: DRN-D-38 (Ours) & \textbf{36.76} & 26.5 \\
\hline
\textit{$\mathcal{T}$}: DeepLab & 42.5 & 44.5 \\
\textit{$\mathcal{S}$}: DRN-D-22 (No dist.) & 30.39 & 16.4 \\
\textit{$\mathcal{S}$}: DRN-D-22 (Ours) & \textbf{36.2} & 16.4 \\
\hline
\hline
\end{tabular}
\caption{ Varying teacher(\textit{$\mathcal{T}$}) and Student(\textit{$\mathcal{S}$}) networks}
\end{subtable}

\begin{subtable}{\linewidth}\centering
\begin{tabular}{ccc}
\\
\hline
\hline
Network & mIoU & Params(M) \\
\hline
\multicolumn{3}{c}{$\mathcal{T}$ and $\mathcal{S}$ DA:AdaptSegnet \cite{tsai2018learning}}\\
$\mathcal{T}$: DRN-D-38 & 32.49 & 26.5 \\
$\mathcal{S}$: DRN-D-22 (Ours) & \textbf{33.36} & 16.4 \\
\hline
\multicolumn{3}{c}{$\mathcal{T}$ DA:ADVENT \cite{vu2019advent}; $\mathcal{S}$ DA: AdaptSegNet \cite{tsai2018learning}}\\
$\mathcal{T}$: DeepLab & 42.5 & 44.5 \\
\textit{$\mathcal{S}$}: DRN-D-22 (No dist.) & 30.39 & 16.4 \\
$\mathcal{S}$: DRN-D-22 (Ours) & \textbf{36.66} & 16.4 \\
\hline
\multicolumn{3}{c}{$\mathcal{T}$ DA:ADVENT \cite{vu2019advent}; $\mathcal{S}$ DA: AdaptSegNet \cite{tsai2018learning}}\\
$\mathcal{T}$: DeepLab & 42.5 & 44.5 \\
\textit{$\mathcal{S}$}: DRN-D-38 (No dist.) & 32.49 & 26.5 \\
$\mathcal{S}$: DRN-D-38 (Ours) & \textbf{37.35} & 26.5 \\
\hline
\hline
\end{tabular}
\caption{Varying DA backbone}
\end{subtable}
\vspace{-0.2cm}
\caption{Experiments by varying backbone networks}
\label{tab:ablation}
\end{table}

\vspace{-0.5cm}
\begin{table}[!htb]
\centering
\begin{tabular}{cccc}
\hline
\hline
Method & Model & Model Size & mIoU \\ 
\hline
AdaptSegNet \cite{tsai2018learning} & DRN-D-22 & 16.4 & 30.39 \\
AdaptSegNet \cite{tsai2018learning} & DRN-D-38 & 26.5 & 32.49 \\
Ours (\cite{tsai2018learning} backbone) & DRN-D-22 & 16.4 & 36.66 \\
Ours (\cite{tsai2018learning} backbone) & DRN-D-38 & 26.5 & 37.35 \\
\hline
\hline
\end{tabular}  
\caption{State-Of-The-Art (SOTA) comparisons}
\label{tab:sota}
\end{table}

\vspace{-0.4cm}
\subsection{Ablation Studies}
\vspace{-0.2cm}
To deduce the performance of the individual distillation loss functions, we conduct extensive ablation studies on the real-to-real adaptation case. We conduct these studies on distillation paradigm case (a) of Fig. \ref{fig:Distillations}. The rationale is that these results should scale for the other three cases as well.
\vspace{-0.5cm}
\subsubsection{Impact of various loss functions:}
\vspace{-0.1cm}
The impact of various distillation loss functions (which also correspond to studies on feature level distillation vs output level distillation) are shown in Table \ref{tab:sourcelossablation}. For each of the distillation loss terms i.e., $L_{KL}$, $L_{MSE}$ and $L_{pseudoT}$, we conduct ablation studies by varying the corresponding hyper parameters (see Table \ref{tab:sourcelossablation}(i),(ii) and (iii)). The ablation studies on loss term $L_{MSE}$ and $L_{CE-pseudo}$ are correspondent with feature level and output level distillation respectively. $L_{KL}$ is employed according to the MLD strategy. We observe that the loss function $L_{pseudoT}$ performs better than both $L_{KL}$ and $L_{MSE}$, by achieving the best mIoU values of 40.27, 39.86 and 42.18 respectively. We observe that $L_{KL}$ performs its best at a weight of 0.4 at the output level(feature level:0.04). Similarly, $L_{MSE}$ performs its best at 0.01 and $L_{pseudoT}$ performs its best when its weight is 1. For pixels where the label (produced by the teacher network) is faithful, this loss function should mimic the ideal case of classical cross entropy loss function. A high hyperparameter (1.0) pushes the objective function towards this goal.

We also conduct experiments on various permutations of the three distillation loss functions, as per MLD scheme (see Table \ref{tab:sourcelossablation}(iv)-(vii)). We observe that with careful tuning, each of the combinations can achieve better results than the individual loss terms involved. Our best results are obtained (mIoU 42.33) when we use all the three.
\vspace{-0.5cm}
\noindent \subsubsection{Impact of $\lambda_{target}$}
\vspace{-0.28cm}
(Table \ref{tab:sourcelossablation}(viii)), ablation studies 
case (case Fig. \ref{fig:Distillations}(c))): The source distillation parameters are set at 0.1, 0.01 and 1.0 for the three loss terms respectively. For target distillation, we use the same parameters scaled by a factor $\lambda_{target}$ (see equation \ref{eq:lambda}). The best results are obtained at a value of 1.0 (shown in bold). 
This can attributed to the same reason of ideal scenario where 1.0 works best for $\lambda_{pseudoT}$. These teacher labels basically serve as a representative for target domain GT, where our problem is completely unsupervised. 

\begin{table}[!htb]
\centering
\footnotesize
\begin{tabular}{ccc}
\hline
\hline
Parameter($\lambda$) & mIoU & mAcc \\ 
\hline
\multicolumn{3}{c}{(i) $L_{KL}$, Distillation paradigm (a)} \\ 
\hline
0.1 & 39.54 & 86.65 \\
\textbf{0.4} & \textbf{40.27} & \textbf{87.15} \\
0.7 & 40.0 & 86.81 \\
1.0 & 38.94 & 86.18 \\ 
\hline
\multicolumn{3}{c}{(ii) $L_{MSE}$, Distillation paradigm (a)} \\
\hline 
0.005 & 38.91 &	86.25 \\
0.05 & 38.71 &	86.44 \\
\textbf{0.01} & \textbf{39.86} &	\textbf{87.17} \\ 
\hline
\multicolumn{3}{c}{(iii) $L_{pseudoT}$, Distillation paradigm (a)} \\ 
\hline 
0.001 & 40.15 &	87.05 \\
0.01 & 40.6 &	87.22 \\
0.05 & 40.72 &	87.4 \\
0.1 & 41.01	& 	87.59 \\
0.5 & 41.45	& 87.98 \\
\textbf{1.0} & \textbf{42.18} &	\textbf{88.28} \\ 
\hline
\multicolumn{3}{c}{(iv) $L_{KL}$+$L_{MSE}$, Distillation paradigm (a)} \\
\hline 
0.4, 0.01 & 39.89  &		87.15 \\
\textbf{0.1, 0.01} & \textbf{40.32}	 &	\textbf{87.37} \\
0.7, 0.05 & 39.7	 &		87.17 \\ 
\hline
\multicolumn{3}{c}{(v)$L_{KL}+L_{pseudoT}$, Distillation paradigm (a)} \\
\hline 
0.1, 0.1 & 41.18 &		87.98 \\
\textbf{0.1, 1} & \textbf{41.92}  &	\textbf{88.3} \\ 
\hline
\multicolumn{3}{c}{(vi) $L_{MSE}+L_{pseudoT}$, Distillation paradigm (a)} \\
\hline 
\textbf{0.01, 1} & \textbf{42.2}  &	\textbf{88.54} \\ 
\hline
\multicolumn{3}{c}{(vii)$L_{KL}+L_{MSE}+L_{pseudoT}$, Distillation paradigm (a)}\\
\hline 
Case (a): \textbf{0.1, 0.01, 1} & \textbf{42.33} &  \textbf{88.42} \\
\hline
\multicolumn{3}{c}{(viii) $\lambda$\textsubscript{target}, Distillation paradigm (c)}  \\
\hline
0.05 & 43.22 & 		88.76 \\
0.1 & 43.05	& 		88.67 \\
0.5 & 43.64 &			88.9 \\
\textbf{1.0} & \textbf{43.97}  & \textbf{89.09} \\
\hline
\hline
\end{tabular}  
\caption{Ablation studies: Impact of various distillation losses on the real-to-real case, table \ref{tab:gta2cs_mainresults}I; Class-wise performance statistics can be found in the supplementary}
\label{tab:sourcelossablation}
\vspace{-0.5cm}
\end{table}
\vspace{-0.4cm}
\section{Conclusion and Future work}
\vspace{-0.2cm}
We proposed a novel method for distilling domain adaptive segmentation knowledge in models with limited memory, which has important implications in autonomous driving scenarios. To the best of our knowledge, this paper marks the first approach towards this goal. In this regard, we presented a multi-level distillation strategy and proposed a tailor-made multifarious loss function to simultaneously handle two key issues in segmentation models, domain gap and memory constraints. We instituted various distillation paradigms. Our experiments and extensive analysis show the success of our proposed method. 

{\small
\bibliographystyle{ieee_fullname}
\bibliography{egbib}
}

\end{document}


\title{Supplementary Material for Domain Adaptive Knowledge Distillation for Unsupervised Semantic Segmentation}
\author{Divya Kothandaraman
\qquad
Athira Nambiar
\qquad
Anurag Mittal \\
{\tt\small ramandivya27@yahoo.in, \{anambiar,amittal\}@cse.iitm.ac.in}\\
Indian Institute of Technology Madras, India
}

\maketitle

\section{Implementation details}
The segmentation network is optimized with Stochastic Gradient Descent (SGD) optimizer (with Nesterov acceleration), where the weight decay is 1e-4 and the momentum is 0.9. The initial learning rate is set at 2.5e-4 and is decreased with a polynomial decay of 0.9. To train the discriminator, we use the Adam optimizer with a learning rate of 10e-4. The polynomial decay is the same as that of the segmentation network. All our experiments are performed on a single NVIDIA GEForce 11 GB GPU, with a batch size of 2.

\section{Quantitative results: Ablation studies}

Table \ref{tab:sourcelossablation} shows class-wise performance for the ablation experiments conducted on the various distillation losses proposed. The ablation corresponds to the source distillation paradigm (paradigm a in the distillation paradigms figure), and the experiments have been conducted on the real-to-real case. 
To deduce the performance of the individual distillation loss functions, we conduct extensive ablation studies on the real-to-real adaptation case. We conduct these studies on distillation paradigm case (a). The rationale is that these results should scale for the other three cases as well. 

\subsubsection{Impact of various loss functions:} The impact of various loss functions has been discussed in the paper.
\\

\noindent \textbf{Class-wise performance:} As with most segmentation models, we notice that our domain-adaptive distilled model performs particularly well on classes such as road, car, vegetation, sky, etc. which have a huge presence in the dataset. Rare classes such as trains have a high probability of being confused with bus; truck and bus can be confusing to differentiate - these can in fact be wrongly classified as cars; wall and fence can be ambiguous and so on. We also notice that detection of small objects like traffic signs and persons in some images gets missed out. This can be attributed to multiple reasons - size of the object, rare occurrence and nuanced boundaries. While our proposed model outperforms both the teacher and the student in most categories, the trends of these models across categories are very similar. Thus, we believe that these issues are innate to the baseline domain adaptation and segmentation models.

\begin{table*}[!ht]
\centering
\begin{center}
\caption{Ablation studies: Impact of various distillation losses for source distillation on the real-to-real case.}
\label{tab:sourcelossablation}
\scalebox{0.615}{
\begin{tabular}{c c c c c c c c c c c c c c c c c c c c c c}
\hline
Parameter($\lambda$) & mIoU & \rotatebox{90}{Road} & \rotatebox{90}{Sidewalk} & \rotatebox{90}{Building} & \rotatebox{90}{Wall} & \rotatebox{90}{Fence} & \rotatebox{90}{Pole} & \rotatebox{90}{Light} & \rotatebox{90}{Sign} & \rotatebox{90}{Veg} & \rotatebox{90}{Terrain} & \rotatebox{90}{Sky} & \rotatebox{90}{Person} & \rotatebox{90}{Rider} & \rotatebox{90}{Car} & \rotatebox{90}{Truck} & \rotatebox{90}{Bus} & \rotatebox{90}{Train} & \rotatebox{90}{MBike} & \rotatebox{90}{Bike} & mAcc \\ \\
\hline
\multicolumn{22}{c}{(i) KL divergence ($L_{KL}$)} \\ 
\hline 
0.1 & 39.54 & 90.08 & 50.58 & 78.99 & 16.41 & 20.71 & 24.43 & 19.44 & 35.11 & 80.65 & 28.63 & 73.07 & 47 & 15.35 & 78.32 & 23.19 & 33.17 & 0.45 & 7.98 & 27.78 & 86.65 \\
\textbf{0.4} & \textbf{40.27} & 90.63 & 51.2 & 79.53 & 18.11 & 21.67 & 24.96 & 20.75 & 35.54 & 81.29 & 29.08 & 74.86 & 47.71 & 14.81 & 79.65 & 23.94 & 34.39 & 0.46 & 8.32 & 28.28 & \textbf{87.15} \\
0.7 & 40.0 & 90.09 & 50.44 & 78.96 & 18.74 & 21.31 & 23.97 & 20.19 & 35.48 & 81.29 & 29.6 & 74.2 & 46.97 & 14.83 & 78.68 & 23.23 & 33.97 & 1.5 & 7.61 & 29.03 & 86.81 \\
1.0 & 38.94 & 89.43 & 50.27 & 78.39 & 17.1 &	20.35 &	23.59 & 	18.23 &	34.28 &	80.45 &	28.48 &	71.35 &	46.52 &	14.66 &	77.28 &	21.06 &	32.46 &	0.93 &	6.21 &	28.89 & 86.18 \\ \\
\hline
\multicolumn{22}{c}{(ii) MSE loss ($L_{MSE}$)} \\
\hline 
0.005 & 38.91 & 89.61 &	50.04 &	78.39 &	17.55 &	20.4 &	22.99 &	17.63 &	33.96 &	80.45 &	28.54 &	71.88 &	46.14 &	14.28 &	77.73 &	20.86 &	32.83 &	1.01 &	6.91 &	28.04 &	86.25 \\
0.05 & 38.71 & 89.94 & 50.56 & 78.55 & 16.34 & 	20.8 & 	21.38 &	17.39 &	33.77 &	80.18 &	28.66 &	72.73 &	45.98 &	12.06 &	78.51 &	21.37 &	33.48 &	1.16 &	5.45 &	27.27 &	86.44 \\
\textbf{0.01} & \textbf{39.86} & 90.92 & 50.78 &	79.45 &	19.59 &	21.23 &	23.55 &	18.82 &	36.34 &	81.06 &	29.12 &	75.31 &	47.05 &	12.5 &	79.99 &	23.05 &	33.07 &	0.86 &	7.48 &	27.21 &	\textbf{87.17} \\ \\
\hline
\multicolumn{22}{c}{(iii) Cross entropy quasi teacher labels ($L_{CE-quasiT}$)} \\ 
\hline 
0.001 & 40.15 & 90.43 &	51.19 & 79.45 &	17.11 &	21.58 &	24.47 &	19.81 &	35.36 & 81.25 &	29.63 &	74.67 &	47.68 &	13.98 &	79.37 &	24.8 &	34.19 &	0.6 &	8.77 &	28.34 &	87.05 \\
0.01 & 40.6 &	90.81 &	52.09 &	79.75 &	16.97 &	22.64 &	25.68 &	20.9 &	35.54 &	81.07 &	29.14 &	74.45 &	48.48 &	17.58 &	80.1 &	24.17 &	33.36 &	0.66 &	10.22 &	27.79 &		87.22 \\
0.05 & 40.72	&	91.17 &	51.15 &	79.65 &	18.25 &	22.5 &	25.54 &	21.42 &	35.81 &	81.58 &	30.55 &	75.66 &	48.05 &	15.54 &	80.05 &	24.28 &	34.46 &	0.44 &	8.99 &	28.57 &	87.4 \\
0.1 & 41.01	& 91.48 &	51.91 &	79.82 &	18.32 &	22.75 &	25.86	& 21.54 &	35.69 &	81.7 &	30.47 &	76.21 &	47.68 &	16.3 &	80.28 &	25.17 &	34.93 &	0.49 &	9.14 &	29.39 &	87.59 \\
0.5 & 41.45	& 91.98 &	52.62 &	80 &	18.58 &	23.55 &	25.72 &	21.67 &	36.44 &	82.29 &	32.84 &	76.98 &	47.8 &	14.87 &	81.08 &	25.76 &	36.33 &	0.29 &	8.6 &	30.2 &		87.98 \\
\textbf{1.0} & \textbf{42.18}	& 92.27 &	55.57 &	80.26 &	19.2 &	24.6 &	25.66 &	21.98 &	36.03 &	82.88 &	34.75 &	77.52 &	48.42 & 17.72 &	82.06 &	24.57 &	37.29 &	0.14 &	9.32 &	31.2 &	\textbf{88.28} \\ \\
\hline
\multicolumn{22}{c}{(iv) Combination of the loss terms KL, MSE (L$_{KL}$+L$_{MSE}$)} \\
\hline 
0.4, 0.01 & 39.89 &	90.83 &	50.88 &	79.42 &	17.11 &	21.29 &	24.66 &	20.2 &	34.78 &	81.03 &	30.08 &	74.75 &	47.23 &	14.3 &	79.85 &	23.15 &	32.26 &	0.39 &	8.03 &	27.65 &		87.15 \\
\textbf{0.1, 0.01} & \textbf{40.32}	& 91.04 &	51.11 &	79.82 &	17.35 &	21.55 &	25.57 &	20.58 &	34.46 &	81.26 &	29.92 &	75.67 &	47.59 &	15.76 &	80.55 &	24.04 &	33.79 &	0.3 &	8.63 &	27.1 &	\textbf{87.37} \\
0.7, 0.05 & 39.7	&	91.04 &	50.92 &	79.37 &	16.83 &	21.24 &	24.19 &	20.51 &	33.92 &	80.8 &	30.43 &	75.02 &	46.48 &	13.71 &	80.13 &	22.73 &	31.51 &	0.56 &	7.17 &	27.76 &		87.17 \\ \\
\hline
\multicolumn{22}{c}{(v)Combination of the loss terms KL, CE-quasiT ($L_{KL}+L_{CE-quasiT}$)} \\
\hline 
0.1, 0.1 & 41.18	&	92.13 &	52.6 &	80.18 &	18.8 &	22.89 &	26.28 &	22.18 &	35.35 &	81.85 &	32.11 &	77.04 &	48 &	15.02 &	81.54 & 25.38 &	33.85 &	0.3 &	7.53 &	29.4 &		87.98 \\
\textbf{0.1, 1} & \textbf{41.92} & 92.4 &	53.8 &	80.44 &	18.62 &	24.52 &	25.61 &	22.17 &	36.95 &	82.63 &	33.45 &	78.14 &	48.18 &	14.92 &	81.72 &	26.63 &	35.9 &	0.09 &	8.95 &	31.43 &	\textbf{88.3} \\ \\
\hline
\multicolumn{22}{c}{(vi) Combination of the loss terms MSE, CE-quasiT ($L_{MSE}+L_{CE-quasiT}$)} \\
\hline 
\textbf{0.01, 1} & \textbf{42.2} & 92.62 &	54.64 &	80.71 &	18.19 &	23.17 &	26.04 &	22.63 &	35.32 &	82.98 &	34.49 &	78.13 &	48.35 &	16.19 &	82.93 &	27.17 &	37.28 &	0.04 &	8.72 &	32.17 &	\textbf{88.54} \\ \\
\hline
\multicolumn{22}{c}{(vii)Combination ($L_{KL}+L_{MSE}+L_{CE-quasiT}$); $\lambda_{KL}$ =0.1,  $\lambda_{MSE}$ = 0.01, $\lambda_{CE-quasiT}$ = 1.0} \\
\hline 
Case (a) & \textbf{42.33} & 92.41 & 53.91 & 80.57 & 19.3 & 22.89 & 26.88 & 23.03 & 36.05 & 82.59 & 34.67 & 77.42 & 48.39 & 15.39 & 82.8 & 27.57 & 40.04 & 0.03 & 9.22 & 31.12 & \textbf{88.42} \\
\hline
\end{tabular}}
\end{center}
\vspace{-0.7cm}
\end{table*}

\section{Qualitative Results}

In this section, we present visual results for our proposed pipeline 'domain-adaptive distillation'. The evaluation is done on the target domain of the student network. The nomenclature is as follows:\\
\begin{itemize}
\item Image: Target domain input image on which evaluation is done
\item GT: Corresponding ground truth
\item Teacher: Teacher network output for the target domain image
\item Student: Student network output for the target domain image
\item Source distillation (a): Output of student network distilled as per distillation paradigm (a) (Source domain distillation), evaluated on the target domain image
(All distillations are as per Fig. 2 in the paper)
\item Target distillation (b): Output of student network distilled as per distillation paradigm (b) (Target domain distillation), evaluated on the target domain image
\item Source + target distillation (c): Output of student network distilled as per distillation paradigm (c) (Source + target domain distillation), evaluated on the target domain image
\item Target init distillation (d): Output of student network distilled as per distillation paradigm (d) (Target domain distillation, initialised with case (c)), evaluated on the target domain image
\end{itemize}
\subsection{BDD to cityscapes}
This section has visual results for the real-to-real adaptation case: Berkeley Deep Drive to Cityscapes. (Fig. 1)
\begin{figure*}
    \centering
    \captionsetup[subfigure]{labelformat=empty}
    \begin{subfigure}[b]{0.22\textwidth}
    \includegraphics[scale=0.1]{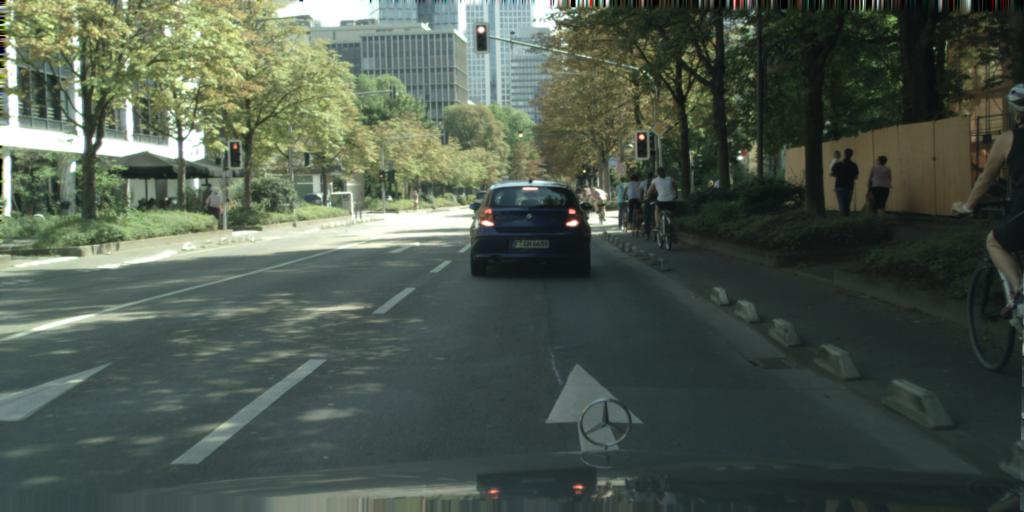}
    \caption{Image 1}
    \end{subfigure}
    \begin{subfigure}[b]{0.22\textwidth}
    \includegraphics[scale=0.1]{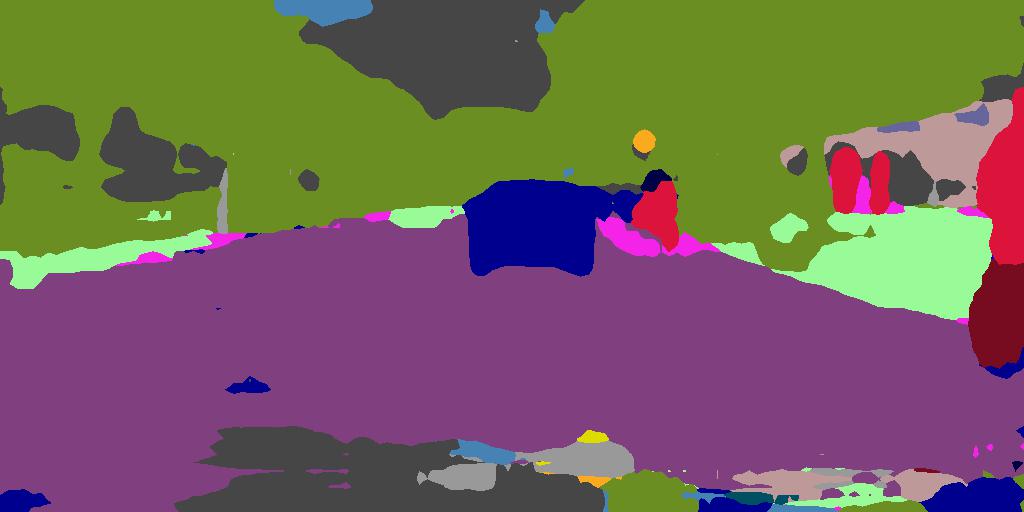}
    \caption{Student}
    \end{subfigure}
    \begin{subfigure}[b]{0.22\textwidth}
    \includegraphics[scale=0.1]{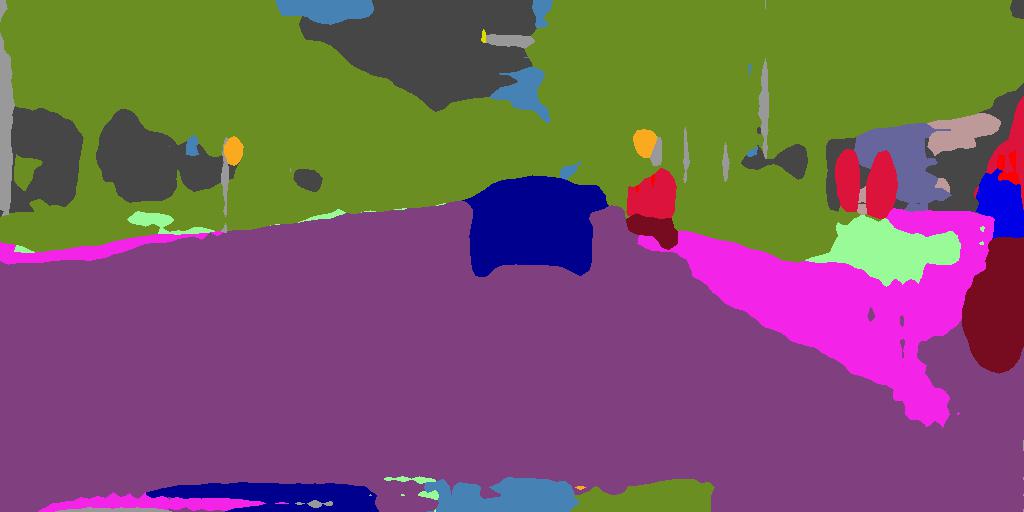}
    \caption{Teacher}
    \end{subfigure}
    \begin{subfigure}[b]{0.22\textwidth}
    \includegraphics[scale=0.1]{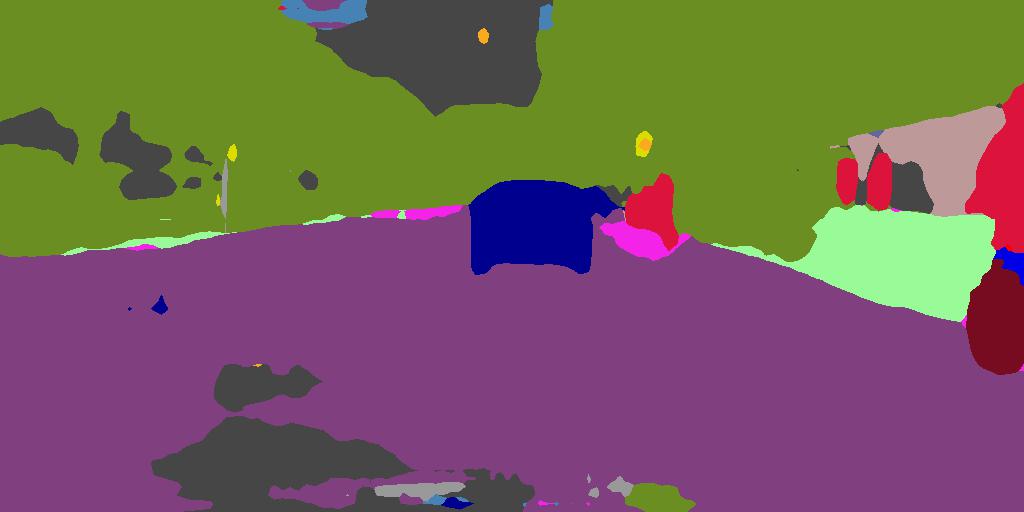}
    \caption{Source dist. (a)}
    \end{subfigure}\\
    \begin{subfigure}[b]{0.22\textwidth}
    \includegraphics[scale=0.1]{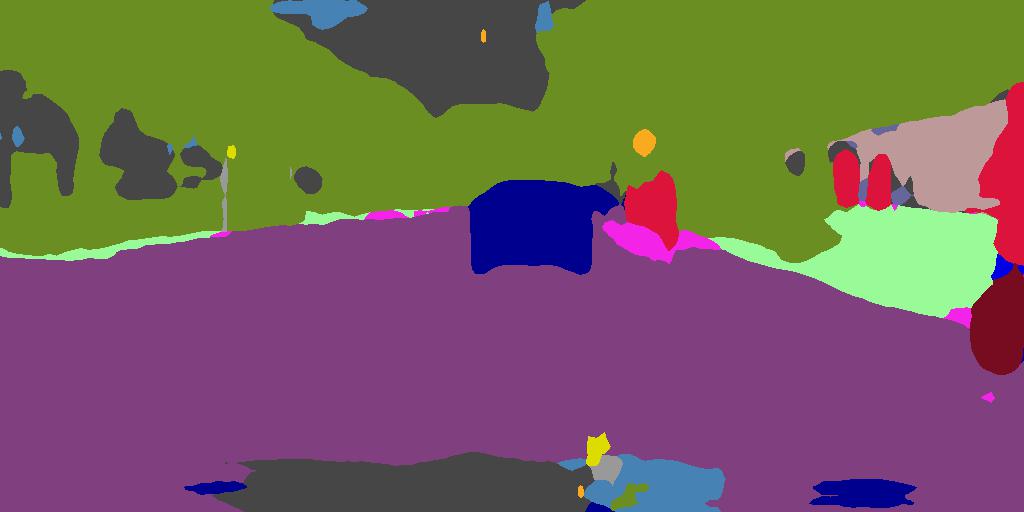}
    \caption{Target dist. (b)}
    \end{subfigure}
    \begin{subfigure}[b]{0.22\textwidth}
    \includegraphics[scale=0.1]{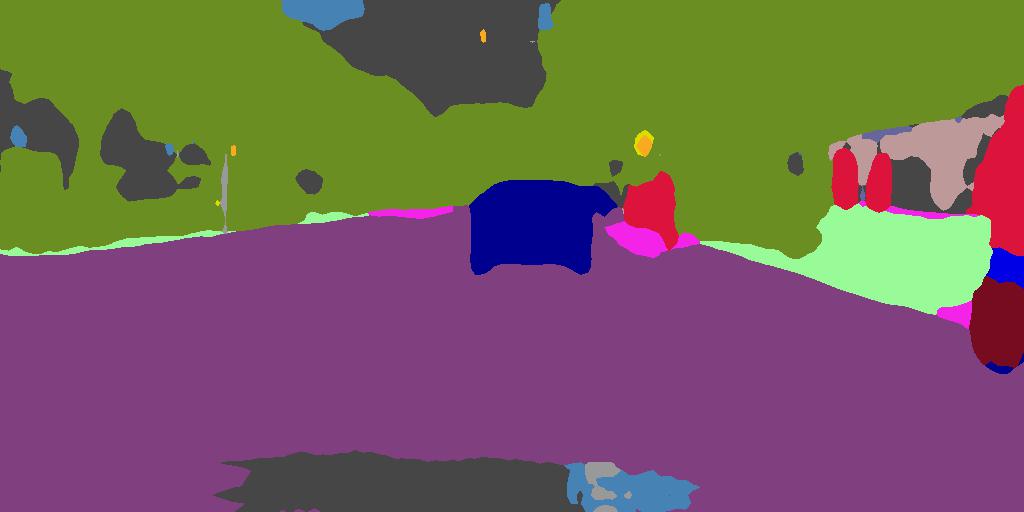}
    \caption{Src + Tgt dist. (c)}
    \end{subfigure}
    \begin{subfigure}[b]{0.22\textwidth}
    \includegraphics[scale=0.1]{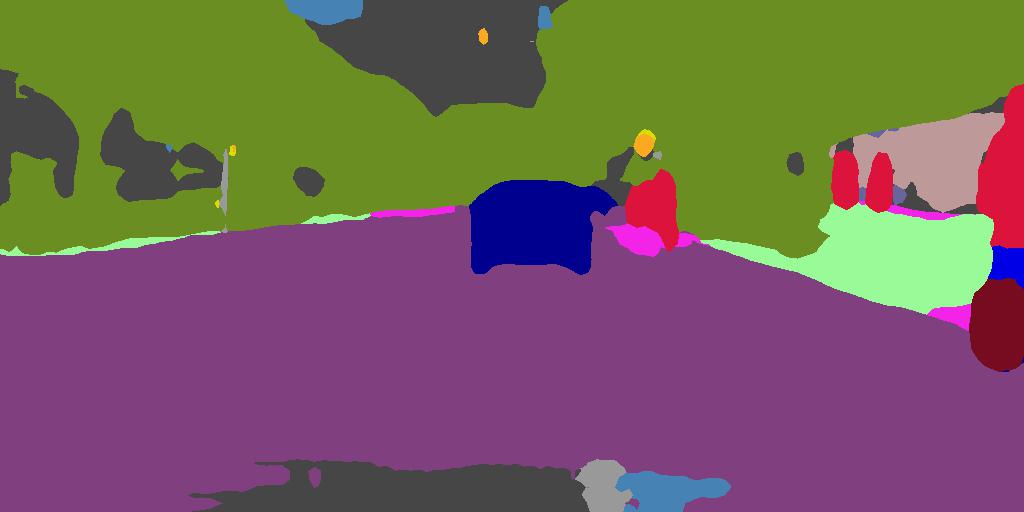}
    \caption{Target init.dist.(d)}
    \end{subfigure}
    \begin{subfigure}[b]{0.22\textwidth}
    \includegraphics[scale=0.1]{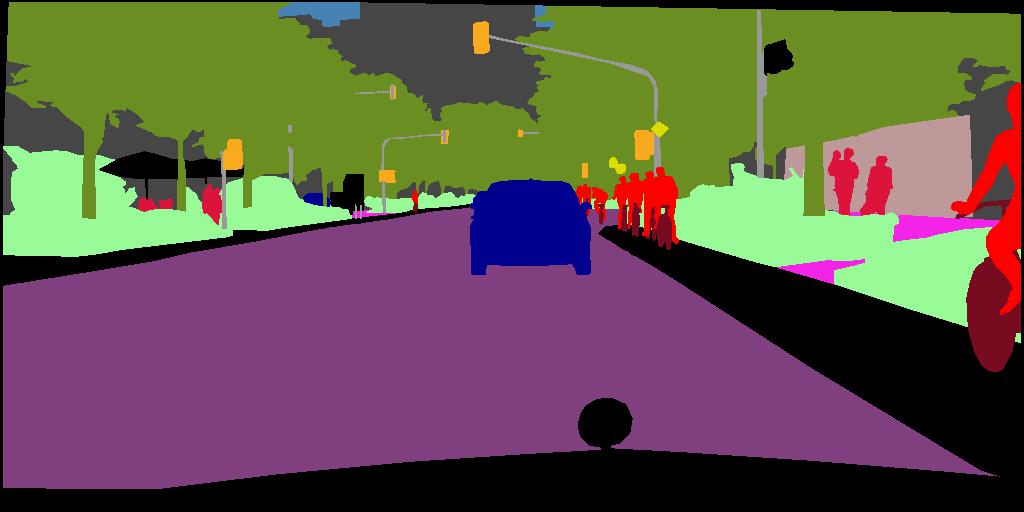}
    \caption{GT}
    \end{subfigure} \\ \\
    
    \begin{subfigure}[b]{0.22\textwidth}
    \includegraphics[scale=0.1]{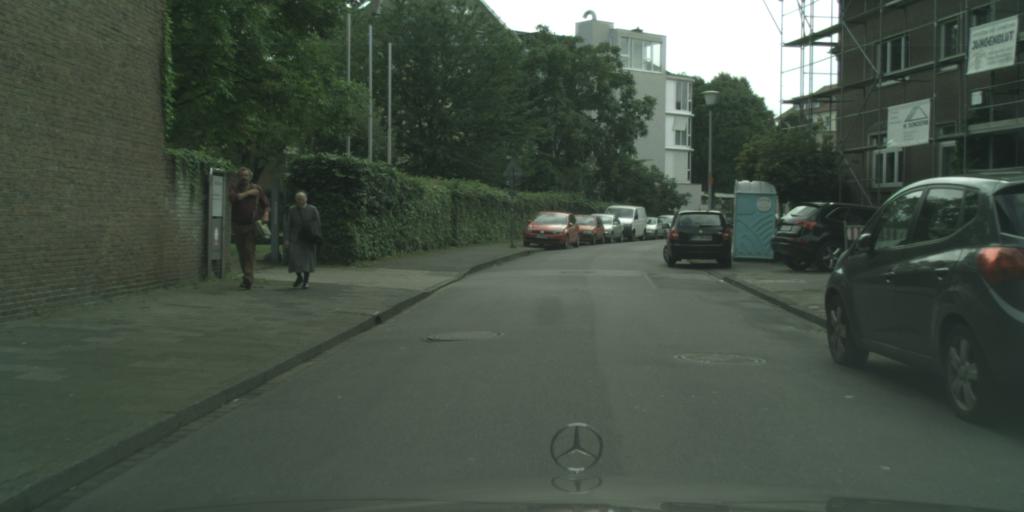}
    \caption{Image 2}
    \end{subfigure}
    \begin{subfigure}[b]{0.22\textwidth}
    \includegraphics[scale=0.1]{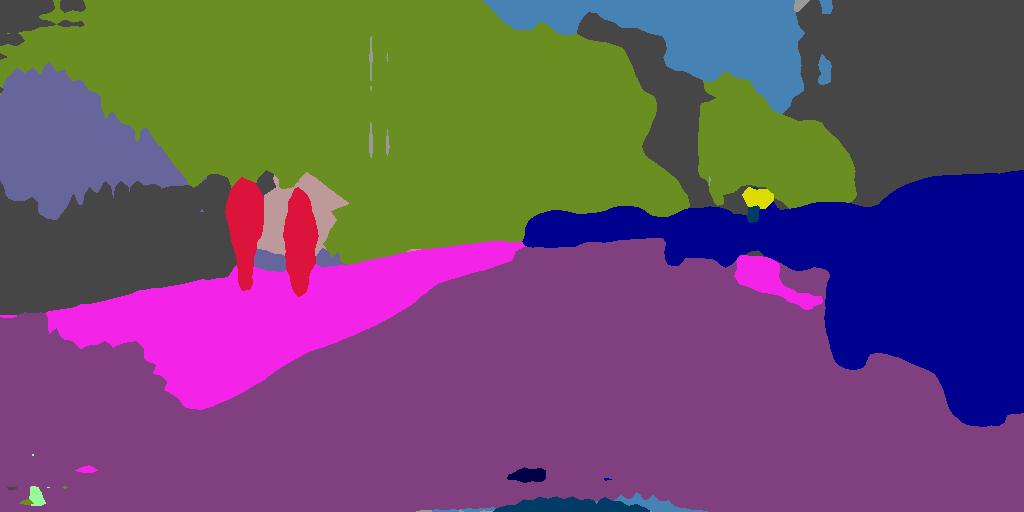}
    \caption{Student}
    \end{subfigure}
    \begin{subfigure}[b]{0.22\textwidth}
    \includegraphics[scale=0.1]{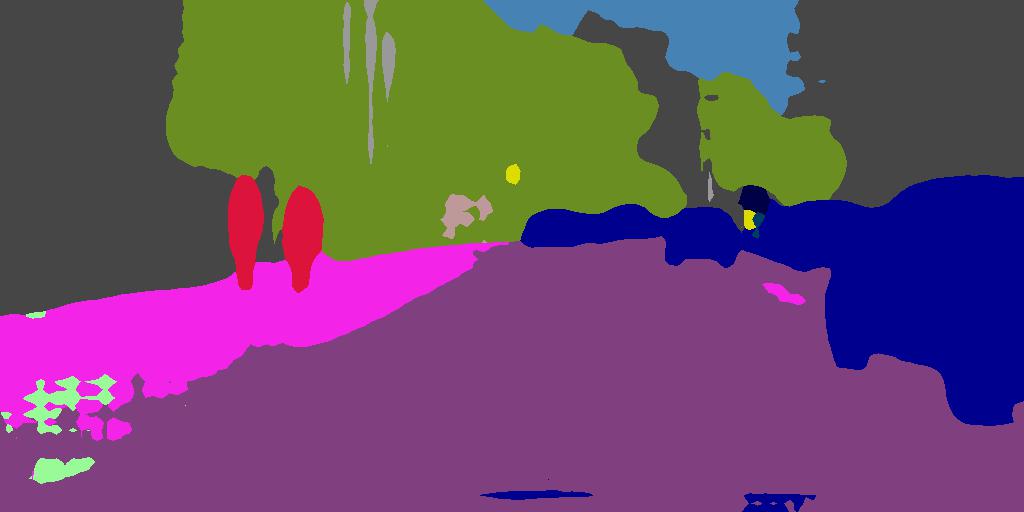}
    \caption{Teacher}
    \end{subfigure}
    \begin{subfigure}[b]{0.22\textwidth}
    \includegraphics[scale=0.1]{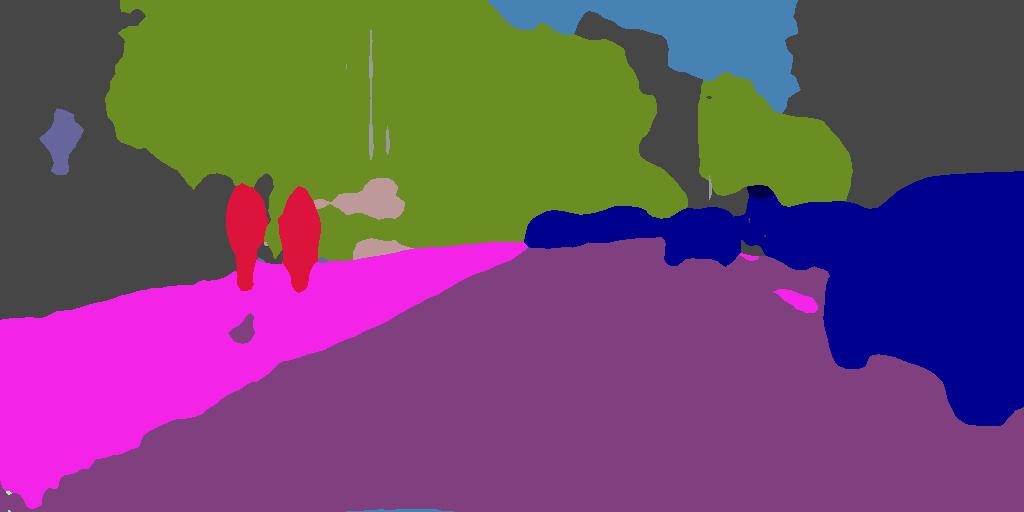}
    \caption{Source dist. (a)}
    \end{subfigure}\\
    \begin{subfigure}[b]{0.22\textwidth}
    \includegraphics[scale=0.1]{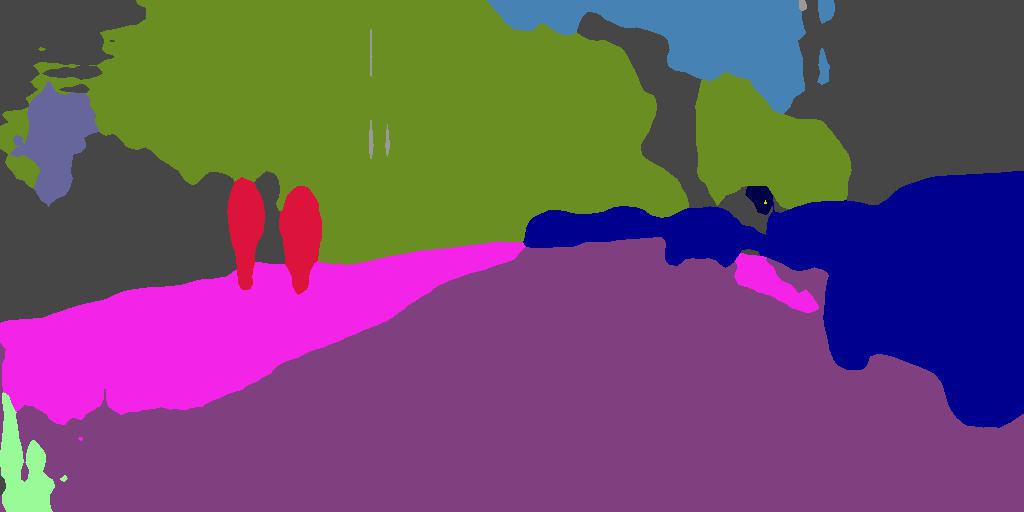}
    \caption{Target dist. (b)}
    \end{subfigure}
    \begin{subfigure}[b]{0.22\textwidth}
    \includegraphics[scale=0.1]{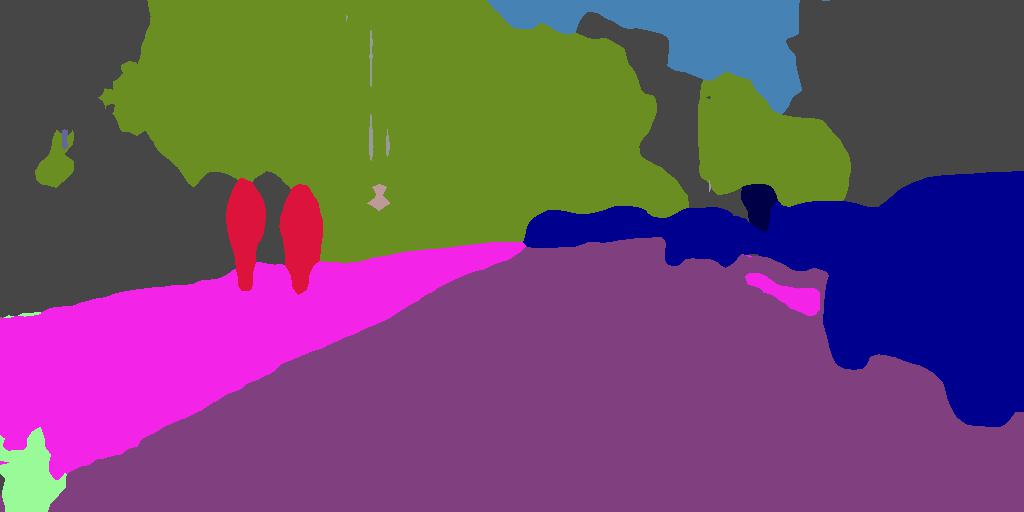}
    \caption{Src + Tgt dist. (c)}
    \end{subfigure}
    \begin{subfigure}[b]{0.22\textwidth}
    \includegraphics[scale=0.1]{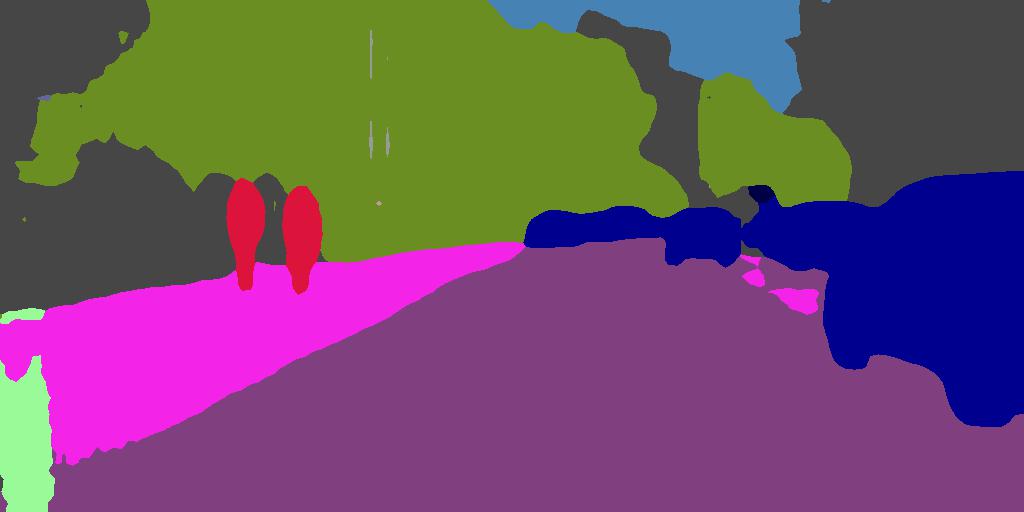}
    \caption{Target init.dist.(d)}
    \end{subfigure}
    \begin{subfigure}[b]{0.22\textwidth}
    \includegraphics[scale=0.1]{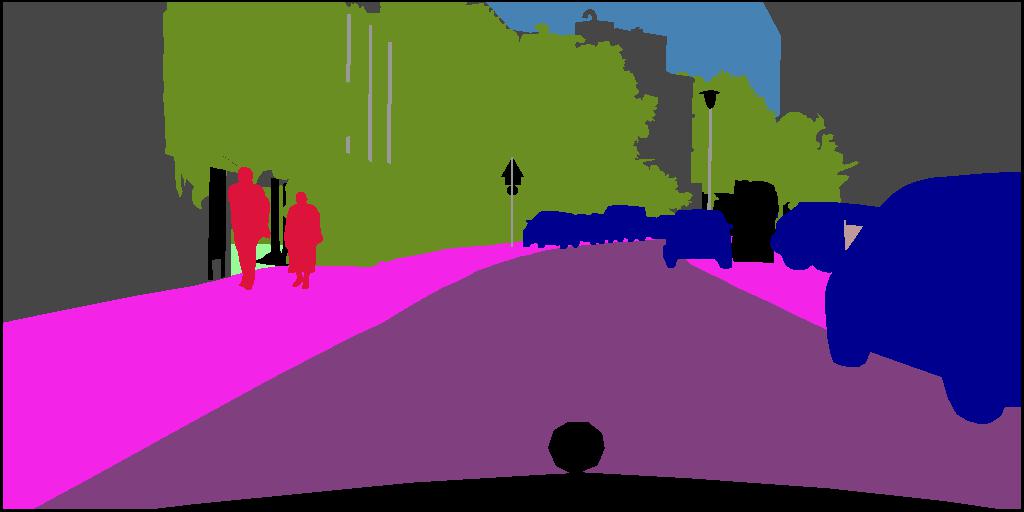}
    \caption{GT}
    \end{subfigure} \\ \\

    \begin{subfigure}[b]{0.22\textwidth}
    \includegraphics[scale=0.1]{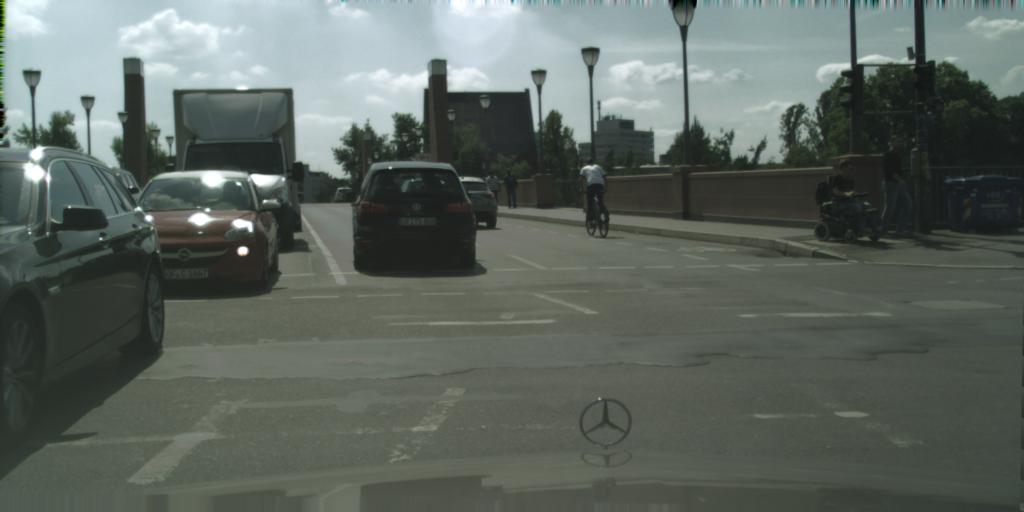}
    \caption{Image 3}
    \end{subfigure}
    \begin{subfigure}[b]{0.22\textwidth}
    \includegraphics[scale=0.1]{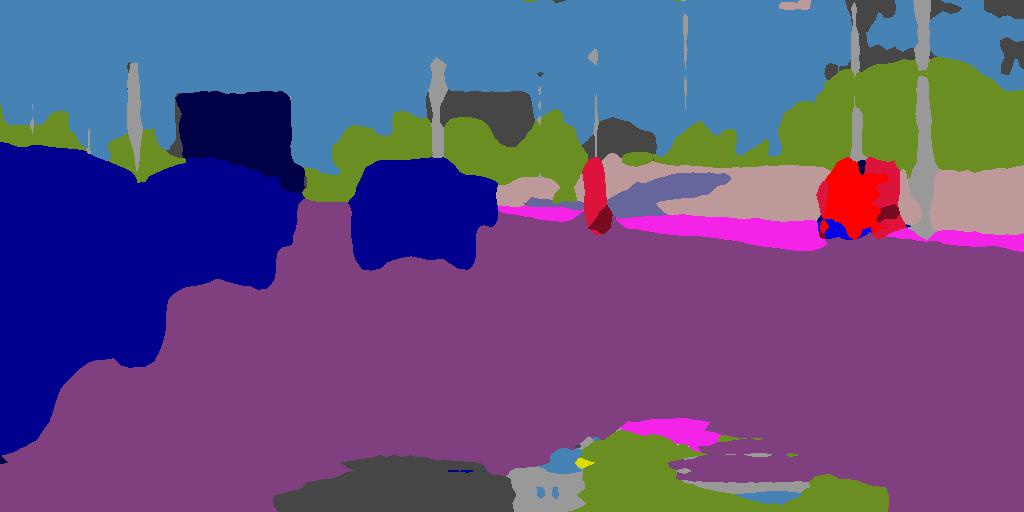}
    \caption{Student}
    \end{subfigure}
    \begin{subfigure}[b]{0.22\textwidth}
    \includegraphics[scale=0.1]{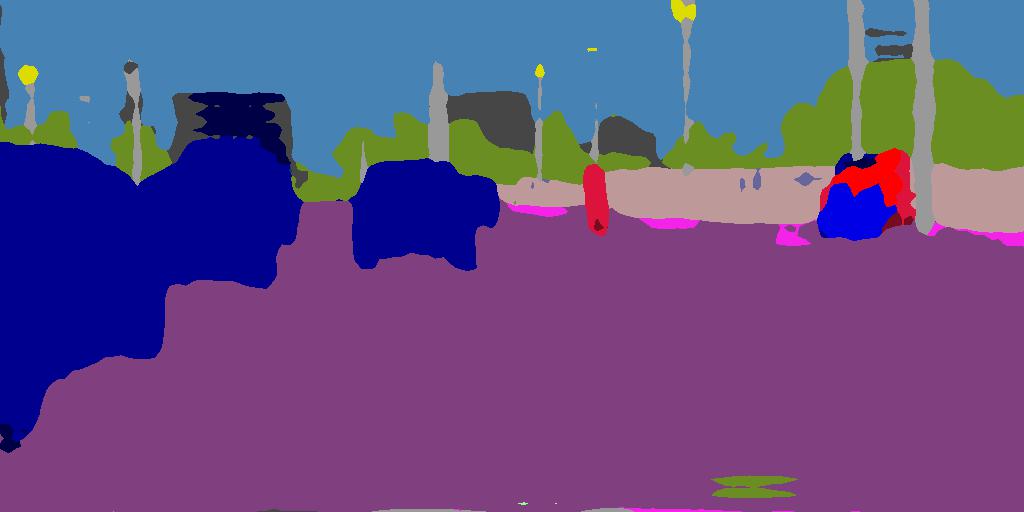}
    \caption{Teacher}
    \end{subfigure}
    \begin{subfigure}[b]{0.22\textwidth}
    \includegraphics[scale=0.1]{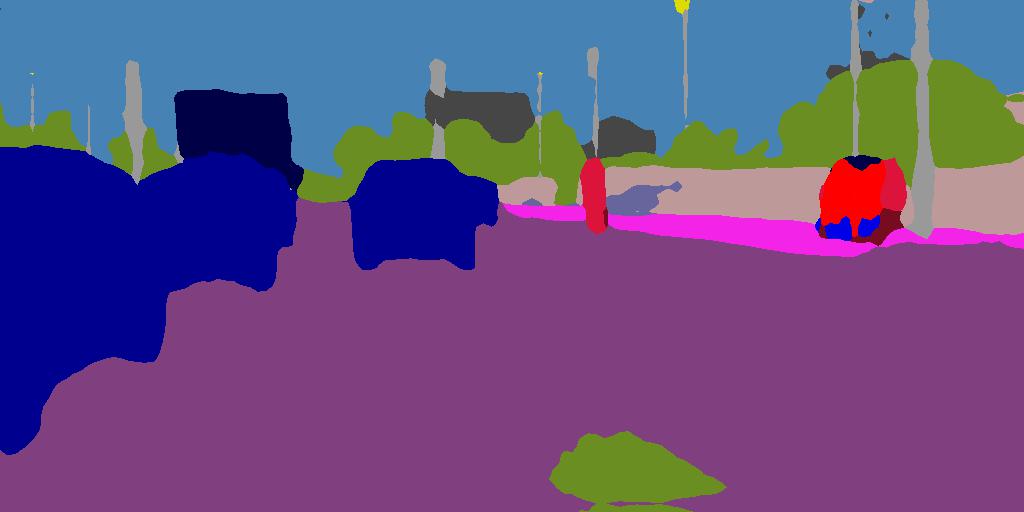}
    \caption{Source dist. (a)}
    \end{subfigure}\\
    \begin{subfigure}[b]{0.22\textwidth}
    \includegraphics[scale=0.1]{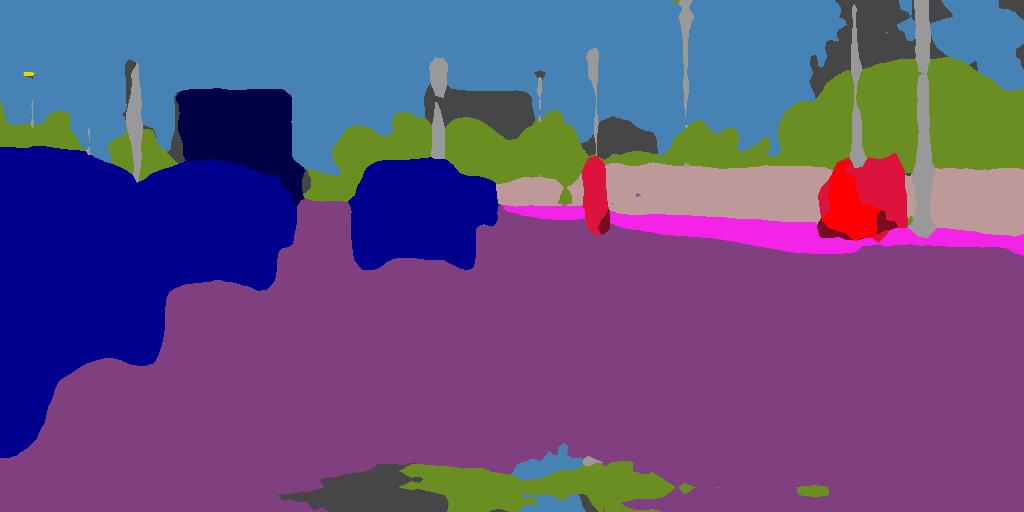}
    \caption{Target dist. (b)}
    \end{subfigure}
    \begin{subfigure}[b]{0.22\textwidth}
    \includegraphics[scale=0.1]{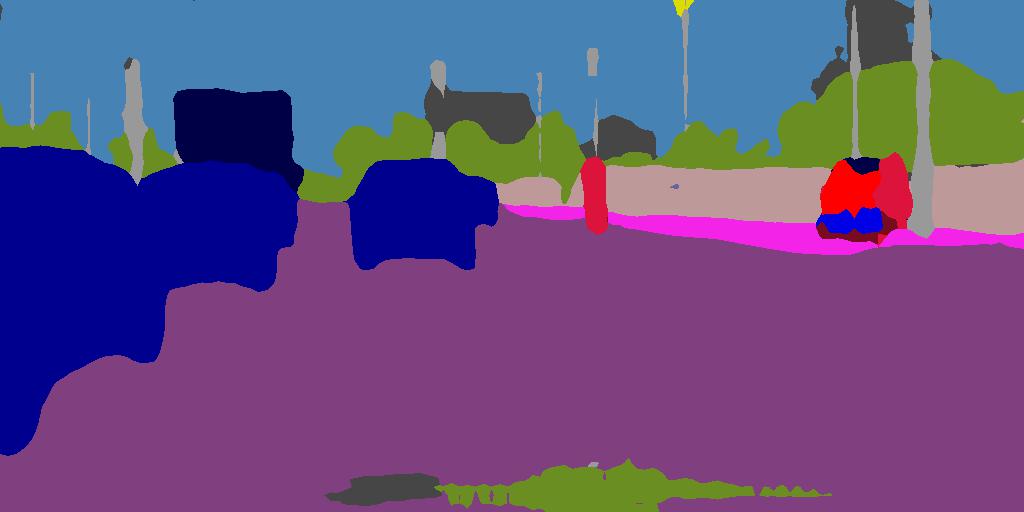}
    \caption{Src + Tgt dist. (c)}
    \end{subfigure}
    \begin{subfigure}[b]{0.22\textwidth}
    \includegraphics[scale=0.1]{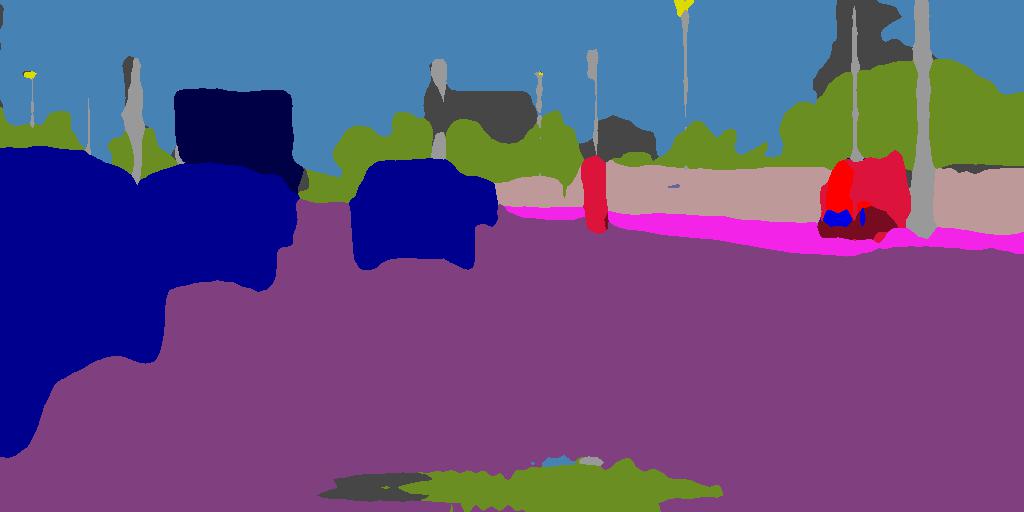}
    \caption{Target init.dist.(d)}
    \end{subfigure}
    \begin{subfigure}[b]{0.22\textwidth}
    \includegraphics[scale=0.1]{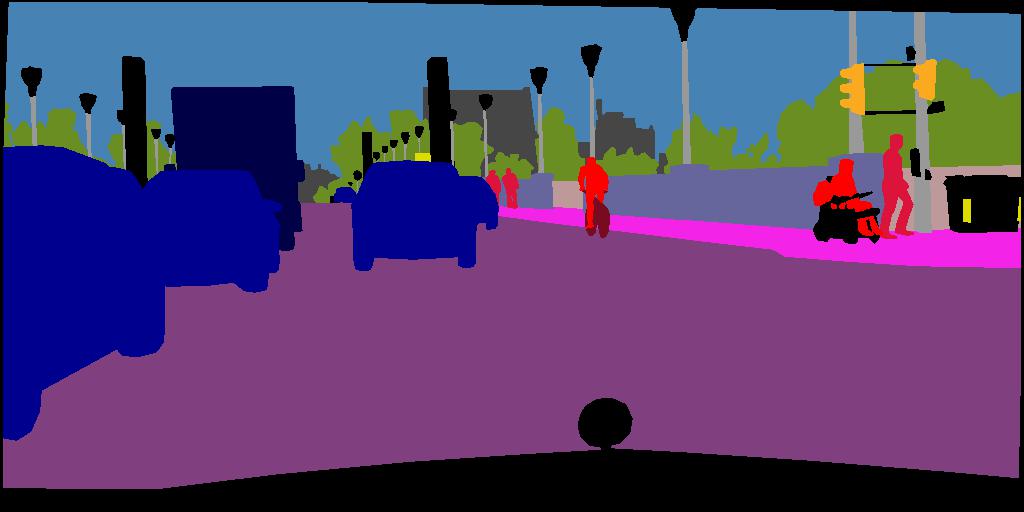}
    \caption{GT}
    \end{subfigure}
    \\ \\
\begin{subfigure}[b]{0.22\textwidth}
    \includegraphics[scale=0.1]{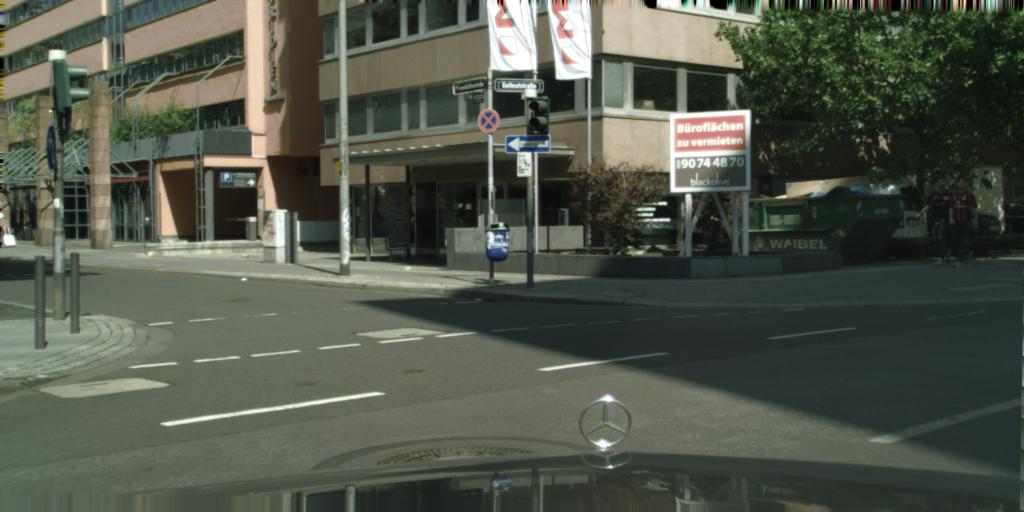}
    \caption{Image 4}
    \end{subfigure}
    \begin{subfigure}[b]{0.22\textwidth}
    \includegraphics[scale=0.1]{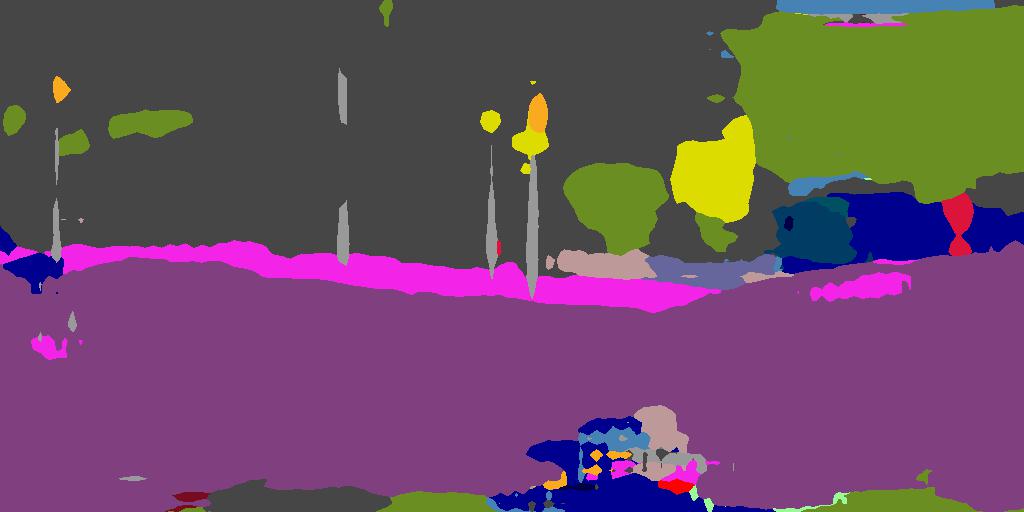}
    \caption{Student}
    \end{subfigure}
    \begin{subfigure}[b]{0.22\textwidth}
    \includegraphics[scale=0.1]{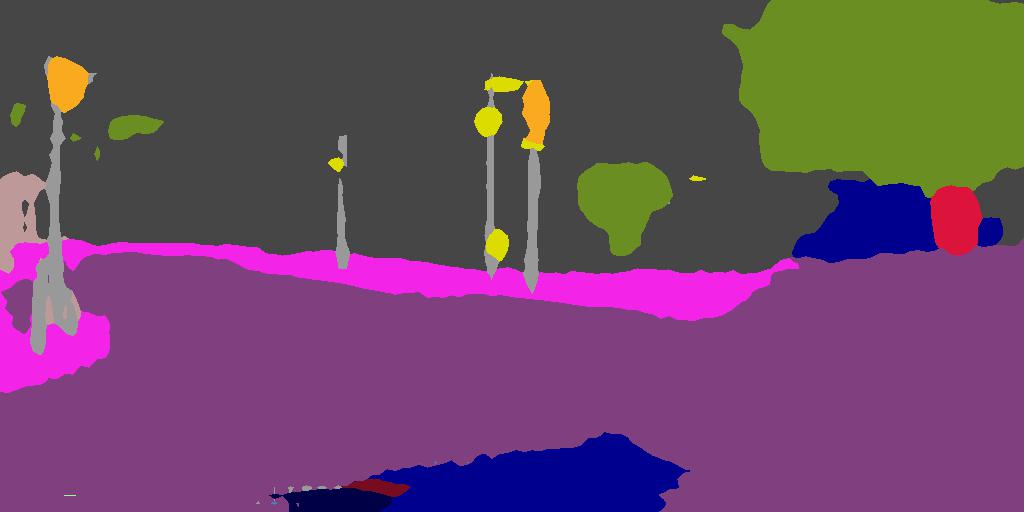}
    \caption{Teacher}
    \end{subfigure}
    \begin{subfigure}[b]{0.22\textwidth}
    \includegraphics[scale=0.1]{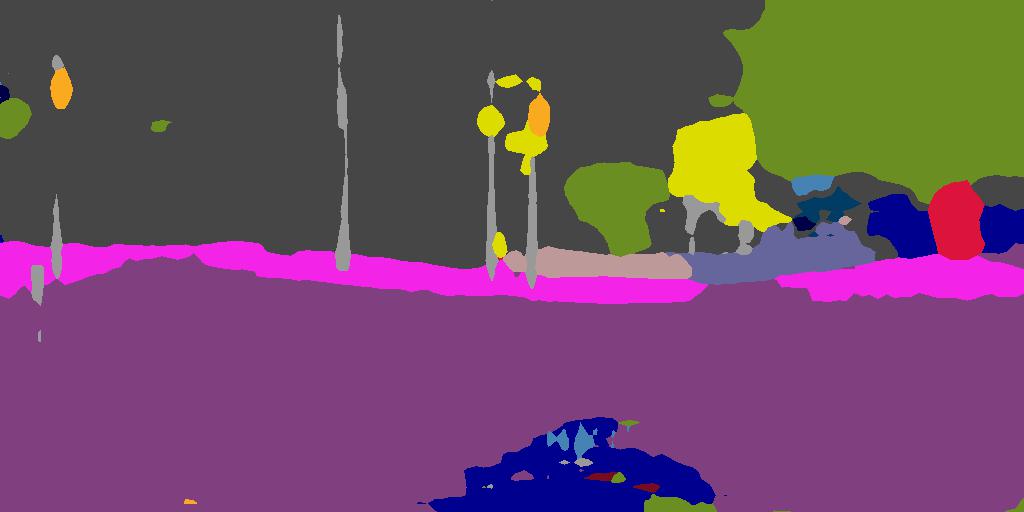}
    \caption{Source dist. (a)}
    \end{subfigure}\\
    \begin{subfigure}[b]{0.22\textwidth}
    \includegraphics[scale=0.1]{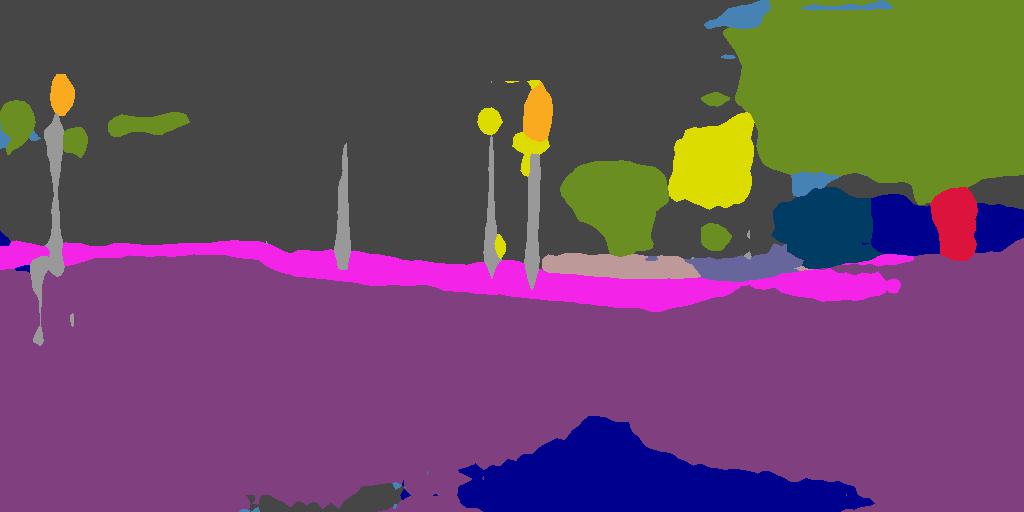}
    \caption{Target dist. (b)}
    \end{subfigure}
    \begin{subfigure}[b]{0.22\textwidth}
    \includegraphics[scale=0.1]{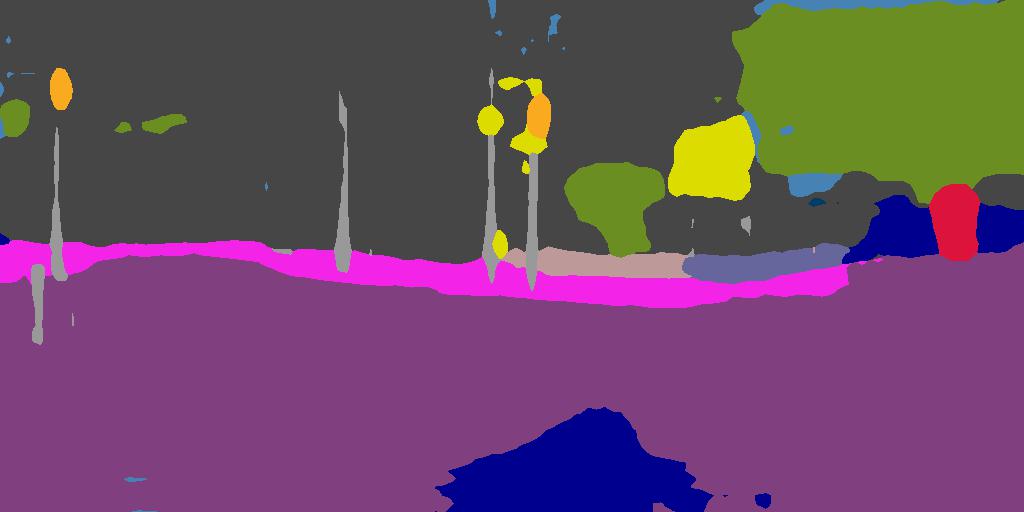}
    \caption{Src + Tgt dist. (c)}
    \end{subfigure}
    \begin{subfigure}[b]{0.22\textwidth}
    \includegraphics[scale=0.1]{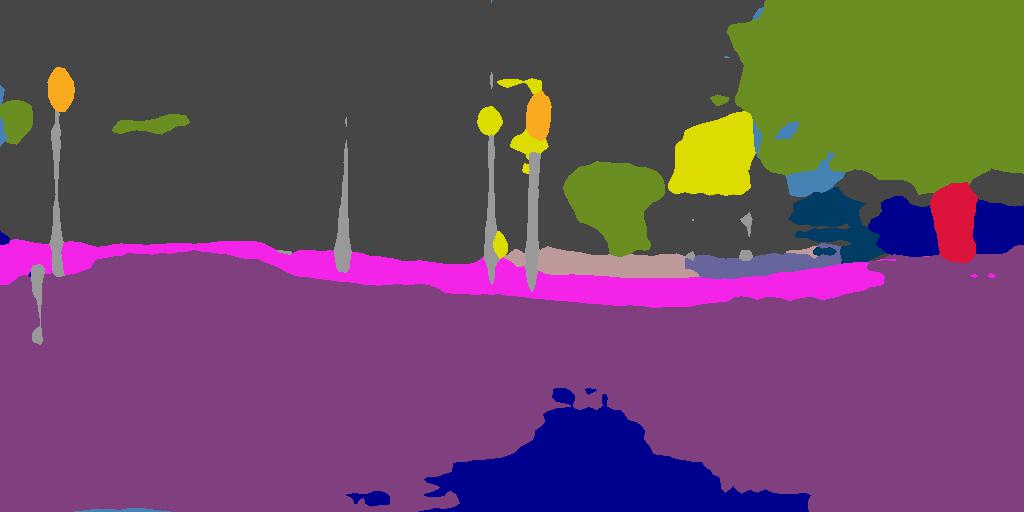}
    \caption{Target init.dist.(d)}
    \end{subfigure}
    \begin{subfigure}[b]{0.22\textwidth}
    \includegraphics[scale=0.1]{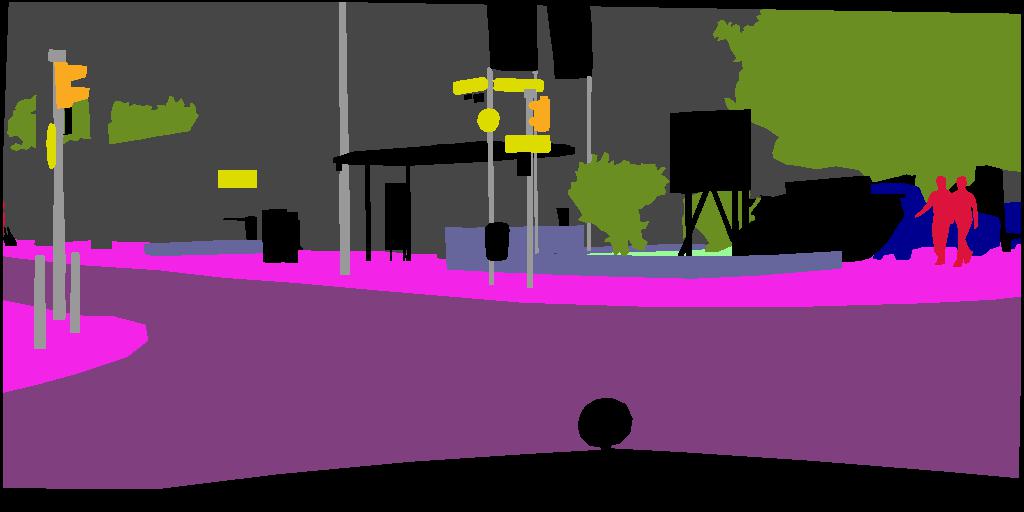}
    \caption{GT}
    \end{subfigure}\\
    
    \begin{subfigure}[b]{0.22\textwidth}
    \includegraphics[scale=0.1]{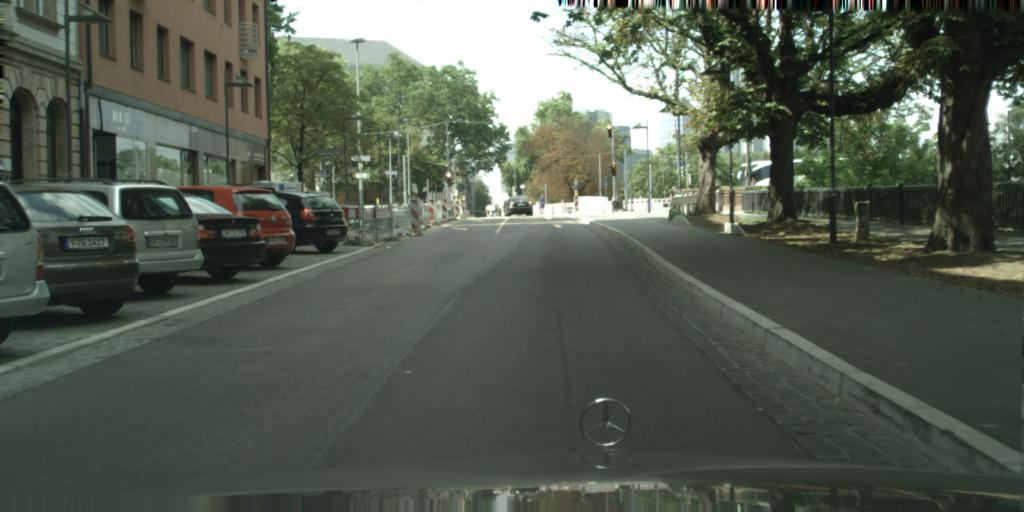}
    \caption{Image 5}
    \end{subfigure}
    \begin{subfigure}[b]{0.22\textwidth}
    \includegraphics[scale=0.1]{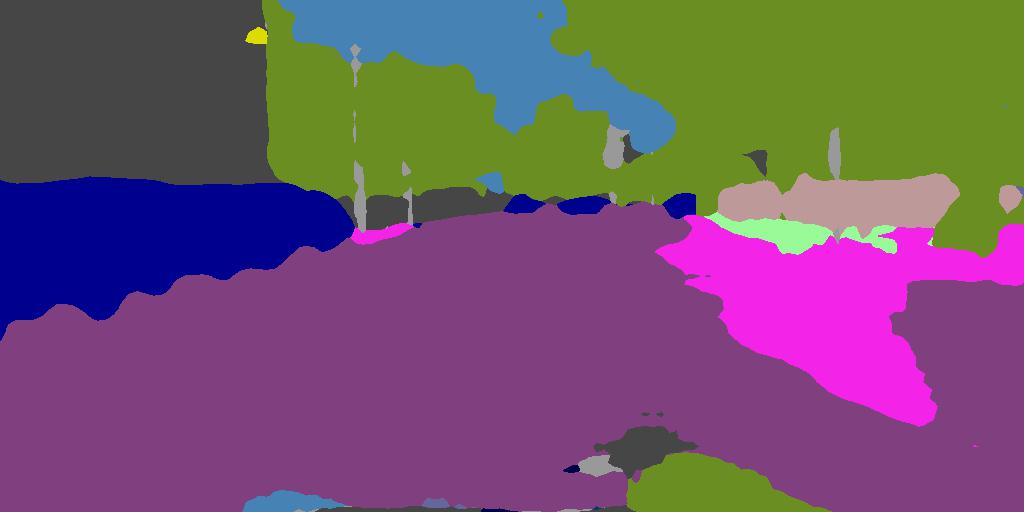}
    \caption{Student}
    \end{subfigure}
    \begin{subfigure}[b]{0.22\textwidth}
    \includegraphics[scale=0.1]{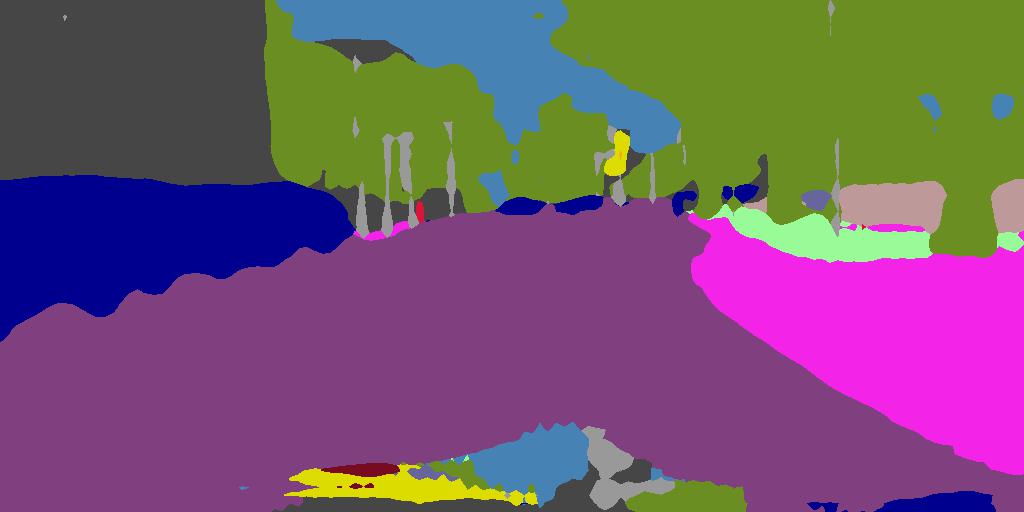}
    \caption{Teacher}
    \end{subfigure}
    \begin{subfigure}[b]{0.22\textwidth}
    \includegraphics[scale=0.1]{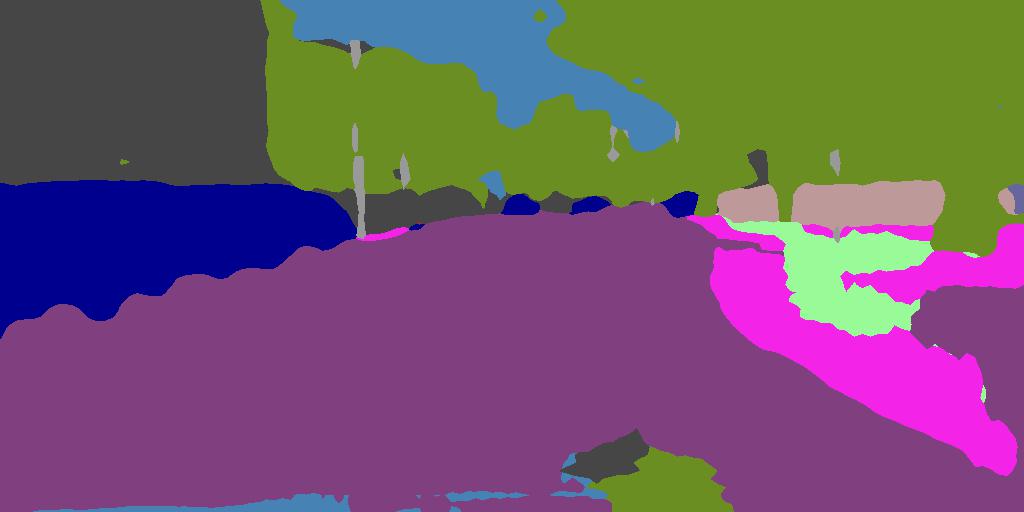}
    \caption{Source dist. (a)}
    \end{subfigure}\\
    \begin{subfigure}[b]{0.22\textwidth}
    \includegraphics[scale=0.1]{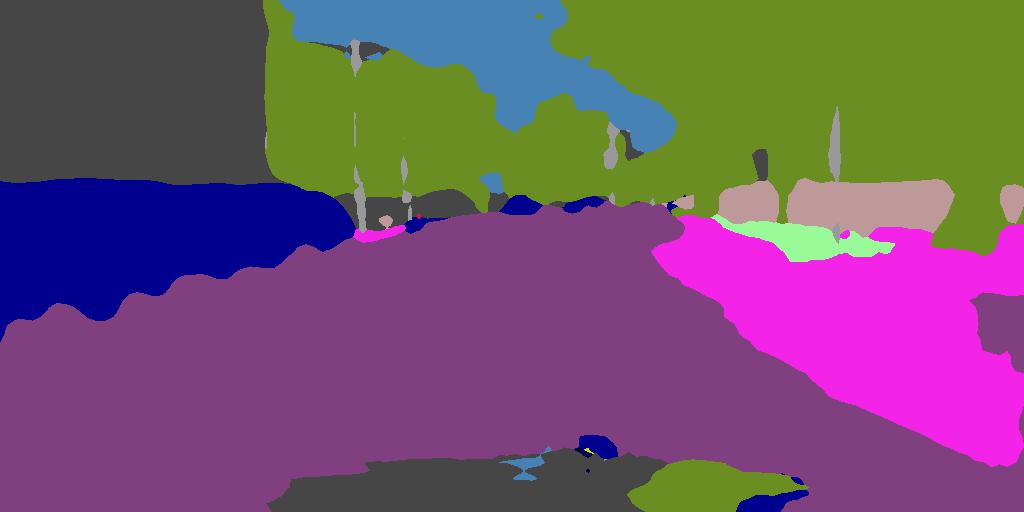}
    \caption{Target dist. (b)}
    \end{subfigure}
    \begin{subfigure}[b]{0.22\textwidth}
    \includegraphics[scale=0.1]{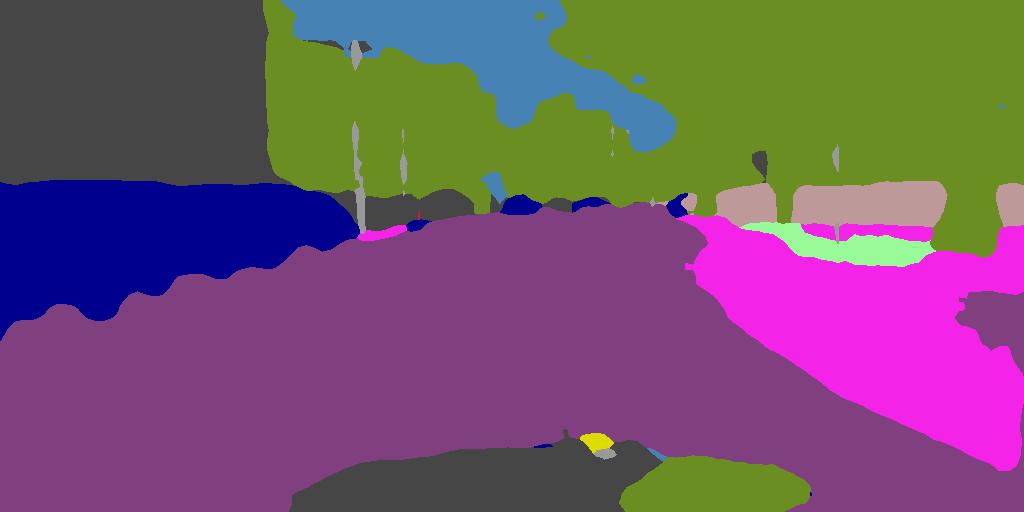}
    \caption{Src + Tgt dist. (c)}
    \end{subfigure}
    \begin{subfigure}[b]{0.22\textwidth}
    \includegraphics[scale=0.1]{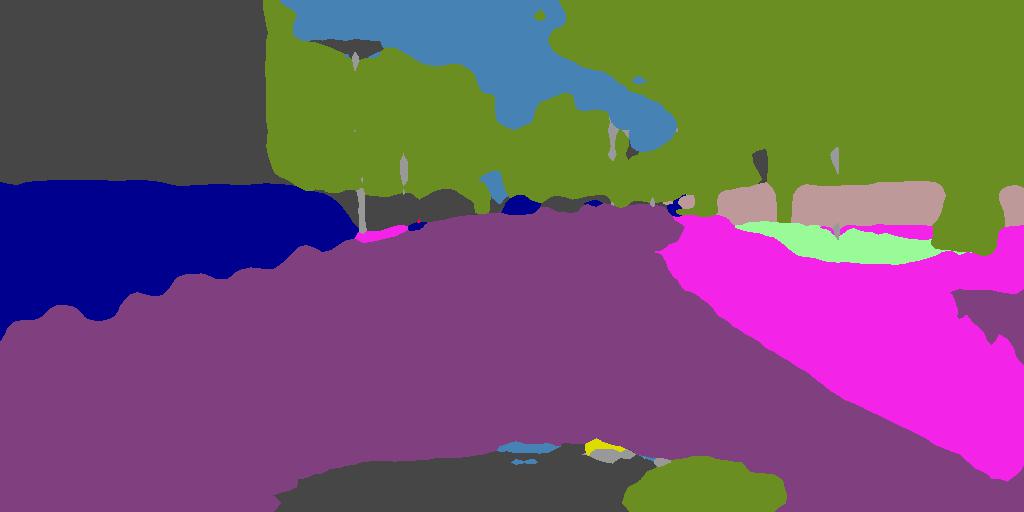}
    \caption{Target init.dist.(d)}
    \end{subfigure}
    \begin{subfigure}[b]{0.22\textwidth}
    \includegraphics[scale=0.1]{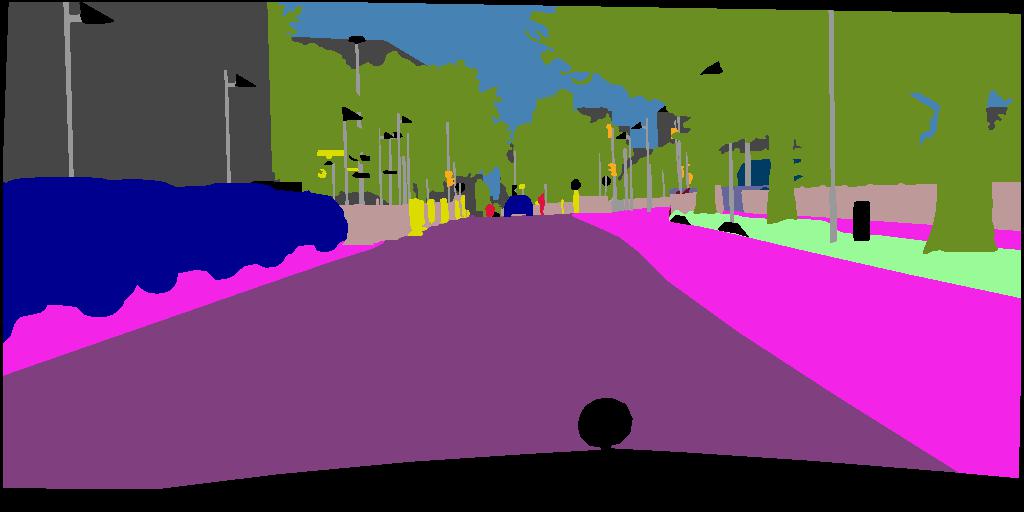}
    \caption{GT}
    \end{subfigure} \\
    \caption{Visual results: BDD to Cityscapes}
    \label{fig:visualisations_bdd2cs}    
\end{figure*}

\subsection{GTA5 to cityscapes}
This section has visual results for the synthetic-to-real adaptation case: GTA5 to Cityscapes. (Fig. 2)
\begin{figure*}[!htbp]
    \centering
    \captionsetup[subfigure]{labelformat=empty}
    \begin{subfigure}[b]{0.22\textwidth}
    \includegraphics[scale=0.1]{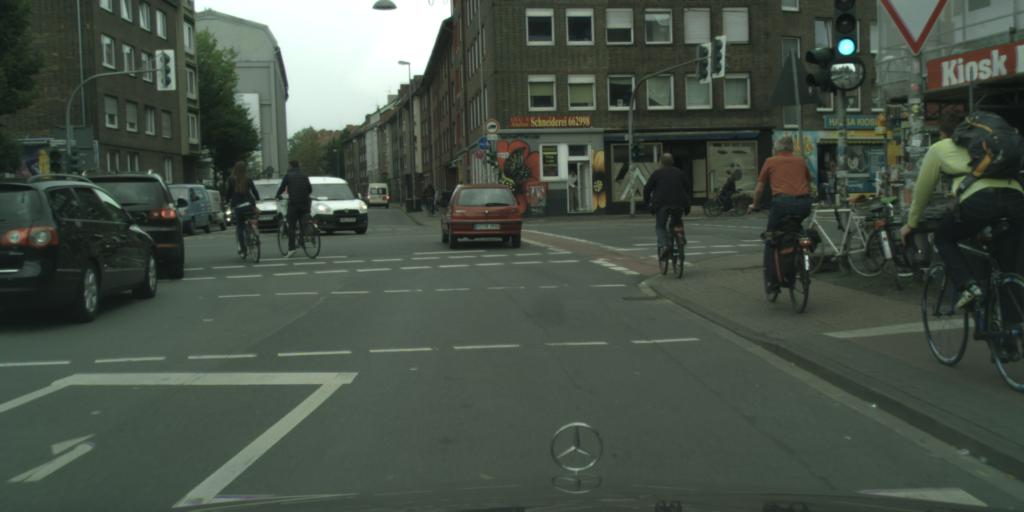}
    \caption{Image 1}
    \end{subfigure}
    \begin{subfigure}[b]{0.22\textwidth}
    \includegraphics[scale=0.1]{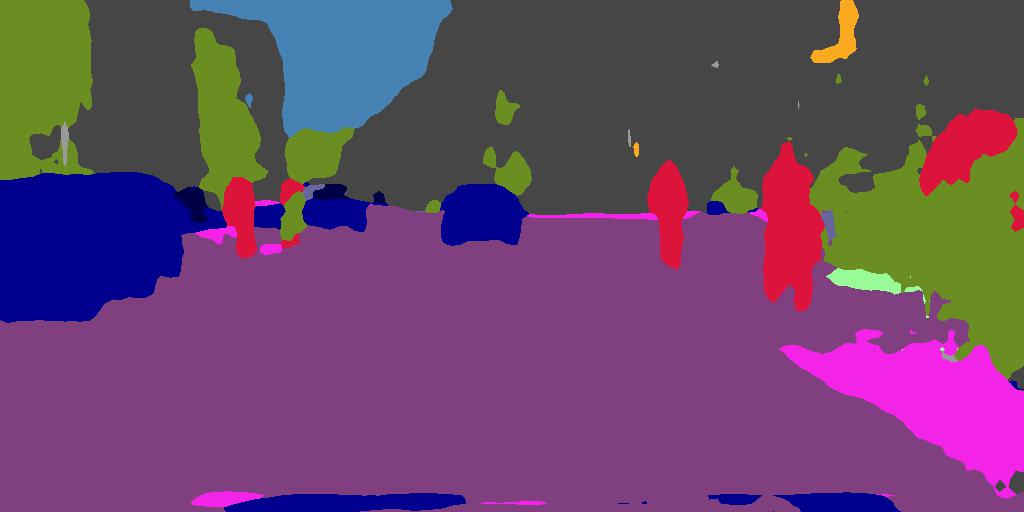}
    \caption{Student}
    \end{subfigure}
    \begin{subfigure}[b]{0.22\textwidth}
    \includegraphics[scale=0.1]{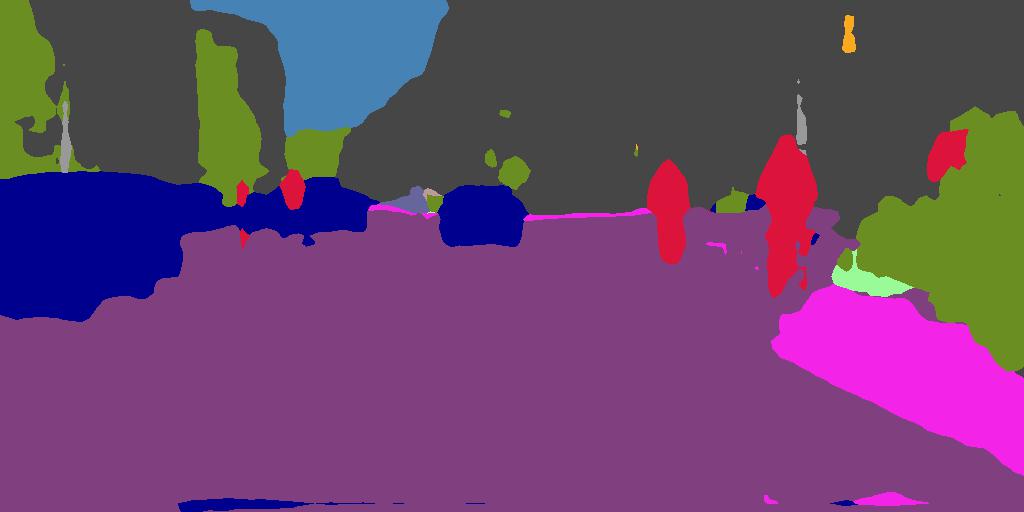}
    \caption{Teacher}
    \end{subfigure}
    \begin{subfigure}[b]{0.22\textwidth}
    \includegraphics[scale=0.1]{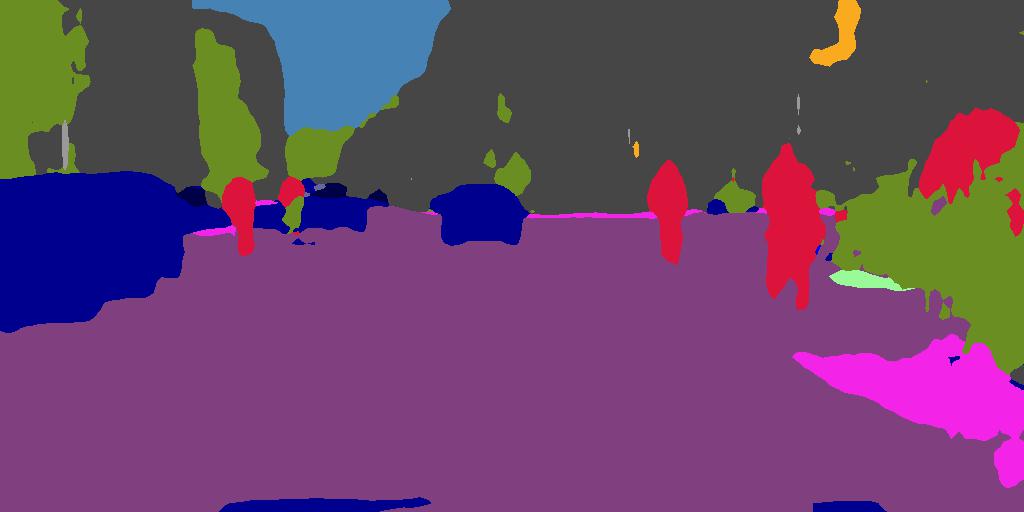}
    \caption{Source dist. (a)}
    \end{subfigure}\\
    \begin{subfigure}[b]{0.22\textwidth}
    \includegraphics[scale=0.1]{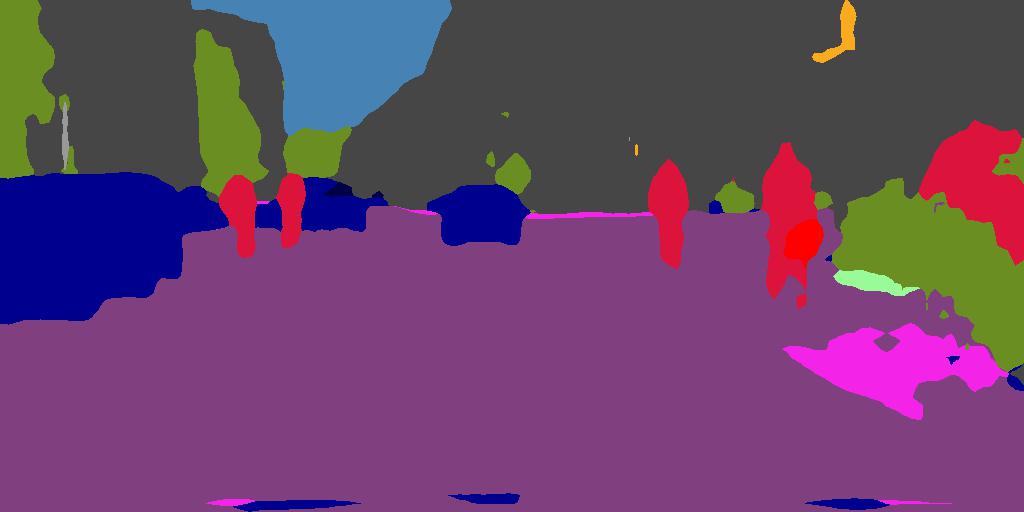}
    \caption{Target dist. (b)}
    \end{subfigure}
    \begin{subfigure}[b]{0.22\textwidth}
    \includegraphics[scale=0.1]{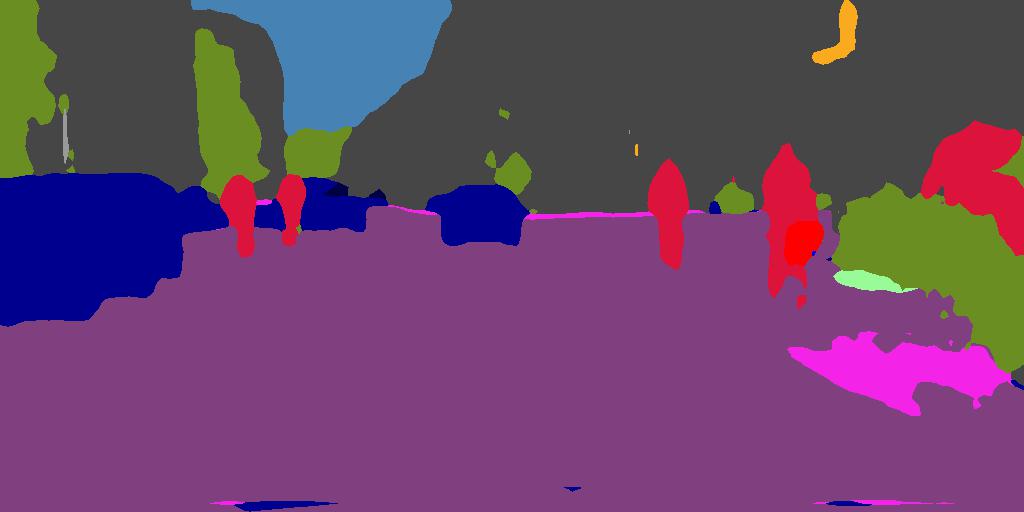}
    \caption{Src + Tgt dist. (c)}
    \end{subfigure}
    \begin{subfigure}[b]{0.22\textwidth}
    \includegraphics[scale=0.1]{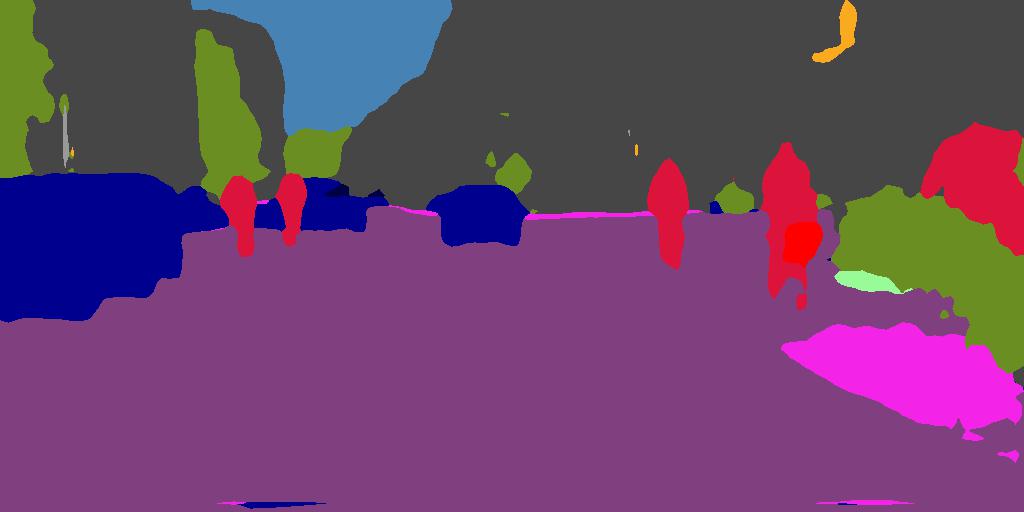}
    \caption{Target init.dist.(d)}
    \end{subfigure}
    \begin{subfigure}[b]{0.22\textwidth}
    \includegraphics[scale=0.1]{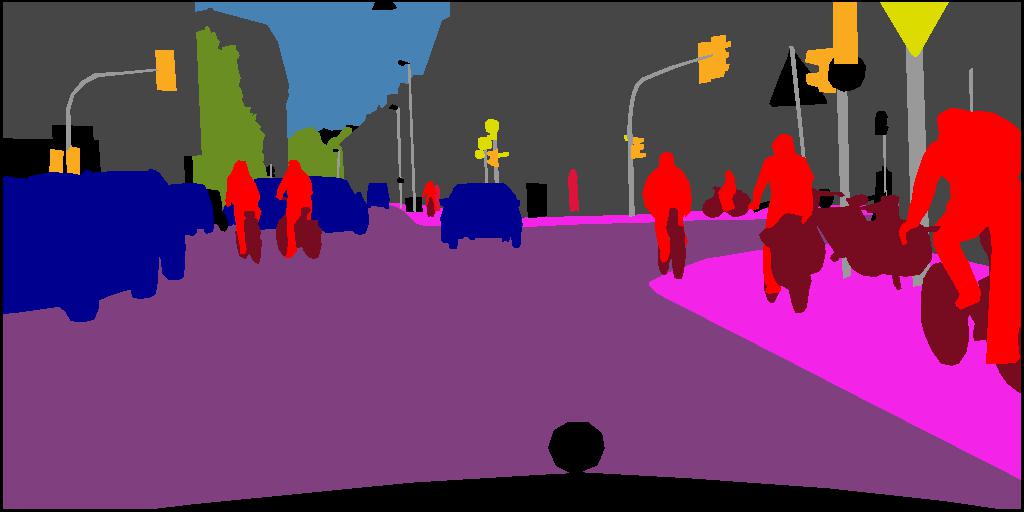}
    \caption{GT}
    \end{subfigure} \\
    
    \begin{subfigure}[b]{0.22\textwidth}
    \includegraphics[scale=0.1]{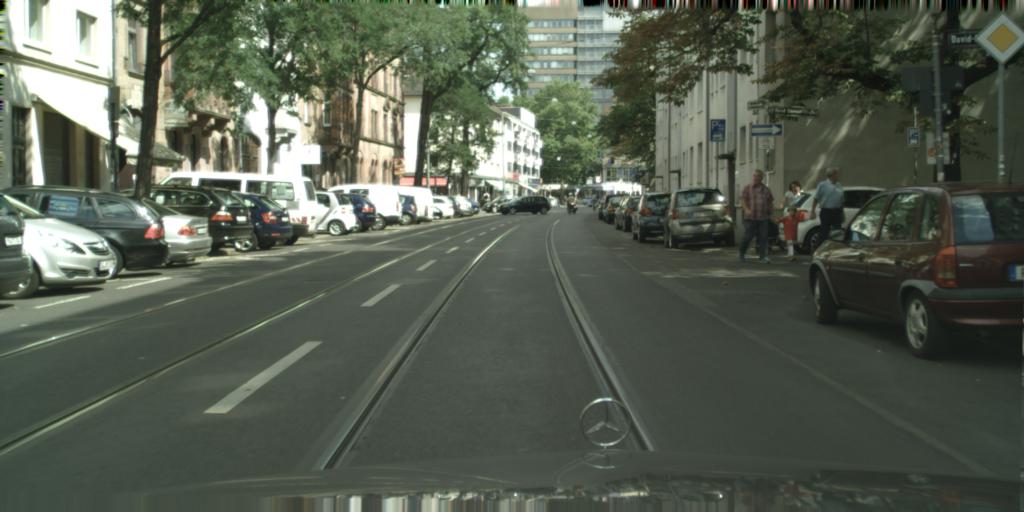}
    \caption{Image 2}
    \end{subfigure}
    \begin{subfigure}[b]{0.22\textwidth}
    \includegraphics[scale=0.1]{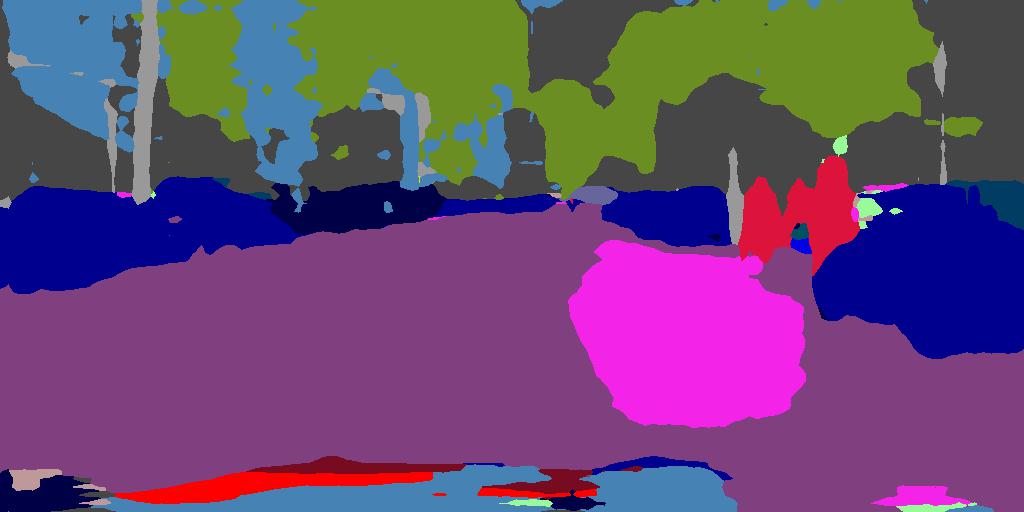}
    \caption{Student}
    \end{subfigure}
    \begin{subfigure}[b]{0.22\textwidth}
    \includegraphics[scale=0.1]{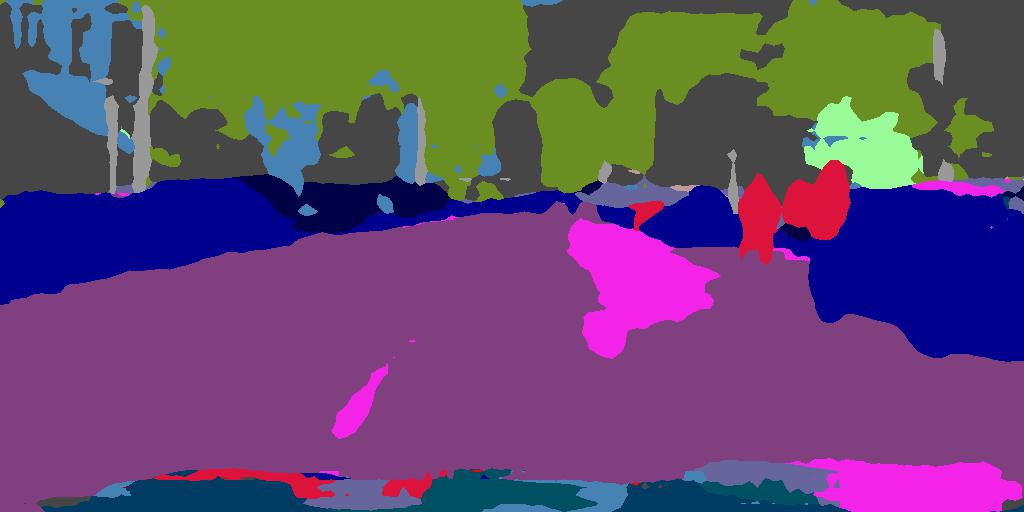}
    \caption{Teacher}
    \end{subfigure}
    \begin{subfigure}[b]{0.22\textwidth}
    \includegraphics[scale=0.1]{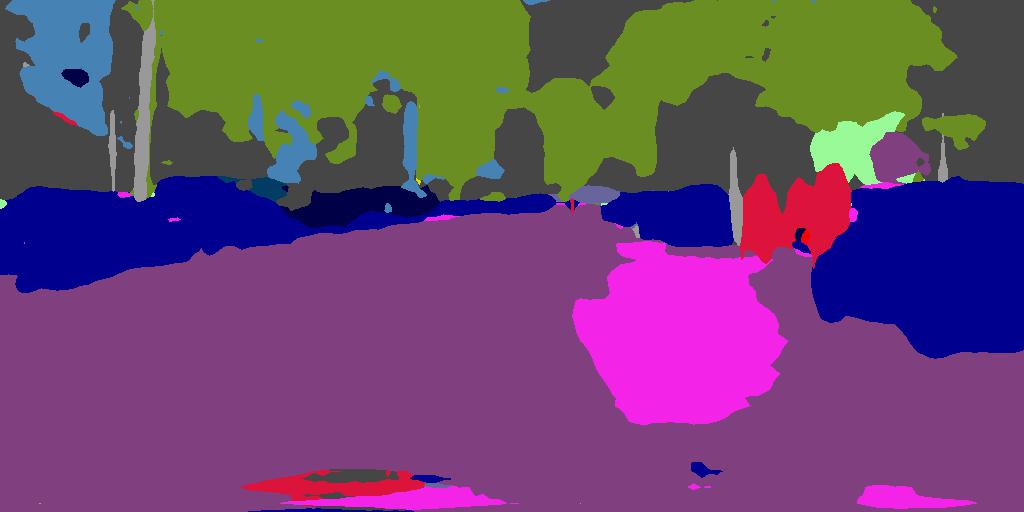}
    \caption{Source dist. (a)}
    \end{subfigure}\\
    \begin{subfigure}[b]{0.22\textwidth}
    \includegraphics[scale=0.1]{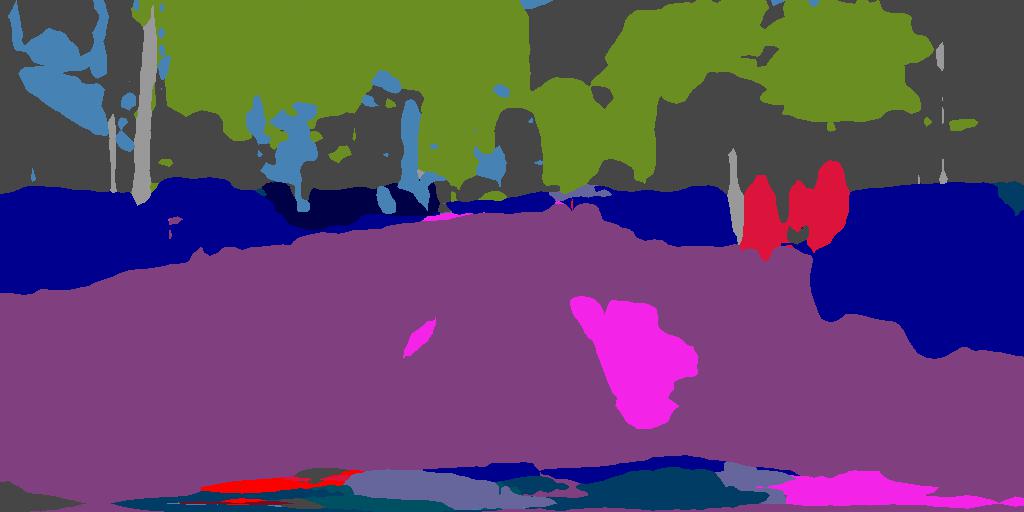}
    \caption{Target dist. (b)}
    \end{subfigure}
    \begin{subfigure}[b]{0.22\textwidth}
    \includegraphics[scale=0.1]{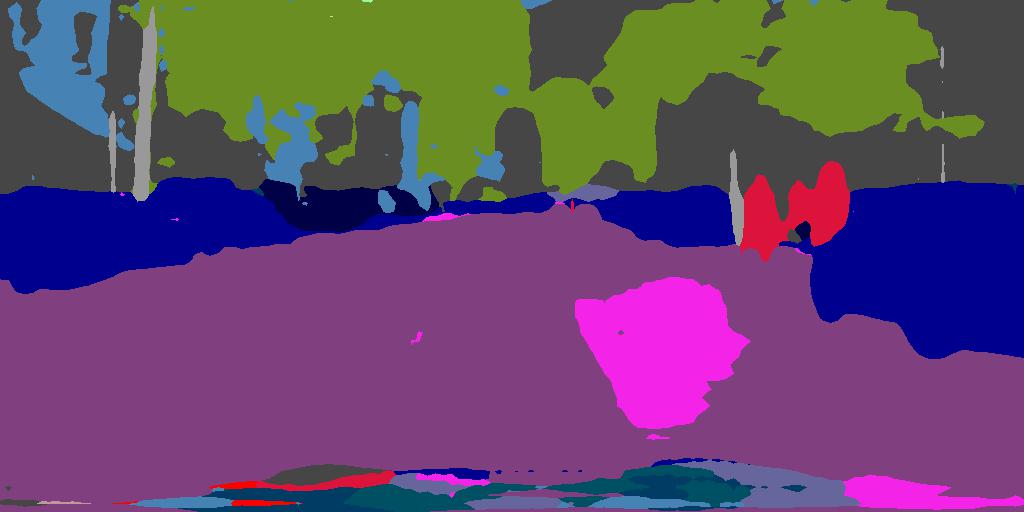}
    \caption{Src + Tgt dist. (c)}
    \end{subfigure}
    \begin{subfigure}[b]{0.22\textwidth}
    \includegraphics[scale=0.1]{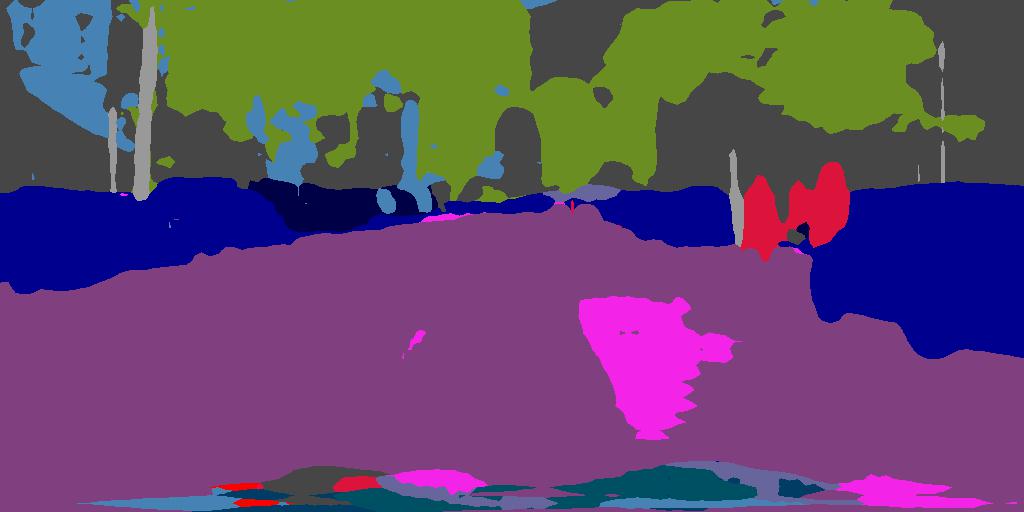}
    \caption{Target init.dist.(d)}
    \end{subfigure}
    \begin{subfigure}[b]{0.22\textwidth}
    \includegraphics[scale=0.1]{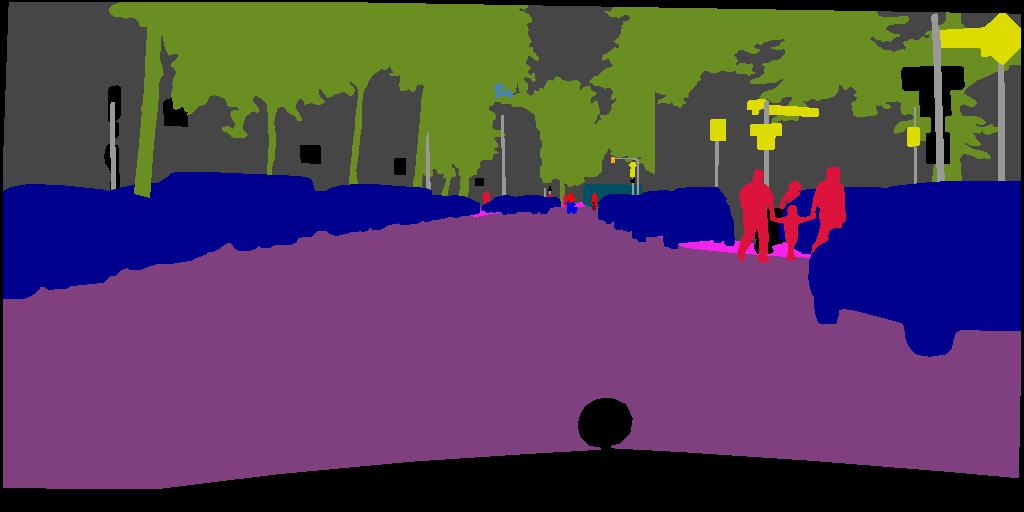}
    \caption{GT}
    \end{subfigure} \\
    
    \begin{subfigure}[b]{0.22\textwidth}
    \includegraphics[scale=0.1]{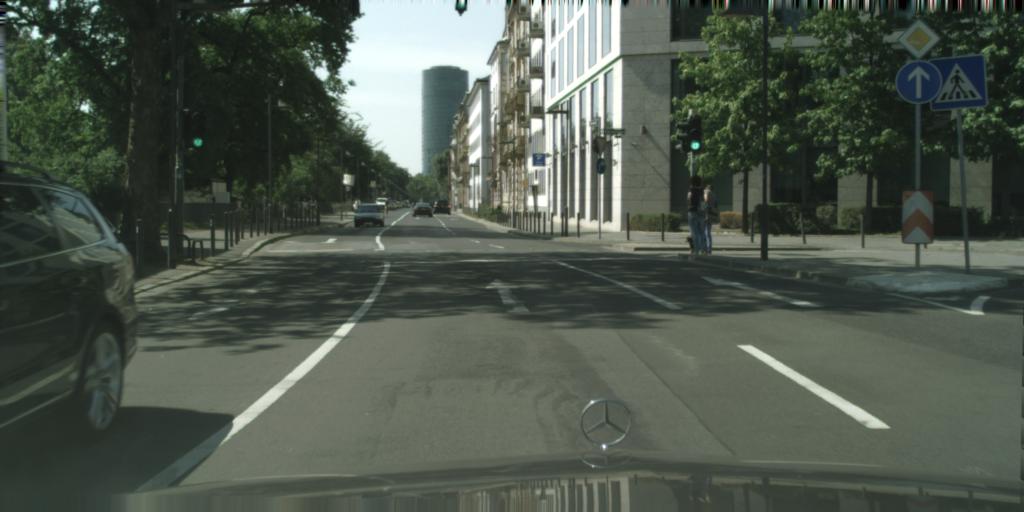}
    \caption{Image 3}
    \end{subfigure}
    \begin{subfigure}[b]{0.22\textwidth}
    \includegraphics[scale=0.1]{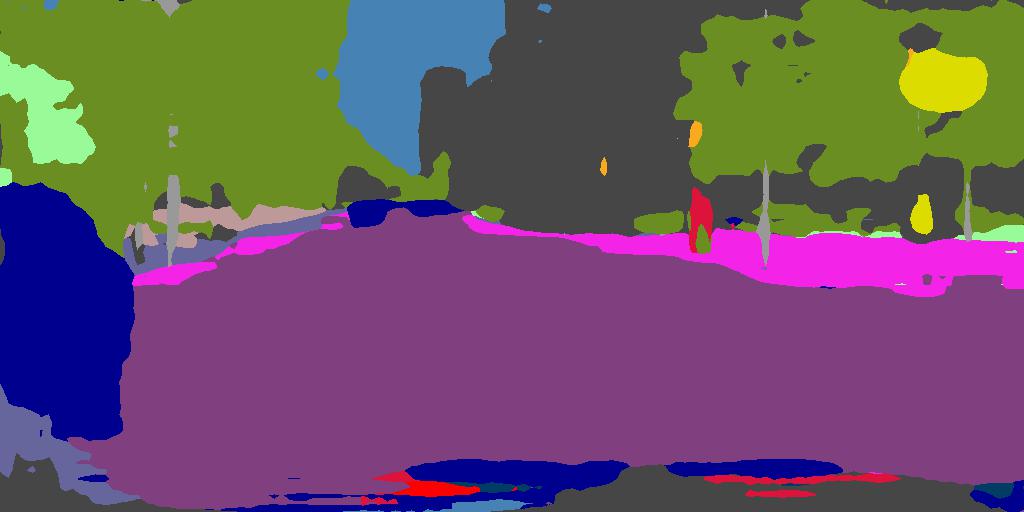}
    \caption{Student}
    \end{subfigure}
    \begin{subfigure}[b]{0.22\textwidth}
    \includegraphics[scale=0.1]{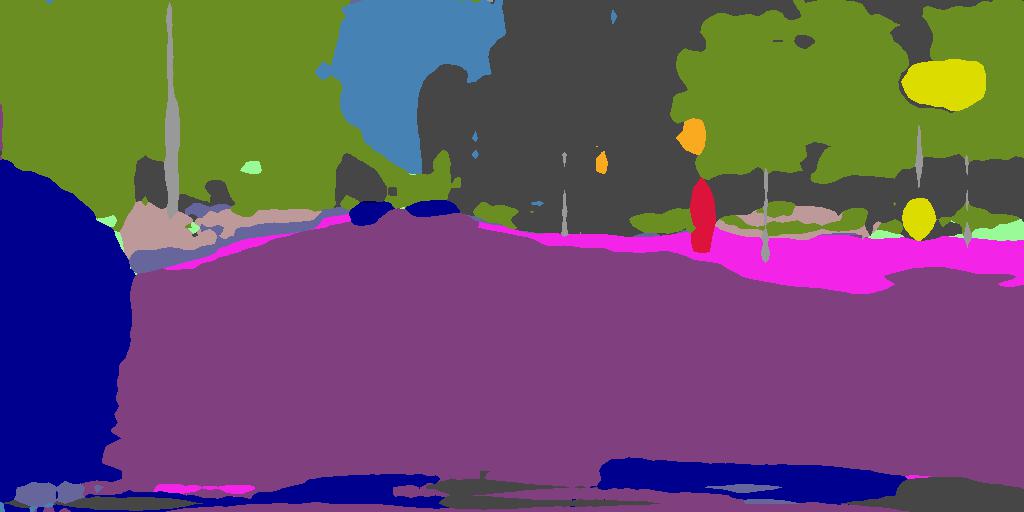}
    \caption{Teacher}
    \end{subfigure}
    \begin{subfigure}[b]{0.22\textwidth}
    \includegraphics[scale=0.1]{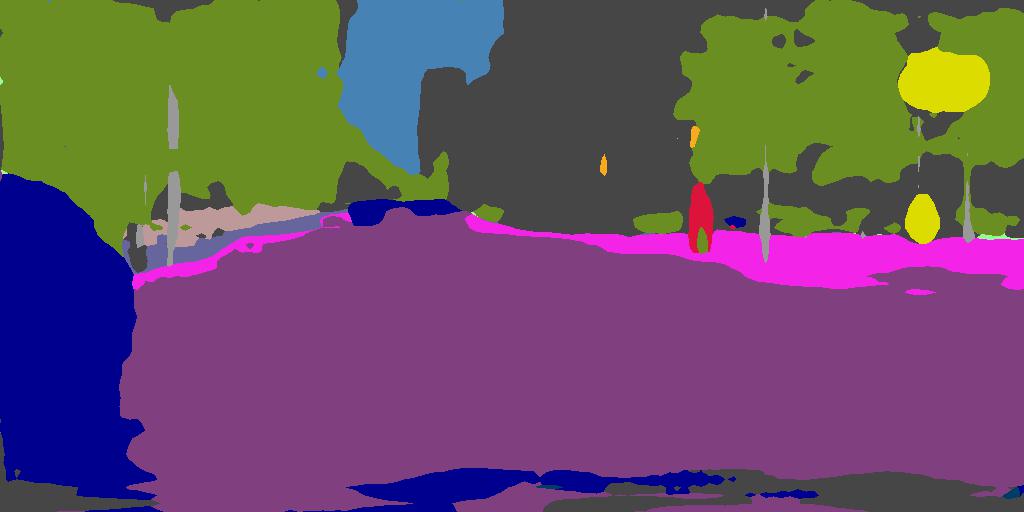}
    \caption{Source dist. (a)}
    \end{subfigure}\\
    \begin{subfigure}[b]{0.22\textwidth}
    \includegraphics[scale=0.1]{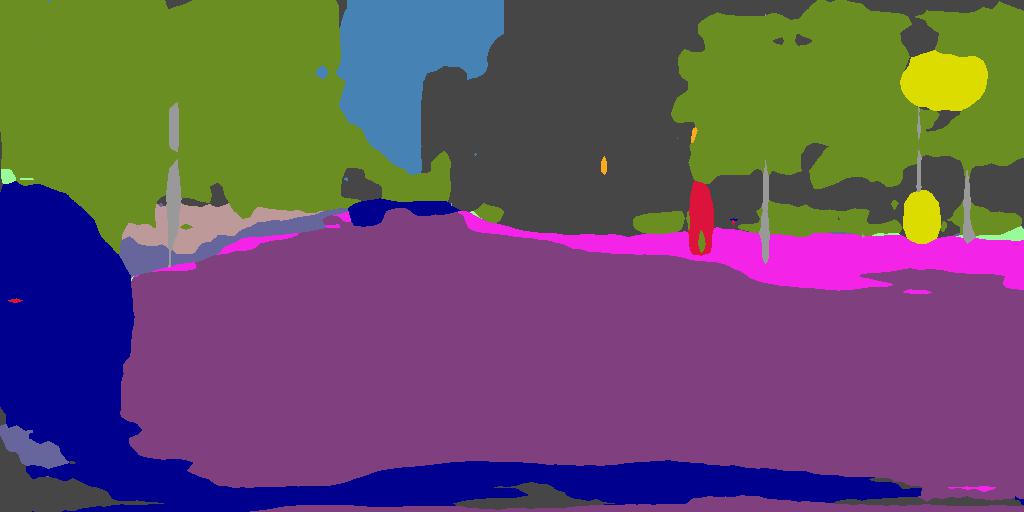}
    \caption{Target dist. (b)}
    \end{subfigure}
    \begin{subfigure}[b]{0.22\textwidth}
    \includegraphics[scale=0.1]{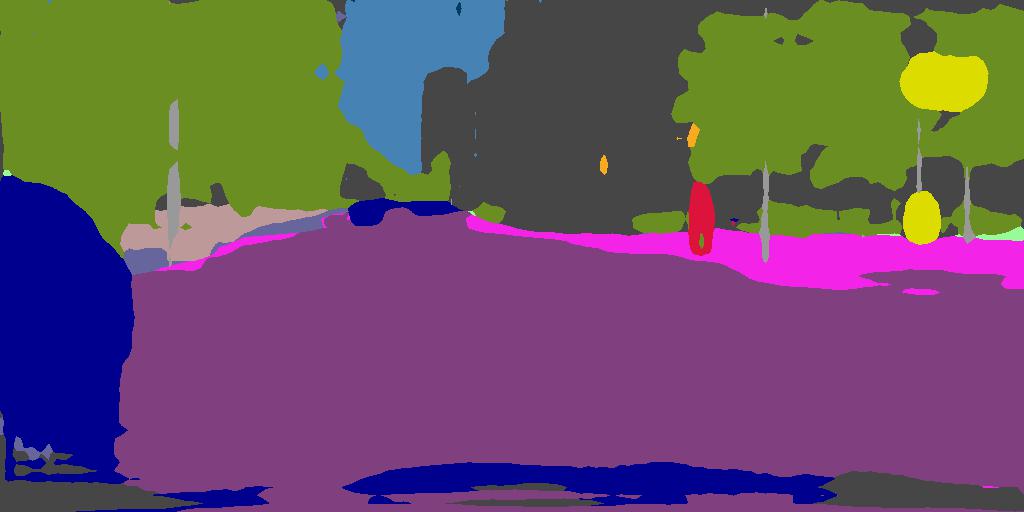}
    \caption{Src + Tgt dist. (c)}
    \end{subfigure}
    \begin{subfigure}[b]{0.22\textwidth}
    \includegraphics[scale=0.1]{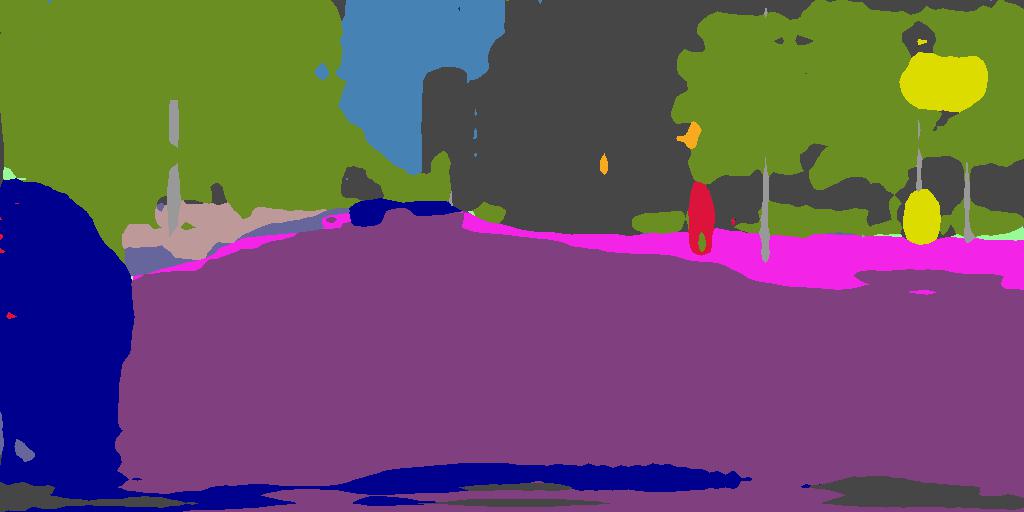}
    \caption{Target init.dist.(d)}
    \end{subfigure}
    \begin{subfigure}[b]{0.22\textwidth}
    \includegraphics[scale=0.1]{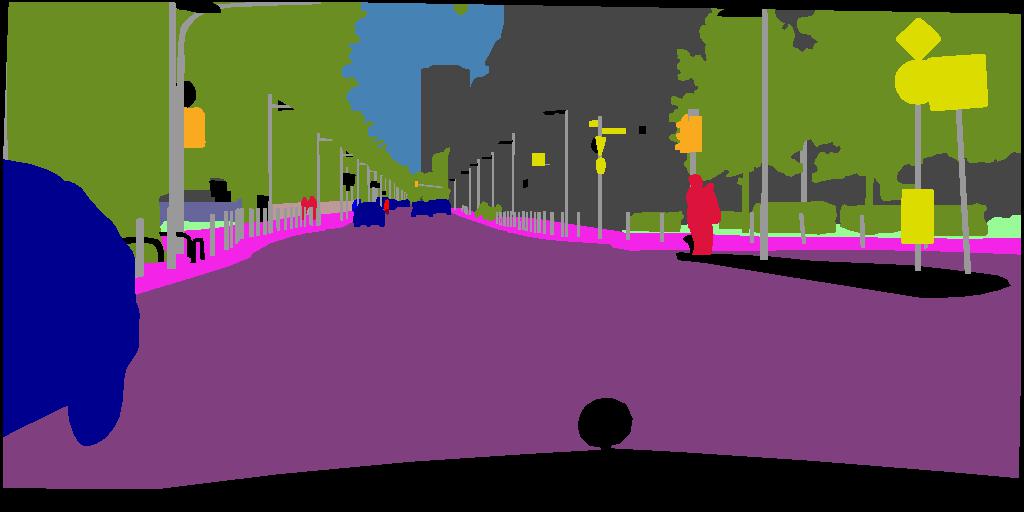}
    \caption{GT}
    \end{subfigure}\\
    
    \begin{subfigure}[b]{0.22\textwidth}
    \includegraphics[scale=0.1]{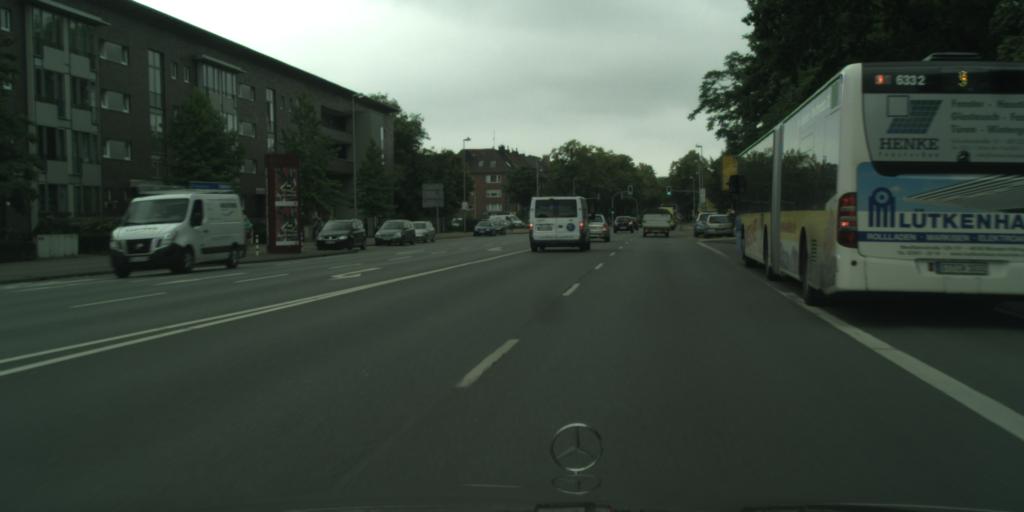}
    \caption{Image 4}
    \end{subfigure}
    \begin{subfigure}[b]{0.22\textwidth}
    \includegraphics[scale=0.1]{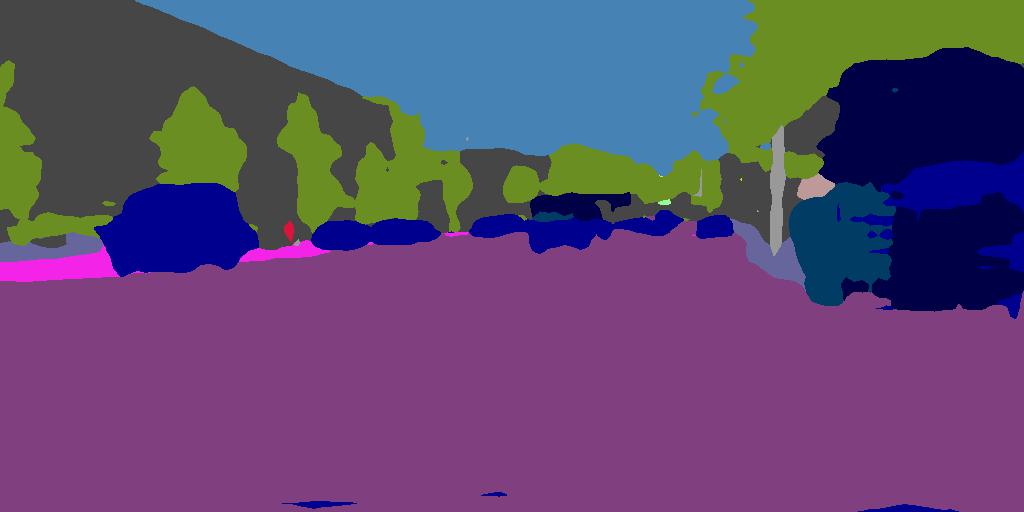}
    \caption{Student}
    \end{subfigure}
    \begin{subfigure}[b]{0.22\textwidth}
    \includegraphics[scale=0.1]{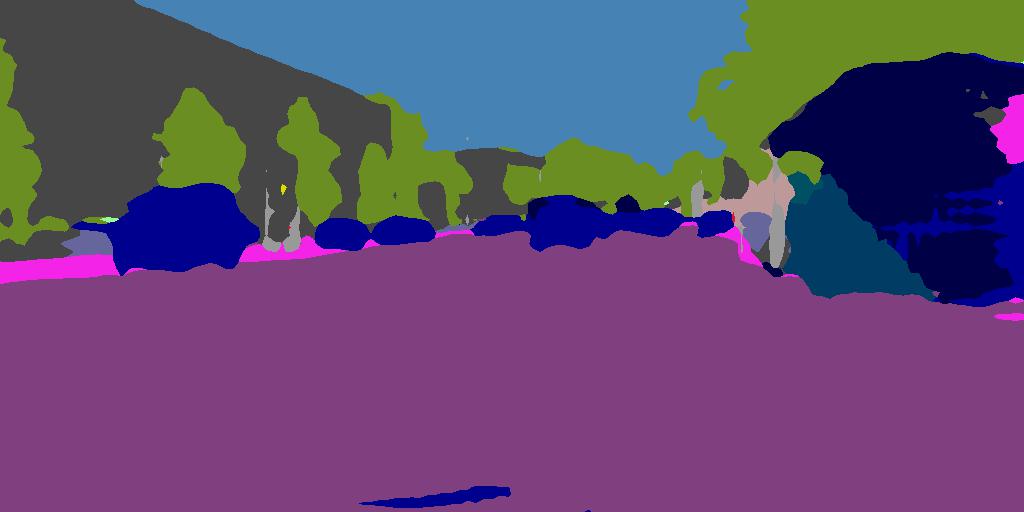}
    \caption{Teacher}
    \end{subfigure}
    \begin{subfigure}[b]{0.22\textwidth}
    \includegraphics[scale=0.1]{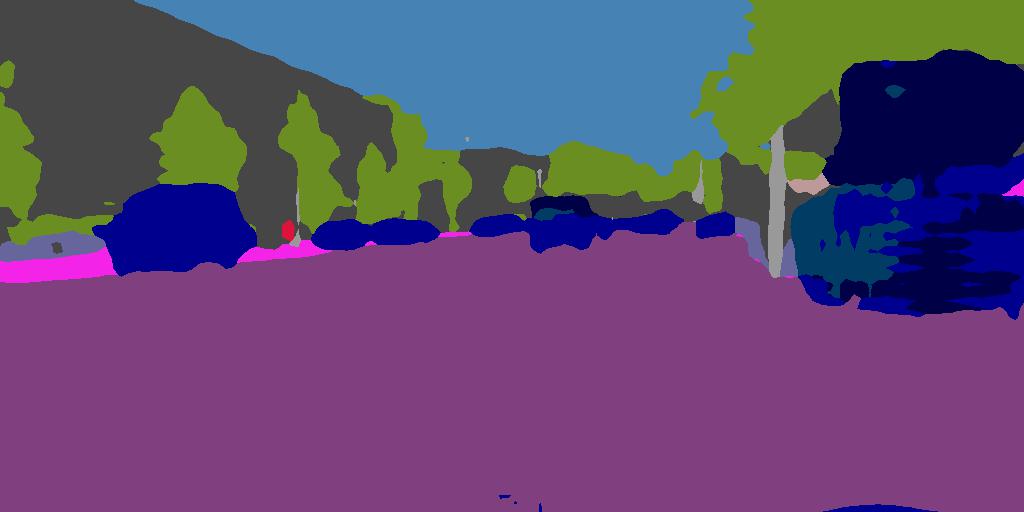}
    \caption{Source dist. (a)}
    \end{subfigure}\\
    \begin{subfigure}[b]{0.22\textwidth}
    \includegraphics[scale=0.1]{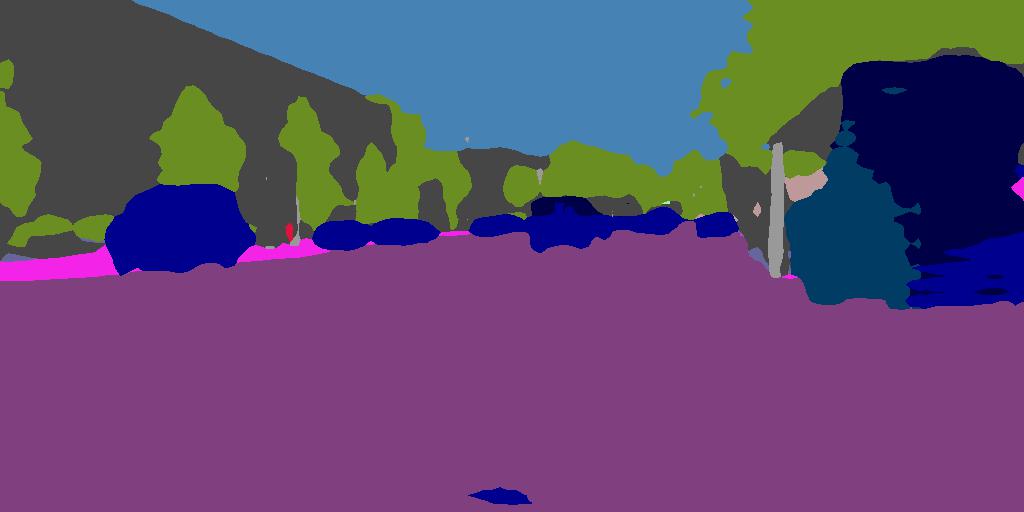}
    \caption{Target dist. (b)}
    \end{subfigure}
    \begin{subfigure}[b]{0.22\textwidth}
    \includegraphics[scale=0.1]{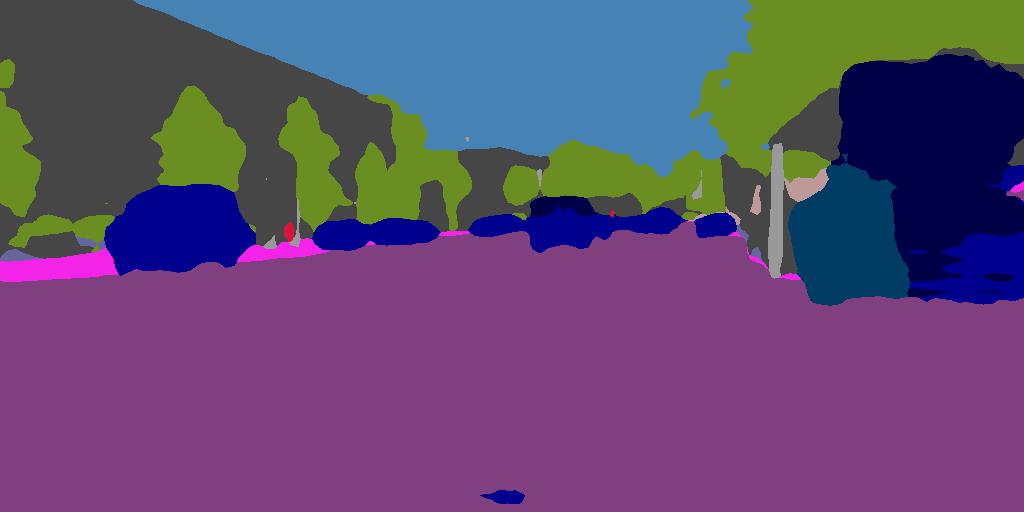}
    \caption{Src + Tgt dist. (c)}
    \end{subfigure}
    \begin{subfigure}[b]{0.22\textwidth}
    \includegraphics[scale=0.1]{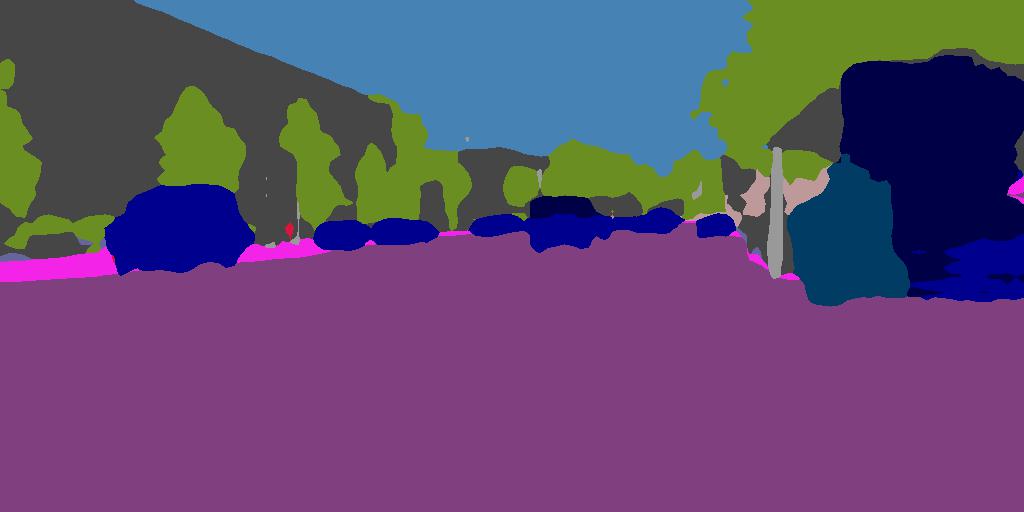}
    \caption{Target init.dist.(d)}
    \end{subfigure}
    \begin{subfigure}[b]{0.22\textwidth}
    \includegraphics[scale=0.1]{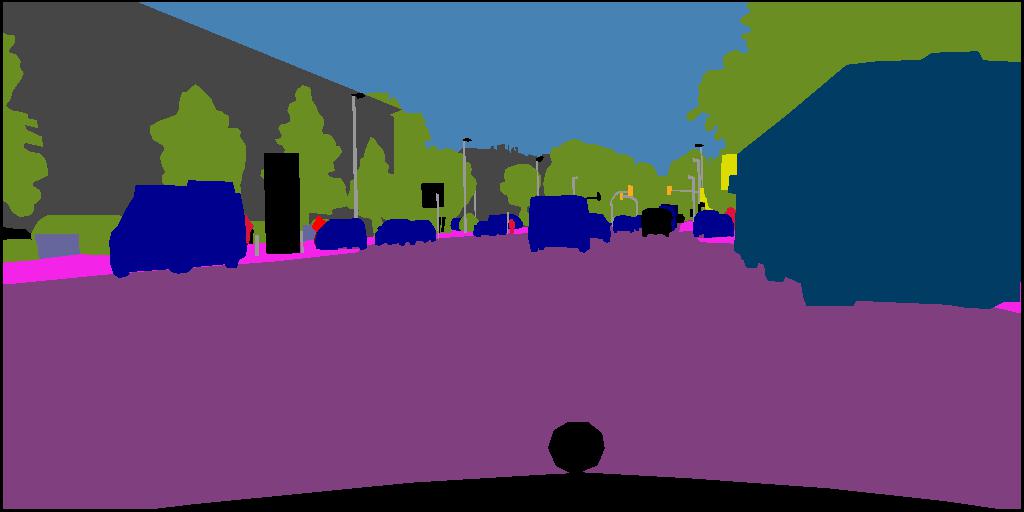}
    \caption{GT}
    \end{subfigure}\\
    
    \begin{subfigure}[b]{0.22\textwidth}
    \includegraphics[scale=0.1]{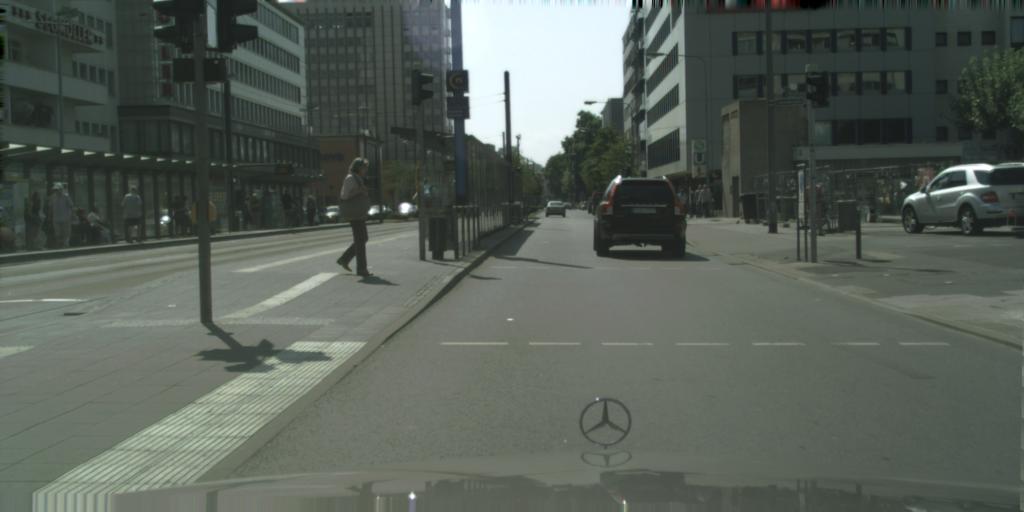}
    \caption{Image 5}
    \end{subfigure}
    \begin{subfigure}[b]{0.22\textwidth}
    \includegraphics[scale=0.1]{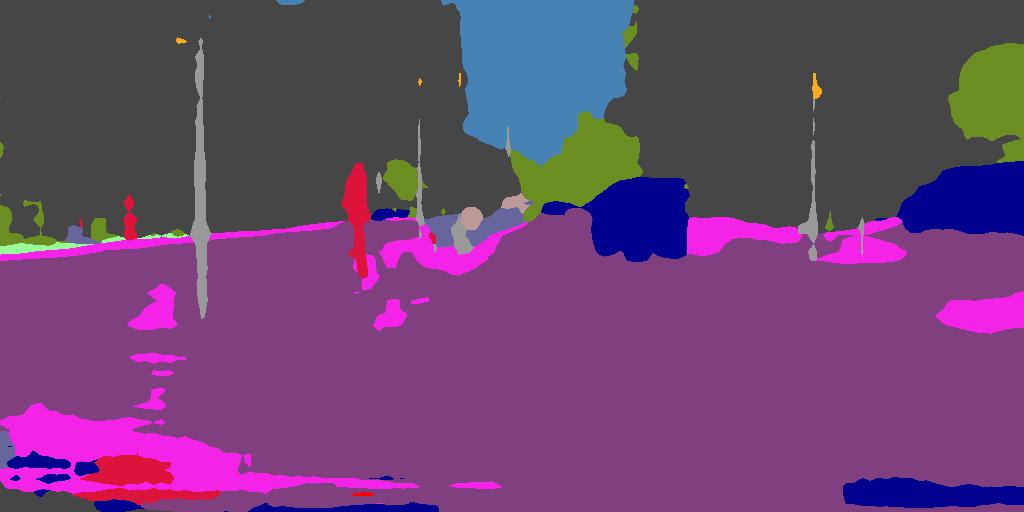}
    \caption{Student}
    \end{subfigure}
    \begin{subfigure}[b]{0.22\textwidth}
    \includegraphics[scale=0.1]{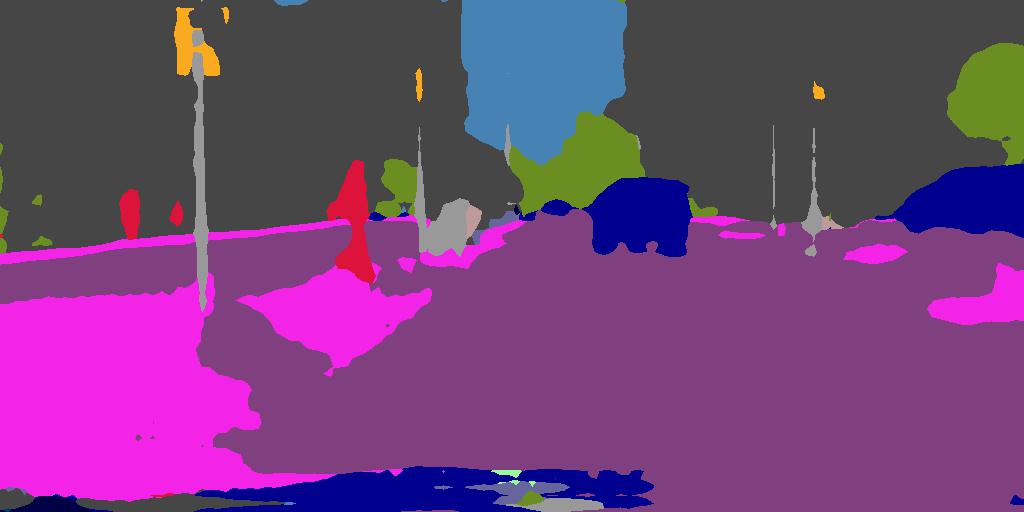}
    \caption{Teacher}
    \end{subfigure}
    \begin{subfigure}[b]{0.22\textwidth}
    \includegraphics[scale=0.1]{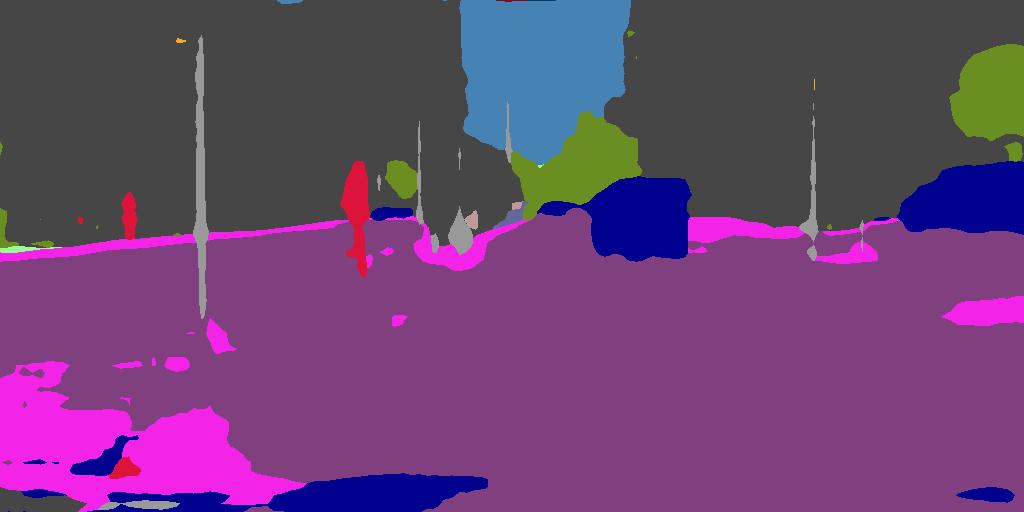}
    \caption{Source dist. (a)}
    \end{subfigure}\\
    \begin{subfigure}[b]{0.22\textwidth}
    \includegraphics[scale=0.1]{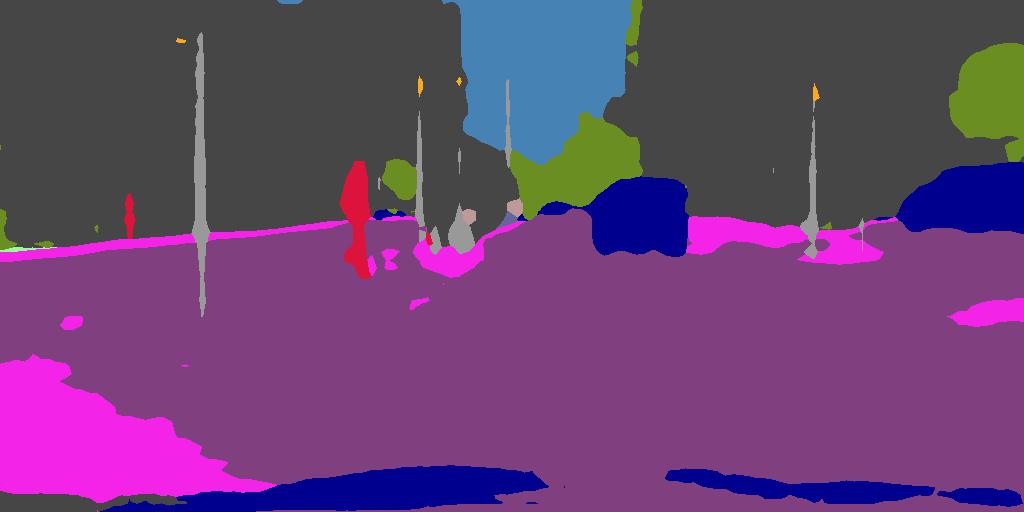}
    \caption{Target dist. (b)}
    \end{subfigure}
    \begin{subfigure}[b]{0.22\textwidth}
    \includegraphics[scale=0.1]{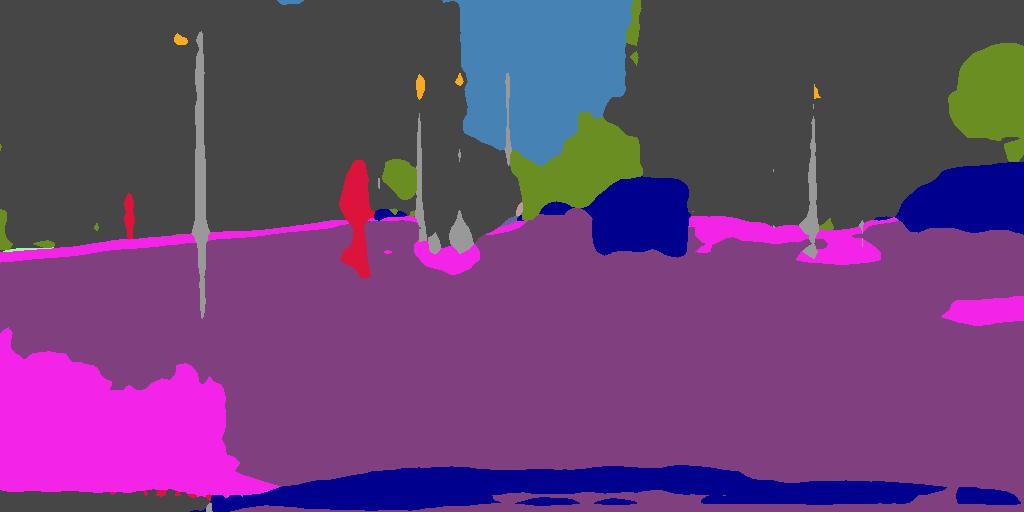}
    \caption{Src + Tgt dist. (c)}
    \end{subfigure}
    \begin{subfigure}[b]{0.22\textwidth}
    \includegraphics[scale=0.1]{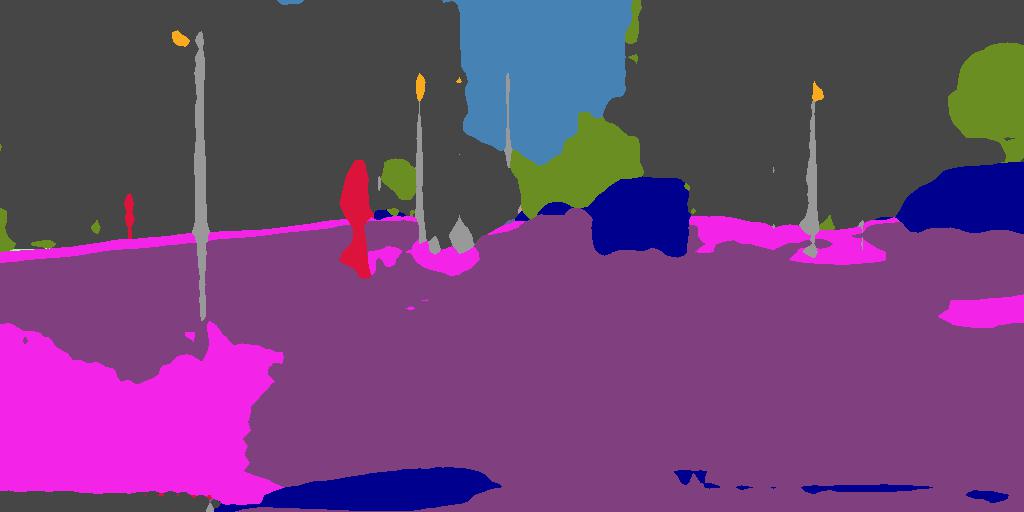}
    \caption{Target init.dist.(d)}
    \end{subfigure}
    \begin{subfigure}[b]{0.22\textwidth}
    \includegraphics[scale=0.1]{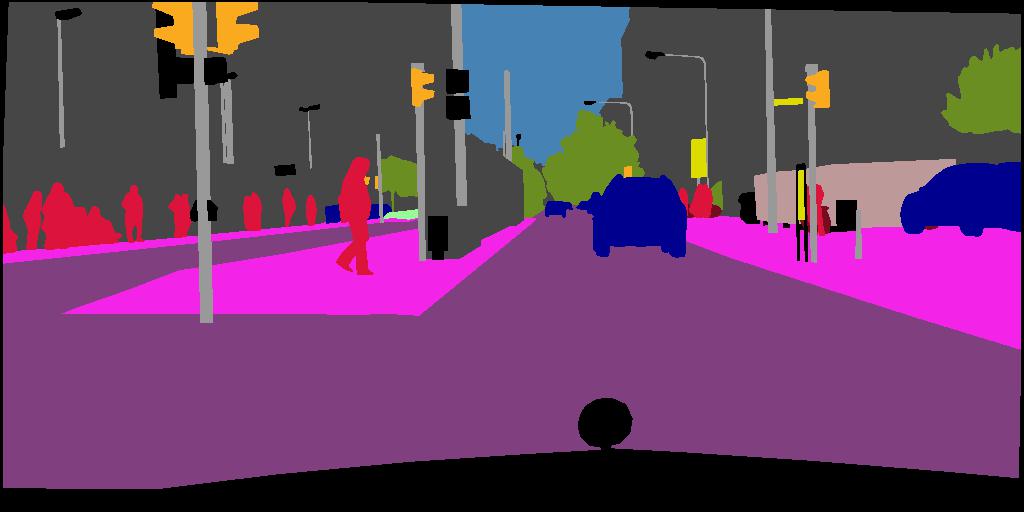}
    \caption{GT}
    \end{subfigure}
    \caption{Visual results: GTA5 to Cityscapes}
    \label{fig:visualisations_gta2cs}
\end{figure*}